\documentclass[letterpaper,twocolumn,10pt]{article}

\usepackage{arXiv}
\usepackage{url}
\usepackage{amsmath, amsthm, amsfonts}
\usepackage{enumerate}
\usepackage{enumitem}
\usepackage{graphicx}
\usepackage{xcolor}
\usepackage{multirow}
\usepackage{comment}
\usepackage[utf8]{inputenc}
\usepackage{mathtools}
\usepackage{subcaption}
\usepackage{tabularx}
\usepackage{xspace}
\usepackage{colortbl}
\usepackage[noend, ruled]{algorithm2e}

\DeclareMathOperator*{\argmin}{arg\,min}

\usepackage{cases}
\usepackage{stfloats}
\usepackage{fancyhdr}
\usepackage{rotating}
\usepackage{oplotsymbl}
\usepackage{booktabs}

\usepackage{amsmath, amsthm, amsfonts}

\usepackage[noend, ruled]{algorithm2e}

\usepackage{booktabs}

\usepackage[pagebackref=true,breaklinks=true,colorlinks,bookmarks=false]{}
\usepackage{hyperref}
\hypersetup{
  colorlinks,
  linkcolor={blue!70!green},
  citecolor={green!70!blue},
  urlcolor={orange!70!red}
}
\setlist[itemize]{leftmargin=*} % 

\newcommand{\etal}{\textit{et al.}\xspace}

\newcommand{\eg}{\textit{e.g.}\xspace}
\definecolor{blue}{RGB}{0,0,0}
\definecolor{black}{RGB}{0,0,0}

\newcommand{\revstart}{\begin{blue}{revision}}
\newcommand{\revend}{~\!\!\end{blue}}

\renewcommand{\smallskip}{\vspace{2pt plus 0.5pt minus 0.5pt}}
\newcommand{\mypara}[1]{\noindent\textbf{#1.} \xspace}

\newcommand{\sysname}{DPMLBench\xspace}

\newcommand{\nalgs}{twelve\xspace}

\newcommand{\pate}{\textsf{PATE}\xspace}
\newcommand{\knn}{\textsf{Priv-kNN}\xspace}
\newcommand{\sigmoid}{\textsf{TanhAct}\xspace}
\newcommand{\loss}{\textsf{FocalLoss}\xspace}
\newcommand{\alloc}{\textsf{AdpAlloc}\xspace}

\newcommand{\dpsgd}{\textsf{DP-SGD}\xspace}
\newcommand{\wavelet}{\textsf{Hand-DP}\xspace}
\newcommand{\dpgen}{\textsf{DPGEN}\xspace}

\newcommand{\rgp}{\textsf{RGP}\xspace}
\newcommand{\gep}{\textsf{GEP}\xspace}
\newcommand{\adpclip}{\textsf{AdpClip}\xspace}
\newcommand{\lpmst}{\textsf{LP-MST}\xspace}
\newcommand{\alibi}{\textsf{ALIBI}\xspace}
\newcommand{\privateset}{\textsf{PrivSet}\xspace}

\newcommand{\simple}{SimpleCNN\xspace}
\newcommand{\resnet}{ResNet\xspace}
\newcommand{\inception}{InceptionNet\xspace}
\newcommand{\vgg}{VGG\xspace}

\usepackage{etoolbox}
\makeatletter
\patchcmd{\hyper@makecurrent}{%
    \ifx\Hy@param\Hy@chapterstring
        \let\Hy@param\Hy@chapapp
    \fi
}{%
    \iftoggle{inappendix}{%true-branch
        \@checkappendixparam{chapter}%
        \@checkappendixparam{section}%
        \@checkappendixparam{subsection}%
        \@checkappendixparam{subsubsection}%
        \@checkappendixparam{paragraph}%
        \@checkappendixparam{subparagraph}%
    }{}%
}{}{\errmessage{failed to patch}}

\newcommand*{\@checkappendixparam}[1]{%
    \def\@checkappendixparamtmp{#1}%
    \ifx\Hy@param\@checkappendixparamtmp
        \let\Hy@param\Hy@appendixstring
    \fi
}
\makeatletter

\newtoggle{inappendix}
\togglefalse{inappendix}

\apptocmd{\appendix}{\toggletrue{inappendix}}{}{\errmessage{failed to patch}}
\begin{document}

\begin{textblock}{16}(1.2,1)
To appear in the 30th ACM SIGSAC Conference on Computer and Communications Security. November 26-30, 2023.
\end{textblock}

\date{}

\title{\sysname: Holistic Evaluation of Differentially Private \\ Machine Learning}

\author{
Chengkun Wei\textsuperscript{{$\mathparagraph$}}\thanks{Chengkun and Minghu contributed equally to this work.}\ \ \
Minghu Zhao\textsuperscript{{$\mathparagraph$}{$\ast$}}\ \ \
Zhikun Zhang\textsuperscript{{$\ddagger$}{$\mathsection$}{$\mathparagraph$}}\thanks{Corresponding authors.}\ \ \
Min Chen\textsuperscript{{$\ddagger$}}\ \ \
\\
Wenlong Meng\textsuperscript{{$\mathparagraph$}}\ \ \
Bo Liu\textsuperscript{{$\|$}}\ \ \
Yuan Fan\textsuperscript{{$\mathparagraph$}}\ \ \
Wenzhi Chen\textsuperscript{{$\mathparagraph$}}\textsuperscript{{$\dagger$}}\ \ \
\\
\textsuperscript{{$\mathparagraph$}}\textit{Zhejiang University} \ \ \ 
\textsuperscript{{$\ddagger$}}\textit{CISPA Helmholtz Center for Information Security} \ \ \ 
\\
\textsuperscript{{$\mathsection$}}\textit{Stanford University} \ \ \
\textsuperscript{{$\|$}}\textit{DBAPPSecurity} \ \ \ 
}

\maketitle
\pagestyle{plain}

\begin{abstract}
Differential privacy (DP), as a rigorous mathematical definition quantifying privacy leakage, has become a well-accepted standard for privacy protection. 
Combined with powerful machine learning techniques, differentially private machine learning (DPML) is increasingly important.
As the most classic DPML algorithm, DP-SGD incurs a significant loss of utility, which hinders DPML's deployment in practice.
Many studies have recently proposed improved algorithms based on DP-SGD to mitigate utility loss.
However, these studies are isolated and cannot comprehensively measure the performance of improvements proposed in algorithms.
More importantly, there is a lack of comprehensive research to compare improvements in these DPML algorithms across utility, defensive capabilities, and generalizability.

We fill this gap by performing a holistic measurement of improved DPML algorithms on utility and defense capability against membership inference attacks (MIAs) on image classification tasks. 
We first present a taxonomy of where improvements are located in the machine learning life cycle.
Based on our taxonomy, we jointly perform an extensive measurement study of the improved DPML algorithms, over \nalgs algorithms, four model architectures, four datasets, two attacks, and various privacy budget configurations.
We also cover state-of-the-art label differential privacy (Label DP) algorithms in the evaluation.
According to our empirical results, DP can effectively defend against MIAs, and sensitivity-bounding techniques such as per-sample gradient clipping play an important role in defense.
We also explore some improvements that can maintain model utility and defend against MIAs more effectively.
Experiments show that Label DP algorithms achieve less utility loss but are fragile to MIAs.
Machine learning practitioners may benefit from these evaluations to select appropriate algorithms.
To support our evaluation, we implement a modular re-usable software, \sysname,\footnote{The implementation can be found at \url{https://github.com/DmsKinson/DPMLBench}} which enables sensitive data owners to deploy DPML algorithms and serves as a benchmark tool for researchers and practitioners.

\end{abstract}

\section{Introduction}
As \textit{machine learning} (ML) continues to evolve, numerous fields are leveraging its power to advance their development~\cite{guo2019survey,brunetti2018computer}; however, this often involves the use of private data, such as medical records.
Previous studies have revealed that the models trained on private data can leak information through a bunch of attacks, such as membership inference~\cite{shokri2017membership}, model inversion~\cite{fredrikson2015model}, and attribute inference~\cite{melis2019exploiting}, which raises critical privacy and security concerns.

\textit{Differential privacy} (DP) is a widely used notion to rigorously formalize and measure the privacy guarantee based on a parameter called \textit{privacy budget}.
Abadi \etal~\cite{abadi2016deep} proposed a general DPML algorithm called \textit{differentially private stochastic gradient descent} (DP-SGD) by integrating per-sample clipping and noise perturbation to the aggregated gradient in the training process.
However, models trained by DP-SGD normally perform badly with respect to model utility.
Recently, researchers proposed many \textit{improved algorithms} with better privacy-utility trade-off~\cite{yu2021large,shamsabadi2021losing,papernot2017semi,zhu2020private,papernot2021tempered,tramer2020differentially,yu2019differentially,ghazi2021deep,chen2022dpgen}.
In the rest of this paper, we refer to DP-SGD as \textit{vanilla DP-SGD} to distinguish between DP-SGD and the improved algorithms.

The improved algorithms modify the vanilla DP-SGD from different aspects but are evaluated in isolation with various settings, which cannot reveal the differences between each other.
Furthermore, existing studies~\cite{jayaraman2019evaluating,iyengar2019towards,zhao2020not,jarin2022dp} fail to report a complete and practical analysis of general DPML algorithms in practical scenarios. 
This motivates us to perform a holistic evaluation and analysis of these improved DPML algorithms.

\subsection{Our Contributions}
\mypara{Algorithm Taxonomy}
We first propose a new taxonomy for the state-of-the-art DPML algorithms based on their improved component in the ML pipeline.
Concretely, we divide the ML pipeline into four phases: Data preparation, model design, model training, and model ensemble (see \autoref{subsec:pipeline} for details), and categorize the DPML algorithms into each phase.
We then perform a theoretical and empirical analysis to obtain an extensive view of the impact of differential privacy on machine learning.

\mypara{Experimental Evaluation}
In this paper, we concentrate on \nalgs state-of-the-art DPML algorithms for image classification tasks.
We then conduct comprehensive experiments for these algorithms on four model architectures (ResNet20~\cite{he2016deep}, VGG16~\cite{simonyan2014very}, InceptionNet~\cite{krizhevsky2012imagenet}, and SimpleCNN) and four benchmark image datasets (MNIST~\cite{lecun1998gradient}, FashionMNIST~\cite{xiao2017fashion}, CIFAR-10~\cite{krizhevsky2009learning}, and SVHN~\cite{netzer2011reading}) to jointly evaluate the tradeoff between privacy protection, model utility, and defense effectiveness.
Furthermore, we evaluate the defensive capabilities of the DPML algorithms on both white-box and black-box \textit{membership inference attacks} (MIAs).
Our measurement aims to answer the following three research questions:

\begin{enumerate}[label=\textbf{RQ$\arabic*$.},itemindent=0.4cm]
  \item What improvements in DPML algorithms are most effective in maintaining model utility?
  \item What improvements in DPML algorithms are most robust in defending membership inference attacks?
  \item What is the impact of dataset and model architecture on algorithms focusing on different stages?
\end{enumerate}

In addition, our measurement covers two state-of-the-art label differential privacy (Label DP) algorithms, which is a variant DPML notion by relaxing the protection of the whole data sample to only protect the label. 
To the best of our knowledge, we are the first to analyze the Label DP algorithms on utility and defense empirically.

\mypara{\sysname}
We implement a toolkit called \sysname to support the comprehensive evaluation of DPML algorithms with respect to model utility and MIA defense.
With a modular design, \sysname can easily integrate additional DPML algorithms, attacks, datasets, and model architectures by implementing new functional codes to the relevant modules.
Our code will be publicly available, facilitating researchers to leverage existing DPML algorithms to provide DP guarantee or benchmark new algorithms.

\mypara{Main Findings}
Our work reveals several interesting findings:
\begin{itemize}
    \item 
    Different improvement techniques can affect the privacy-utility trade-offs of the algorithm from different perspectives.
    For example, we find that reducing the dimension of the parameter improves the performance of DPML on large models but may impair utility when the privacy budget is large.
    In addition, DP synthetic algorithms and algorithms in the model ensemble category are the most robust in defending against MIAs.
    \item DP can effectively defend against MIAs.
    Also, sensitivity-bounding techniques such as per-sample gradient clipping play an important role in defense.
    \item Some model architecture design choices for non-private ML models are ineffective for private ML models.
    For instance, using Tanh as the activation function and GroupNorm can reduce the utility loss on vanilla DP-SGD.
    However, we also find that using Tanh and GroupNorm together would have a negative effect.
    \item Compared to standard DP, Label DP has less utility loss but is more fragile to MIAs.
\end{itemize}

\section{Preliminaries}

\begin{figure*}[!t]
    \centering
    \includegraphics[width=0.95\linewidth]{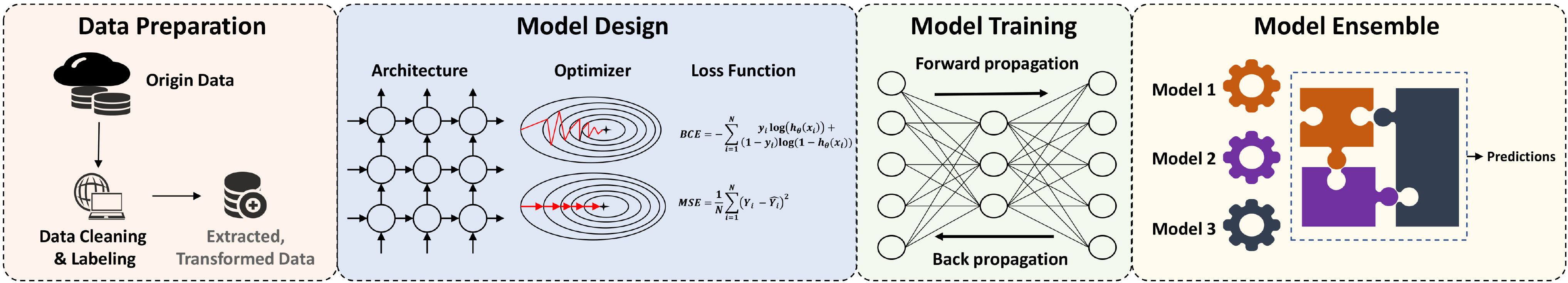}    
    \caption{Machine learning pipeline.
    }
    \label{fig:pipeline}
    \vspace{-0.3cm}
\end{figure*}

\subsection{Machine Learning Pipeline} 
\label{subsec:pipeline}

\autoref{fig:pipeline} illustrates a typical machine learning pipeline, which consists of four phases: Data preparation, model design, model training, and model ensemble.

The \emph{data preparation} phase aims to explore the underlying distribution of data for learning algorithms.
Commonly used techniques in this phase include data cleaning~\cite{chu2016data}, data labeling~\cite{fredriksson2020data}, and feature extraction~\cite{khalid2014survey}.
Feature extraction transforms the input data into a low-dimensional subspace that reveals the most relevant information~\cite{chumerin2006comparison}.
Low dimensional information can downgrade the difficulty of the following training procedures~\cite{yu2021not,yu2021large,zhou2020bypassing}.

In the \emph{model design} phase, we aim to select components such as the model architectures, loss functions, and optimization algorithms that are appropriate for the task.
There are plenty of studies on this topic~\cite{he2016deep,simonyan2014very}.

The \emph{model training} phase is the process of computing the following optimization objective:
$$
\argmin_{\theta} \frac{1}{|\mathit{D}_{train}|} \sum_{(x, y)\in \mathit{D}_{train}} \mathit{L}(y, \mathit{M}(x;\theta)),
$$
where $(x,y)$ is the data sample in the training dataset $\mathit{D}_{train}$;
$\mathit{L}$ and $\mathit{M}$ represent loss function and model architecture, respectively.
The parameters $\theta$ in model $\mathit{M}$ are optimized to minimize the objective function $\mathit{L}$ on training data during the model training phase.

The \emph{model ensemble} phase combines multiple models while deploying the model.
Previous studies show that aggregating multiple models' predictions can obtain better generalization performance than a single model~\cite{opitz1999popular}.

\subsection{Differential Privacy} 
\label{subsec:dp}

Differential privacy (DP)~\cite{dwork2008Differential} is a rigorous mathematical definition quantifying how much privacy preservation a mechanism can provide.
DP provides a privacy guarantee by bounding the impact of a single input on the mechanism's output.

\begin{definition}[($\epsilon$, $\delta$)-Differential Privacy] 
Given two neighboring datasets $D$ and $D'$ differing by one record, a mechanism $\mathit{M}$ satisfies ($\epsilon$, $\delta$)-differential privacy if
$$ 
Pr[\mathit{M}(D)\in S] \leq e^{\epsilon } \cdot Pr[\mathit{M}(D') \in S] + \delta,
$$
where $\epsilon$ is the privacy budget, and $\delta$ is the failure probability.
\end{definition} 

The privacy budget quantifies the maximum information a mechanism $\mathit{M}$ can expose.
A smaller privacy budget indicates better privacy preservation.
$\delta$ indicates the probability that $\mathit{M}$ fails to satisfy $\epsilon$-DP.
When $\delta = 0$, we achieve pure $\epsilon$-DP, a stronger notion, and a more rigorous privacy guarantee.

\mypara{Bounded DP and Unbounded DP}
How to interpret neighboring datasets distinguishes between \textit{bounded DP} and \textit{unbounded DP}~\cite{kifer2011no}.
In \textit{unbounded DP}, D and D$'$ are neighbors if D can be obtained from D$'$ by adding or removing one element.
In \textit{bounded DP}, D and D$'$ are neighbors if D can be obtained from D$'$ by replacing one element.
When using \textit{bounded DP}, two datasets should have the same number of elements.
Furthermore, any algorithms that satisfy $\epsilon$-\textit{unbounded DP} also satisfy 2$\epsilon$-\textit{bounded DP} because replacing one element can be achieved by removing then adding one element.
All algorithms in \autoref{tab:methods_table} satisfy the \textit{unbounded DP}.

\mypara{Gaussian Mechanism}
Adding noise sampled from Gaussian distribution is a commonly used approach to achieve ($\epsilon$, $\delta$)-DP, known as Gaussian mechanism~\cite{dwork2014algorithmic}.
Formally, applying the Gaussian mechanism to a function $\mathit{f}$ can be defined as:
$$
    \mathit{M}(d) = \mathit{f}(d) + \mathbf{N}(0, S_\mathit{f}^2 \cdot \sigma^2 ),
$$
where $\mathbf{N}(0, S_\mathit{f}^2 \cdot \sigma^2 )$ is the Gaussian distribution with mean 0 and standard deviation $S_\mathit{f}^2 \cdot \sigma^2$,
where $\sigma$ is called noise multiplier and $S_\mathit{f}$ is the sensitivity of function $\mathit{f}$.

\begin{definition}
(Sensitivity). Given two neighboring datasets $D$ and $D'$, the global sensitivity of a mechanism $\mathit{M}$, denoted by $\mathit{S}_\mathit{M}$, is given below
$$ \mathit{S}_\mathit{M} = \max_{D, D'}| \mathit{M}(D) - \mathit{M}(D') |.  $$
\end{definition}

\mypara{Composition}
The composition theorems calculate the total privacy budget when we apply DP on the private dataset multiple times.
The most straightforward composition strategy is summing up the privacy budget of each individual DP algorithm.
Formally, for $k$ DP algorithms with privacy budget $\epsilon_1, \epsilon_2, \epsilon_3, \cdots, \epsilon_k$, the total privacy budget is $\epsilon = \epsilon_1 + \epsilon_2 + \epsilon_3 + \cdots + \epsilon_k$.
Mironov~\cite{mironov2017renyi} et al. propose R\'enyi differential privacy to achieve a tighter analysis of cumulative privacy budgets.

\begin{definition}[$(\alpha, \delta)$-R\'enyi Differential Privacy (RDP)~\cite{mironov2017renyi}]

A randomized mechanism $\mathit{M}$ is said to satisfy $\epsilon$-Rényi differential privacy of order $\alpha$ (which can be abbreviated as $(\alpha, \delta)$-RDP), if for any adjacent datasets $D$, $D'$, it holds that
$$ 
D_\alpha(\mathit{M}(D)||\mathit{M}(D')) \leq \epsilon, 
$$
where $D_\alpha(\mathit{M}(D)||\mathit{M}(D'))$ is the $\alpha$-R\'enyi divergence between the distribution of $\mathit{M}(D)$ and the distribution of $\mathit{M}(D')$.
Parameter $\alpha$ controls the momentum of the privacy loss random variable.
\end{definition}

Note that larger $\alpha$ leads to more weight being assigned to worst-case events, \eg, $(\infty, \epsilon)$-RDP is equivalent to $\epsilon$-DP.
If $M$ satisfies $(\alpha, \epsilon)\text{-}RDP$, it also satisfies $(\epsilon+\frac{log\frac{1}{\delta} }{\alpha-1}, \delta)\text{-}DP$.
Applying $k$ algorithms with $\left(\alpha, \epsilon_1\right)$-RDP, $\left(\alpha, \epsilon_2\right)$-RDP, $\cdots$ ,$\left(\alpha, \epsilon_k\right)$-RDP on same dataset sequentially leads to an algorithm with $\left(\alpha, \epsilon_1+\epsilon_2+\cdots+\epsilon_k\right)$-RDP.
By selecting $\alpha$ delicately, accumulating privacy loss in RDP and then converting to DP can derive a tighter upper bound than composite $(\epsilon, \delta)$-DP directly.

\mypara{Post-processing}
The post-processing property guarantees that no matter what additional processing one performs on the output of an algorithm that satisfies ($\epsilon$,$\delta$)-DP, the composition of the algorithm and the post-processing operations still satisfy ($\epsilon$,$\delta$)-DP.

\subsection{Differentially Private Machine Learning} 
\label{sec:dpml}

Abadi \etal~\cite{abadi2016deep} integrated differential privacy with stochastic gradient descent (SGD) and proposed a general learning algorithm named differential privacy stochastic gradient descent (DP-SGD). 
Compared to SGD, DP-SGD introduced a few modifications to make the algorithm satisfy differential privacy.
Firstly, the sensitivity of each gradient is bounded by clipping each gradient in the $l_2$ norm.
\begin{equation}
    clip(\textbf{g},C) = \textbf{g}/max(1,\frac{||\textbf{g}||_2}{C}). 
\end{equation}
Per-sample clipping bounds the contribution of each sample to model parameters to $C$.
Moreover, DP-SGD applies a Gaussian mechanism to the aggregated clipped gradient.
Formally,
\begin{equation} \label{equ:gau-mech}
    \Tilde{\textbf{g}} = \textbf{g} + \mathit{N}(0,C^2\sigma^2), 
\end{equation}
where $\Tilde{\textbf{g}}$ is the noisy gradient used to update parameters and $\sigma$ controls privacy level.
After the above two steps, the gradients used to update the parameters satisfy DP.

Nevertheless, gradient clipping and noise perturbation introduce deviation in the training process, which impairs the model's utility. 
Recently, researchers proposed a number of improved DPML algorithms to reduce the utility loss incurred by vanilla DP-SGD~\cite{papernot2018scalable,papernot2021tempered,yu2021not,tramer2020differentially}.
However, these improved DPML algorithms were evaluated on different models and datasets with different assumptions.
Therefore, it is a pressing need to design a holistic benchmark to comprehensively evaluate these DPML algorithms to gain a deeper insight.

\begin{table*}[!t]
  \centering
  \footnotesize
  \caption{Overview and comparison of DPML algorithms. 
  *: Evaluation is based on subsequent private model training on generated data.
  \circlet: same as non-private training.
  \circletfillhl: with modification but no noise adding.
  \circletfill: with modification and noise adding.}
  \label{tab:methods_table}
  \renewcommand{\arraystretch}{1.3}
  \resizebox{0.9\linewidth}{!}{
    % Table generated by Excel2LaTeX from sheet 'properties'

    \begin{tabular}{cc|c|c|c|c|c|c|c}
    \toprule
    \multicolumn{2}{c|}{} & \textbf{Algorithms} & \textbf{Auxiliary Data} & \textbf{Private Data} & \textbf{Model Architecture} & \textbf{ Gradient} & \textbf{ Loss Function} & \textbf{Perturbation} \\
    \hline
    % \hline
    \multicolumn{2}{c|}{\multirow{3}{*}{Data Preparation}} & \wavelet & \circlet     & \circletfillhl     & \circlet     & \circletfill     & \circlet     & Gradient \\
\cline{3-9}    \multicolumn{2}{c|}{} & \privateset* & \circlet     & \circletfill     & \circlet     & \circlet     & \circlet     & Input \\
\cline{3-9}    \multicolumn{2}{c|}{} & \dpgen* & \circlet     & \circletfill     & \circlet     & \circlet     & \circlet     & Input \\
    \hline
    \multicolumn{2}{c|}{\multirow{2}{*}{Model Design}} & \sigmoid  & \circlet     & \circlet     & \circletfillhl     & \circletfill     & \circlet     & Gradient \\
\cline{3-9}    \multicolumn{2}{c|}{} & \loss  & \circlet     & \circlet     & \circlet     & \circletfill     & \circletfillhl     & Gradient \\
    \hline
    \multicolumn{2}{c|}{\multirow{5}{*}{Model Training}} & Vanilla DP-SGD & \circlet     & \circlet     & \circlet     & \circletfill     & \circlet     & Gradient \\
\cline{3-9}    \multicolumn{2}{c|}{} & \rgp   & \circlet     & \circlet     & \circletfillhl     & \circletfill     & \circlet     & Gradient \\
\cline{3-9}    \multicolumn{2}{c|}{} & \gep   & \circletfillhl     & \circlet     & \circlet     & \circletfill     & \circlet     & Gradient \\
\cline{3-9}    \multicolumn{2}{c|}{} & \alloc & \circlet     & \circlet     & \circlet     & \circletfill     & \circlet     & Gradient \\
\cline{3-9}    \multicolumn{2}{c|}{} & \adpclip & \circlet     & \circlet     & \circlet     & \circletfill     & \circlet     & Gradient \\
\hline
\multicolumn{2}{c|}{\multirow{2}{*}{Model Ensemble}} & \pate  & \circletfill     & \circlet     & \circlet     & \circlet     & \circlet     & Input \\
\cline{3-9}    \multicolumn{2}{c|}{} & \knn & \circletfill     & \circlet     & \circlet     & \circlet     & \circlet     & Input \\
    \bottomrule
    \end{tabular}%

  }
\end{table*}

\subsection{Membership Inference in Machine Learning Models} 
\label{sec:MIA}

The MIAs have become one of the most widely studied~\cite{salem2018ml,nasr2019comprehensive} attacks against ML models after Shokri \etal proposed in~\cite{shokri2017membership}.
The MIA aims to infer whether a data sample is used to train the target ML model.
Formally, MIA $\mathit{A}$ can be defined as:
$$
    \mathit{A} : \mathit{I}, \mathit{M}, \mathbf{x}  \xrightarrow{}  \{ 0,1 \},
$$
where $\mathit{I}$ is the auxiliary knowledge of adversary, $\mathit{M}$ is the model to be attacked, and $\mathbf{x}$ is a data sample. 
$\mathit{A}$ can be seen as a binary classifier, where 1 means the data sample $\mathbf{x}$ is used for training model $\mathit{M}$, namely a member, and 0 otherwise.
It is natural to use MIAs to evaluate the defensive capabilities of DPML algorithms, as in many previous studies~\cite{jayaraman2019evaluating,jarin2022dp,song2021systematic}.

Based on the information an attacker can obtain, MIAs can be classified into two categories: White-box and black-box.
The white-box attacks have full access to the target model, while black-box attacks only have query access to the target model and obtain the prediction confidence vector. 
We adopt both types of MIAs to comprehensively evaluate the defensive capabilities of the DPML algorithms (in \autoref{subsec:defense-analysis}).

\section{Taxonomy}
\label{sec:taxonomy}

In this section, we provide an overview of our taxonomy and give survey-style descriptions of the DPML algorithms.

\subsection{Overview}   
\label{subsec:tax-overview}

We first propose a new taxonomy for the DPML algorithms based on the component they improve in the ML pipeline discussed in \autoref{subsec:pipeline}.
We introduce this taxonomy due to the following reasons: (1) The training phases of ML are independent, meaning the improvements in different phases might be combined to achieve better model utility.
(2) It provides future researchers with a clear roadmap to improve the DPML algorithms, which we hope can benefit the community.
(3) It is domain-agnostic and can be easily extended to evaluate the DPML algorithms in other domains, such as graph and NLP data.

\autoref{tab:methods_table} summarizes all the improved DPML algorithms and their corresponding categories.
We also discuss the properties of all the DPML algorithms.
For instance, vanilla DP-SGD falls in the model training category and modifies the gradient to provide the DP guarantee, whereas \pate belongs to the model ensemble category and leverages auxiliary data to provide a DP guarantee.
Auxiliary data generally refers to data with the same distribution as sensitive data but is publicly available, which is a common assumption in DPML~\cite{yu2021not,papernot2017semi,zhou2020bypassing}.

\mypara{Data Preparation}
The algorithms in this category pre-process the original training data.
Feature extraction and DP synthetic data are two typical approaches in this category. 
Feature extraction aims to reduce the difficulty of private training.
Using a pre-trained network before classifier~\cite{abadi2016deep,yu2019differentially,kurakin2022toward} can be seen as a variant of feature extraction.
DP synthetic data aims to provide a DP guarantee for training data.
Applying DP mechanisms to data directly, such as the Gaussian mechanism, downgrades the utility of data, especially when data is in high dimension (\eg, image).
DP synthetic data is an alternative that aims to generate data in a DP manner with a similar distribution as sensitive data.
Training models on synthetic data with traditional machine learning algorithms can derive a model with DP guaranteed according to post-processing property~\cite{chen2020gswgan,chen2022dpgen,torkzadehmahani2019dpcgan}.
In this category, we pick three algorithms, of which \wavelet~\cite{tramer2020differentially} leverages a feature extractor, and the other two (\privateset~\cite{chen2022private} and \dpgen~\cite{chen2022dpgen}) belong to DP synthetic data algorithms.

\mypara{Model Design}
Algorithms in this category focus on designing more adapted model designs to DPML.
Deep learning in non-private settings has been widely studied, and many rules have been summarized to train a standard neural network.
However, these design guidelines do not perform well in vanilla DP-SGD~\cite{kurakin2022toward} due to gradient clipping and noise perturbation.
For instance, larger models often mean better performance in non-private settings.
However, smaller models tend to get better performance on vanilla DP-SGD.
Some existing studies focus on exploring more adapted model design rules to DPML~\cite{papernot2021tempered,cheng2022dpnas}.
We select two algorithms in this category, and they propose improvements from the activation function~ (\sigmoid~\cite{papernot2021tempered}) and loss function (\loss~\cite{shamsabadi2021losing}), respectively.

\mypara{Model Training}
Algorithms in this category explore DP mechanisms with less impact on model utility in the DP-SGD training phase.
The vanilla DP-SGD~\cite{abadi2016deep} bounds the $l_2$-norm of gradient $g$ by clipping the gradient to the threshold $C$; thus, a straightforward improvement strategy is to find an optimal clipping strategy~\cite{pichapati2019adaclip,andrew2021differentially}.
On the other hand, the noise perturbation leads to bias during model updating, which impairs the model's utility. 
Therefore, designing a better noise perturbation mechanism to alleviate the noise effect is another optimization option~\cite{yu2021not,zhou2020bypassing,yu2019differentially}.
In this category, we select four algorithms, excluding vanilla DP-SGD.
\adpclip~\cite{andrew2021differentially} proposes an improved clipping strategy, and the rest of them (\rgp~\cite{yu2021large}, \gep~\cite{yu2021not}, and \alloc~\cite{yu2019differentially}) explore better noise perturbation mechanisms.

\mypara{Model Ensemble}
This category contains algorithms providing DP guarantee through the model ensemble.
The vanilla DP-SGD has poor scalability because it requires modifications to the training process.
Papernot \etal~\cite{papernot2017semi} propose Private Aggregation of Teacher Ensemble (PATE) by leveraging model ensemble.
PATE treats the training phase of the model as a black box so that it has better scalability than vanilla DP-SGD for less modification to the training process.
DP mechanism is applied while aggregating the prediction of multiple models.
Since then, many DPML algorithms based on the model ensemble have emerged~\cite{papernot2018scalable,zhu2020private}.
We select \pate and \knn in our measurement.

\subsection{Data Preparation} 
\mypara{\wavelet~\cite{tramer2020differentially}} 
Tramer \etal leverage Scattering Network (ScatterNet)~\cite{oyallon2015deep}, a feature extractor that encodes images using a cascade of wavelet transforms to extract the features.
To achieve the DP guarantee, they fine-tuned a model on top of extracted features through DP-SGD.

\mypara{\dpgen~\cite{chen2022dpgen}} 
It is an instantiation of the DP variant of the Energy-based Model (EBM)~\cite{du2019implicit,lecun2006tutorial}, which aims to privatize Langevin Markov Chain Monte Carlo (MCMC) sampling method~\cite{younes1999convergence} to synthesize images, of which an energy-based network guides the movement directions.
\dpgen achieves DP by using Randomized Response (RR) in movement direction selection.
Compared to other DP-SGD based synthesis methods~\cite{xie2018differentially,torkzadehmahani2019dpcgan}, \dpgen can generate higher-resolution images. 

\mypara{\privateset~\cite{chen2022private}}
It leverages dataset condensation to generate data in a differentially private manner.
It directly optimizes for a small set of samples promising to derive approximate results under downstream tasks instead of imitating the complete data distribution.
Specifically, they use DP-SGD to optimize a gradient-matching objective for the downstream task that minimizes the difference between the gradient on the real data and the generated data.

\subsection{Model Design}
\mypara{\sigmoid~\cite{papernot2021tempered}}
Considering the need for DP to bound sensitivity,
Papernot \etal~\cite{papernot2021tempered} replace ReLU with tempered sigmoid as the activation function. 
The authors found that the bounded property of tempered sigmoid functions, especially Tanh, can effectively limit the $l_2$-norm of the gradient while training models with DP-SGD.
Thus, less information can be lost in gradient clipping.

\mypara{\loss~\cite{shamsabadi2021losing}}
It introduces a loss function adapted to vanilla DP-SGD, which combines three terms: 
The summed squared error $\mathit{L}_{\text {Focal }}$, the focal loss $\mathit{L}_{\mathrm{SSE}}$~\cite{lin2017focal}, and a regularization penalty on the intermediate pre-activations $\mathit{L}_{\text {Reg }}$.
Finally, they proposed loss function $\mathit{L}$:
\begin{equation}
  \mathit{L}=\alpha \mathit{L}_{\text {Focal }}+(1-\alpha) \mathit{L}_{\mathrm{SSE}}+\frac{(1-\alpha)}{\beta} \mathit{L}_{\text {Reg }},
\end{equation}
where $\alpha=\textit{Sigmoid}(e_c-e_t)$ (current epoch $e_c$, and threshold epoch $e_t$), $\beta$ is the hyperparameter controlling the strength of the regularization.
These terms consider convergence speed, emphasis on complex samples, and sensitivity during training.
The new loss function can better control the gradient sensitivity in the training procedure.

\subsection{Model Training}

\mypara{\rgp~\cite{yu2021large}}
It adopts a reparametrization scheme to replace the model weight in each layer with two low-dimensional weight matrices and a residual weight matrix:
\begin{equation}
  \mathbf{W} \rightarrow \mathbf{L} \mathbf{R}+\tilde{\mathbf{W}}.\text{stop\_gradient}\left(\right).
\end{equation}
By making the gradient carriers $\{\mathbf{L},\mathbf{R}\}$ consist of orthonormal vectors, a projection of the gradient of $\mathbf{W}$ can be constructed from the noisy gradients of $\tilde{\mathbf{L}}$ and $\tilde{\mathbf{R}}$.
$\{\mathbf{L},\mathbf{R}\}$ are trained by DP-SGD separately to achieve the DP guarantee and finally combined to obtain the gradient for updating the model.
Note that the dimensionality of $\mathbf{L}$ and $\mathbf{R}$ is much smaller than that of $\mathbf{W}$.
Thus RGP can reduce the storage consumption and the noise added to the model.

\mypara{\gep~\cite{yu2021not}}
Yu \etal observe that the number of noise increases with the growth of model size in vanilla DP-SGD and figure out a solution, \gep~\cite{yu2021not}, to reduce the dimension of the gradient before adding noise. 
GEP first computes an anchor subspace that contains some gradients of public data via the power method.  
Then, it projects the gradient of private data into the anchor subspace to produce a low-dimensional gradient embedding and a small-norm residual gradient.
The two parts are applied with the DP mechanism separately and combined to update the original weight.
Compared to RGP, GEP leverages public data to  decompose the original model parameters for dimensionality reduction.

\mypara{\alloc~\cite{yu2019differentially}}
It proposes a dynamic noise-adding mechanism instead of keeping noise multiplier $\sigma$ constant every training epoch in vanilla DP-SGD.
It replaces the variance in the Gaussian mechanism with a function of the epoch:
\begin{equation}
  \mathit{M}(d) = \mathit{f}(d) + \mathbf{N}(0, S_\mathit{f}^2 \cdot \sigma_t^2 ),
\end{equation}
the value of $\sigma_t$ depends on the final privacy budget, epoch, and schedule function.
The schedule function defines how the noise scale is adjusted during training.
Yu \etal proposed several pre-defined schedules.
We select \textit{Exponential Decay} in our evaluation, which has the best average performance in \cite{yu2019differentially}.
The mathematical form of \textit{Exponential Decay} is 
$\sigma_t=\sigma_0 e^{-k t}$, 
where $k (k>0)$ is decay rate and $\sigma_0$ is the initial noise scale.

\mypara{\adpclip~\cite{andrew2021differentially}}
It uses an adaptive clipping threshold mechanism, which sets the clip threshold to a specified quantile of the update norm distribution every epoch.
Formally, clipping threshold $C_t$ in epoch $t$ can be computed as
$  C_t = C_{t-1} \cdot exp(-\eta_C(\overline{b}-\gamma ) )$
, where $\gamma \in \left[0,1\right] $ is a quantile to be matched, $ \overline{b}  \triangleq \frac{1}{m} \sum_{i \in[m]} \mathbb{I}_{x_i \leq C}$ is the empirical fraction of samples with value at most $C$ , and $\eta_C$ is the learning rate with default value of 0.2 in \cite{andrew2021differentially}.
To address the issue that $\overline{b}$ reveals private information, Gaussian mechanism is applied to $\overline{b}$: $\tilde{b}^t=\frac{1}{m}\left(\sum_{i \in \mathit{Q}^t} b_i^t+\mathit{N}\left(O, \sigma_b^2\right)\right)$.
The method consumes a negligible privacy budget to track the quantile closely.
AdpClip was originally designed for federated learning (FL) but can be extended to traditional centralized learning scenarios.

\subsection{Model Ensemble}
\label{subsec:model-ensemble}
\mypara{\pate~\cite{papernot2017semi}} 
It first trains multiple teacher models with disjoint private data.
The teacher ensemble is later used to label the public data, and the noise perturbation is applied to the voting aggregation before generating a prediction.
The student model, which gives the final output, is trained from labeled public data and cannot directly access private data.
The privacy budget is determined by the noise added to the votes and the number of queries to the teacher ensemble.
Additionally, PATE leverages a semi-supervised learning method to reduce the queries to the teacher ensemble.

\mypara{\knn~\cite{zhu2020private}}
In PATE, a larger number of teacher models lead to a larger absolute lead gap while aggregating votes, potentially allowing for a larger noise level. 
At the same time, splitting data makes each teacher model hold only partial original training data, which causes a model utility drop.
Thus, Zhu \etal~\cite{zhu2020private} propose a data-efficient scheme based on the private release of k-nearest neighbor (kNN) queries to replace teacher ensemble, which avoids splitting the training dataset.
For every given data sample from the public domain, \knn subsamples a random subset from the entire private dataset.
Then it picks the $k$ nearest neighbors from the subset in feature space, equivalent to $k$ teachers' prediction in vanilla PATE.

\section{\sysname}
\begin{figure}[!t]
    \centering
    \includegraphics[width=0.85\linewidth]{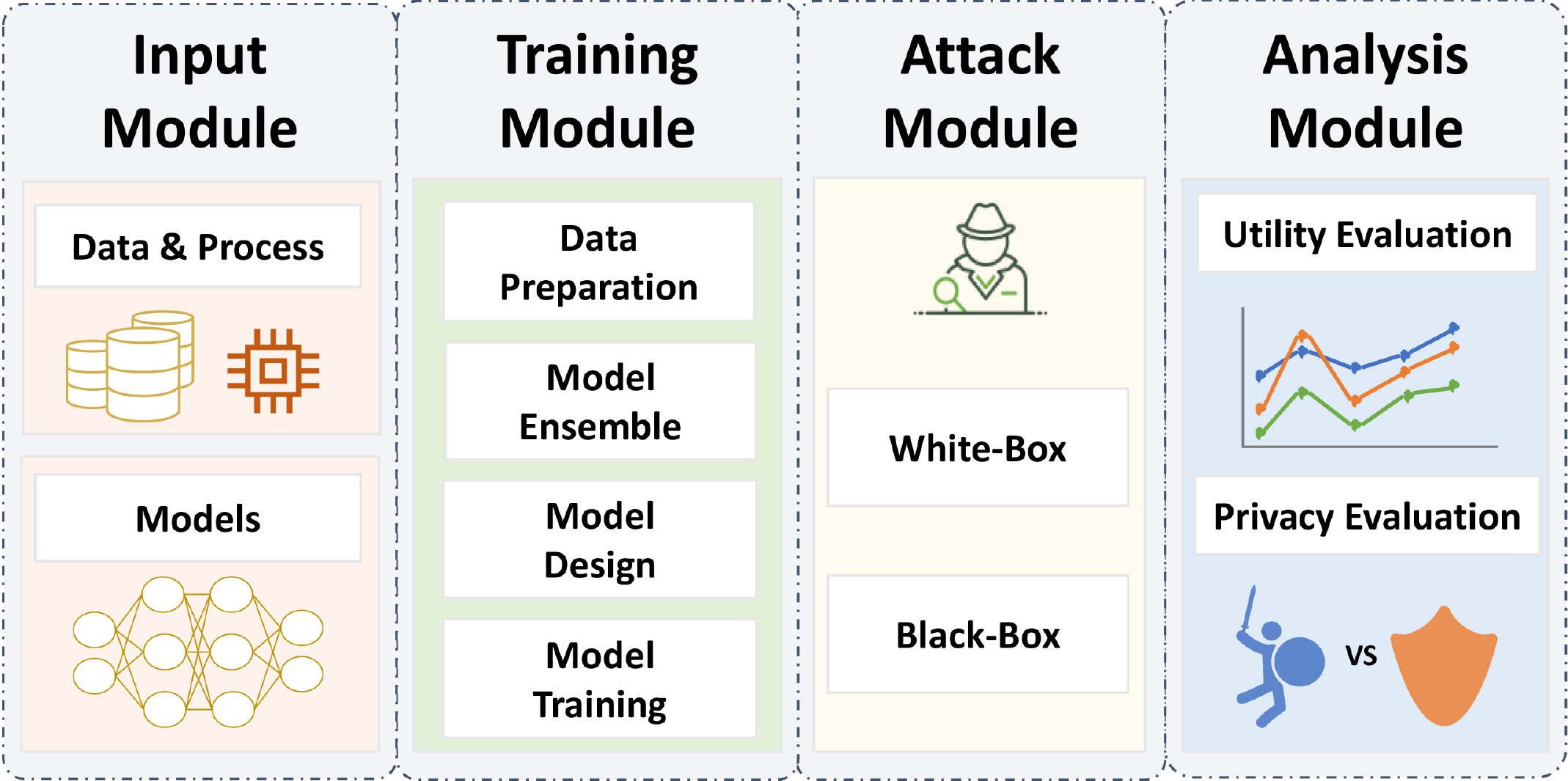}
    \caption{Overview of \sysname.}
    \label{fig:Overview}
\end{figure}

This section introduces \sysname, a modular toolkit designed to evaluate DPML algorithms' performance on utility and privacy.
\autoref{fig:Overview} illustrates the four modules of \sysname.

\begin{enumerate}

    \item \mypara{Input} 
    This module prepares the dataset and model for the following modules.
    For dataset, it involves dataset partition and preprocessing \eg, normalization.
    For the model, it constructs model architectures and does necessary modifications for private training (see \autoref{subsec:setup}).
        
    \item \mypara{Training} 
    This module performs the DPML algorithms to train DPML models.
    It currently supports \nalgs different DPML algorithms into four categories (see \autoref{sec:taxonomy}).
    
    \item \mypara{Attack}
    This module performs two MIAs on models trained from the training module.

    \item \mypara{Analysis}
    This module evaluates the performance of DPML algorithms on utility and privacy.
\end{enumerate}

\sysname follows a modular design that makes it flexible to integrate new algorithms, attacks, datasets, and models.
We envisage that \sysname can be used for the following purposes:
\begin{itemize}
    \item As we have implemented \nalgs representative DPML algorithms, \sysname enables data owners to train their privacy-preserving models with these DPML algorithms efficiently.

    \item \sysname comprehensively assesses different DPML algorithms in utility and privacy.
    Researchers can re-use \sysname as a benchmark tool to evaluate other DPML algorithms and attacks in the future.

    \item Since \sysname follows a modular design, modules are connected through abstract interfaces.
    To integrate a new DPML algorithm and attack or to extend \sysname into different domains, users can re-implement processing functions in the corresponding modules and reuse other modules directly.
\end{itemize}

\section{Experiments}
Based on the proposed taxonomy, we present a series of comprehensive experiments to answer the following questions:
\begin{enumerate}
[label=\textbf{RQ$\arabic*$.},itemindent=0.4cm]
    \item What improvements in DPML algorithms are most effective in maintaining model utility?
    \item What improvements in DPML algorithms are most robust in defending MIAs?
    \item What is the impact of dataset and model architecture on algorithms focusing on different stages?
\end{enumerate}

\subsection{Experimental Setup}
\label{subsec:setup}

\mypara{DPML Algorithms}
We implement \nalgs DPML algorithms; their details can be found in \autoref{sec:taxonomy}.
For \gep, \rgp, \knn, \dpgen, and \privateset, we use implementations of authors and modify codes to adapt for our evaluation.
The rest of the algorithms are implemented by PyTorch~\cite{paszke2019pytorch} and Opacus~\cite{yousefpour2021opacus}.

\mypara{Datasets}
We conduct experiments on four datasets: MNIST~\cite{lecun1998gradient}, FashionMNIST~\cite{xiao2017fashion}, CIFAR-10~\cite{krizhevsky2009learning}, and SVHN~\cite{netzer2011reading}, which are widely used in evaluating privacy-preserving machine learning~\cite{abadi2016deep,yu2021not,papernot2017semi,zhu2020private}.
We resize all images to 32x32 in our evaluation.

Since our attacks are all under the assumption that the attacker has an auxiliary dataset that shares similar distribution with the training data, we split each dataset into four disjoint parts: shadow training set, shadow testing set, target training set, and target testing set.
Additionally, we allocate $90\%$ of the data originally used for testing as public data for the algorithms in the ensemble category.

\begin{table}[!tbp]
  \centering
  \caption{The testing accuracy, tailored AUC of MIAs in black-box/white-box of baseline models. The number of parameters follows each model name. (Accuracy(\%)/black-box/white-box)}
  \label{tab:baseline}
  \resizebox{\linewidth}{!}
  {
    \begin{tabular}{ccccc}
    \toprule
     Target Model & MNIST & FMNIST & SVHN  & CIFAR-10 \\
    \midrule
    \textbf{SimpleNet} (0.17M) & 98.42/0.50/0.50 & 88.04/0.54/0.54 & 87.69/0.64/0.53 & 69.50/0.78/0.72 \\
    \textbf{ResNet} (0.26M) & 99.12/0.50/0.50 & 89.16/0.52/0.54 & 92.88/0.57/0.59 & 66.56/0.77/0.63 \\
    \textbf{InceptionNet} (1.97M) & 99.18/0.51/0.50 & 90.92/0.56/0.53 & 95.08/0.55/0.57 & 83.52/0.71/0.68 \\
    \textbf{VGG} (128.8M) & 98.70/0.50/0.52 & 90.74/0.59/0.56 & 91.91/0.62/0.56 & 72.96/0.78/0.73 \\
    \bottomrule
    \end{tabular}
  }
\end{table}

\mypara{Model Architectures}
We focus on four model architectures, including ResNet20~\cite{he2016deep}, VGG16~\cite{simonyan2014very}, InceptionNet~\cite{krizhevsky2012imagenet}, and a simple three convolution layer network as SimpleCNN.
Batch normalization makes each sample's normalized value depend on its peers in a batch, making it hard to restrict a single data contribution to the output.
To adapt differential privacy, we replace all batch normalization~\cite{sergey2015batch} with group normalization~\cite{Wu2018group}.
We regard the models trained with the same hyperparameters without DP as the baseline to evaluate utility loss.
\autoref{tab:baseline} shows the performance of the baseline model across datasets, including testing accuracy and tailored AUC against black-box/white-box MIAs.

We use MLPs for black-box and white-box model architecture for the attack implementation as in \cite{liu2021ml-doctor,nasr2019comprehensive}.
A detailed description of the model architecture can be found in \autoref{apx:model-details}.

\mypara{Hyperparameters}
We use R\'enyi DP to accumulate the overall privacy budget and precompute the required noise scale ($\sigma \text{ in DP-SGD}$) numerically~\cite{abadi2016deep,mironov2019renyi}.
We keep $\delta=10^{-5}$ and use different privacy budgets: $\epsilon = \{0.2,0.3,0.4,0.5,1,2,4,8,100,1000\}$.
All algorithms' clipping threshold $C$ are fixed to 4 unless the algorithm has special clipping strategies.

We use the hyperparameters obtained by grid search on DP-SGD if the original paper does not mention the setting.
While searching hyperparameters, we refer to the guides of recent studies on hyperparameter settings for private training~\cite{de2022unlocking,kurakin2022toward}.
For simplicity, we ignore the privacy leakage caused by hyperparameter tuning in our experiment~\cite{papernot2021hyperparameter}.
For the attack models, we follow the settings in \cite{liu2021ml-doctor}, where the batch size is 64, the epoch is 50, the optimizer is Adam, and the learning rate is $10^{-5}$.
\autoref{apx:hyper-setting} shows the detailed hyperparameter settings.

\mypara{Metrics}
Following previous studies~\cite{jayaraman2019evaluating,liu2021ml-doctor,jarin2022dp,zhao2020not}, we use accuracy ACC to evaluate the models' utility and the area under ROC curve (AUC) to evaluate the defense ability of the model.
In MIAs, AUC lower than 0.5 indicates that the inference attack performs worse than a random guess and tends to infer non-members as members.
Thus we set the lower bound of AUC to 0.5 for analysis convenience, indicating that AUC=0.5 implies no privacy leakage.
We process the AUC metric as follows:
$$
 \widetilde{\text{AUC}} = \textit{max}(\text{AUC},0.5),
$$
We name $\widetilde{\text{AUC}}$ as tailored AUC, which is always between 0.5 and 1.

To compare the performance of DPML algorithms and non-private algorithms more directly, we define proportional metric \textbf{utility loss} and \textbf{privacy leakage}, respectively:
\begin{equation}
    \text{Utility Loss} = 1 - \frac{\text{ACC}_{\mathit{M}_{pri}}}{\textit{ACC}_{\mathit{M}_{base}}},
\end{equation}
    
\begin{equation}
  \text{Privacy Leakage} = \frac{\widetilde{\text{AUC}}_{\mathit{M}_{pri}}-0.5}{\widetilde{\text{AUC}}_{\mathit{M}_{base}}-0.5},
\end{equation}
where $\mathit{M}_{pri}$ presents a private model trained by a DPML algorithm and $\mathit{M}_{base}$ presents a non-private model trained by vanilla SGD with the same settings as $\mathit{M}_{pri}$.
The utility loss denotes the percentage loss in accuracy of the DP model on the same test set relative to the normal model.
The private leakage denotes the proportion of privacy models’ privacy leakage compared to the normal model.

\subsection{Evaluation on Utility Loss} 
\label{subsec:utility-analysis}

\begin{table*}[!t]
    \centering
    \setlength{\tabcolsep}{0.11cm}
    \setlength{\abovecaptionskip}{0.2cm}
    \caption{Overview of algorithms' utility loss on different model architectures, datasets, and privacy budgets.  
    For each privacy budget, we bold the value with the best performance (with the \textbf{smallest} value of utility loss).
    The experimental results of GEP on VGG are unavailable due to memory limits.
    }  
    \label{tab:acc}
    \resizebox{\textwidth}{!}
    {
        \begin{tabular}{c|l|ccccc|ccccc|ccccc}
\toprule
 \multirow{2}{*}{} & \multirow{2}{*}{} & \multicolumn{5}{c|}{SimpleCNN} & \multicolumn{5}{c|}{ResNet} & \multicolumn{5}{c}{VGG} \\
 &  & 0.2 & 1 & 4 & 100 & 1000 & 0.2 & 1 & 4 & 100 & 1000 & 0.2 & 1 & 4 & 100 & 1000 \\
\midrule
\multirow[c]{12}{*}{\rotatebox{90}{MNIST}} 
& \wavelet & $89.19\pm1.04$ & $18.98\pm1.67$ & $8.46\pm0.35$ & $4.19\pm0.17$ & $3.73\pm0.14$
& $24.38\pm1.93$ & $11.58\pm0.76$ & $5.73\pm0.46$ & $2.88\pm0.50$ & $2.01\pm0.68$ 
& $88.63\pm1.25$ & $88.89\pm1.23$ & $90.46\pm0.16$ & $7.98\pm0.15$ & $3.80\pm0.78$ \\
 & \privateset & $77.26\pm5.89$ & $47.54\pm9.23$ & $30.86\pm2.26$ & $19.72\pm2.28$ & $17.58\pm2.60$ 
 & $89.70\pm0.20$ & $82.24\pm2.70$ & $57.01\pm6.11$ & $20.83\pm3.60$ & $17.86\pm2.67$ 
 & $81.36\pm11.11$ & $58.61\pm5.03$ & $51.32\pm22.10$ & $52.40\pm27.64$ & $66.52\pm18.65$ \\
& \dpgen & $60.99\pm8.05$ & $88.26\pm0.83$ & $14.95\pm0.58$ & $2.48\pm0.43$ & $2.70\pm0.26$ 
& $70.75\pm4.84$ & $84.07\pm3.96$ & $73.73\pm3.77$ & $3.64\pm0.36$ & $3.79\pm0.25$ 
& $90.08\pm0.04$ & $90.08\pm0.04$ & $58.28\pm30.47$ & $\textbf{1.76}\pm0.14$ & $2.20\pm0.28$ \\
 
 & \sigmoid & $85.19\pm2.58$ & $18.30\pm1.29$ & $\textbf{3.13}\pm0.39$ & $\textbf{1.74}\pm0.14$ & $\textbf{1.74}\pm0.16$ 
 & $38.16\pm3.27$ & $18.61\pm3.42$ & $7.05\pm0.54$ & $3.89\pm0.44$ & $3.12\pm0.19$ 
 & $90.21\pm0.04$ & $89.59\pm0.12$ & $90.15\pm0.31$ & $6.14\pm0.08$ & $\textbf{1.91}\pm0.03$ \\
 & \loss & $87.99\pm1.97$ & $29.11\pm2.59$ & $7.44\pm0.40$ & $3.32\pm0.07$ & $2.40\pm0.01$ 
 & $43.09\pm4.17$ & $10.99\pm1.70$ & $6.32\pm0.85$ & $2.72\pm0.12$ & $1.78\pm0.12$ 
 & $82.10\pm1.77$ & $88.16\pm0.01$ & $88.91\pm0.39$ & $11.46\pm11.64$ & $3.66\pm0.52$ \\
 & \dpsgd & $89.61\pm1.12$ & $21.98\pm0.19$ & $7.50\pm0.42$ & $3.58\pm0.21$ & $3.04\pm0.23$ 
 & $28.17\pm3.48$ & $11.33\pm1.17$ & $5.38\pm0.85$ & $2.88\pm0.35$ & $2.24\pm0.36$ 
 & $88.79\pm1.02$ & $88.74\pm0.47$ & $90.56\pm0.77$ & $13.23\pm4.42$ & $3.55\pm0.06$ \\
 & \rgp & $\textbf{36.65}\pm1.04$ & $\textbf{13.23}\pm0.78$ & $10.17\pm1.02$ & $6.72\pm0.09$ & $6.37\pm0.23$ 
 & $31.87\pm2.62$ & $21.24\pm3.90$ & $33.03\pm7.12$ & $34.06\pm5.62$ & $37.66\pm8.97$ 
 & $90.30\pm0.29$ & $\textbf{6.59}\pm0.95$ & $\textbf{3.86}\pm0.27$ & $6.33\pm2.24$ & $4.78\pm0.14$ \\
 & \gep & $90.30\pm0.29$ & $90.25\pm0.34$ & $14.37\pm1.83$ & $2.67\pm0.52$ & $1.52\pm0.02$ & $86.22\pm2.16$ & $17.61\pm1.35$ & $\textbf{4.36}\pm0.22$ & $\textbf{1.00}\pm0.04$ & $\textbf{0.46}\pm0.26$ & - & - & - & - & - \\
 & \alloc & $89.23\pm0.81$ & $18.79\pm1.24$ & $6.57\pm0.20$ & $3.59\pm0.45$ & $3.14\pm0.29$ & $\textbf{24.26}\pm3.70$ & $10.04\pm2.13$ & $4.91\pm0.59$ & $3.25\pm0.57$ & $2.48\pm0.37$ 
  & $90.24\pm0.22$ & $89.00\pm0.58$ & $89.85\pm1.04$ & $6.44\pm0.26$ & $3.12\pm0.04$ \\
 & \adpclip & $88.17\pm4.04$ & $75.55\pm11.05$ & $7.79\pm0.37$ & $8.00\pm0.30$ & $8.10\pm0.28$ 
 & $59.66\pm2.95$ & $\textbf{7.46}\pm0.53$ & $4.85\pm0.40$ & $4.17\pm0.35$ & $4.21\pm0.41$ 
 & $88.92\pm0.22$ & $88.13\pm0.24$ & $89.06\pm0.75$ & $14.00\pm2.44$ & $4.78\pm0.15$ \\
 & \pate & $82.83\pm3.94$ & $71.81\pm4.09$ & $33.36\pm12.33$ & $11.89\pm4.04$ & $10.98\pm2.78$ 
 & $90.86\pm0.09$ & $85.89\pm6.08$ & $30.74\pm15.79$ & $7.12\pm3.66$ & $10.38\pm1.16$ 
 & $84.94\pm3.47$ & $76.17\pm3.78$ & $44.08\pm23.93$ & $32.15\pm40.09$ & $32.66\pm39.74$ \\
 & \knn & $61.22\pm2.13$ & $34.97\pm3.37$ & $33.13\pm0.62$ & $34.77\pm0.94$ & $33.80\pm1.56$ 
 & $25.03\pm5.76$ & $9.70\pm0.87$ & $8.43\pm0.63$ & $9.30\pm0.74$ & $9.91\pm1.25$ 
 & $\textbf{49.05}\pm1.62$ & $17.80\pm2.17$ & $17.16\pm0.94$ & $16.29\pm0.39$ & $14.74\pm1.43$ \\
\cline{1-17}
\multirow[c]{12}{*}{\rotatebox{90}{CIFAR-10}} 
 & \wavelet & $90.06\pm0.16$ & $86.88\pm4.04$ & $48.67\pm0.96$ & $43.28\pm0.74$ & $44.14\pm0.18$ 
 & $84.29\pm2.26$ & $58.34\pm0.50$ & $50.74\pm0.58$ & $41.95\pm2.33$ & $39.29\pm3.11$ 
 & $90.03\pm0.18$ & $89.67\pm0.07$ & $89.86\pm0.12$ & $79.74\pm7.71$ & $37.58\pm0.82$ \\
 & \privateset & $\textbf{88.85}\pm0.65$ & $87.78\pm1.28$ & $86.78\pm0.71$ & $88.83\pm0.94$ & $89.43\pm0.78$ 
 & $89.56\pm0.83$ & $89.28\pm0.37$ & $89.45\pm1.18$ & $87.91\pm2.03$ & $85.43\pm2.43$  
 & $89.77\pm0.29$ & $88.14\pm0.78$ & $89.07\pm0.38$ & $90.04\pm0.26$ & $87.64\pm2.81$ \\
& \dpgen & $90.16\pm0.11$ & $89.86\pm0.09$ & $89.66\pm0.70$ & $69.98\pm2.28$ & $76.98\pm2.26$ 
& $90.00\pm0.35$ & $90.00\pm0.16$ & $90.59\pm1.39$ & $79.51\pm1.01$ & $83.38\pm4.20$ 
& $90.52\pm0.30$ & $89.72\pm0.21$ & $89.47\pm0.29$ & $87.24\pm3.07$ & $88.86\pm1.43$ \\
 & \sigmoid & $89.74\pm0.64$ & $69.95\pm1.13$ & $\textbf{45.39}\pm0.92$ & $\textbf{32.93}\pm0.55$ & $32.21\pm0.21$ 
 & $82.55\pm1.17$ & $62.11\pm0.33$ & $55.52\pm0.32$ & $48.95\pm1.17$ & $49.28\pm2.73$ 
 & $90.22\pm0.00$ & $90.07\pm0.25$ & $90.13\pm0.10$ & $64.46\pm1.43$ & $\textbf{34.26}\pm0.28$ \\
 & \loss & $89.88\pm0.08$ & $88.17\pm2.46$ & $52.42\pm0.47$ & $38.55\pm0.79$ & $38.47\pm0.90$ 
 & $84.36\pm2.43$ & $62.12\pm0.53$ & $52.06\pm0.43$ & $40.65\pm2.12$ & $39.00\pm2.98$  
 & $90.17\pm0.26$ & $89.75\pm0.15$ & $89.85\pm0.23$ & $66.36\pm6.19$ & $36.60\pm0.33$ \\
 & \dpsgd & $89.80\pm0.30$ & $89.13\pm1.30$ & $48.79\pm0.24$ & $40.03\pm0.93$ & $40.48\pm0.86$ 
 & $81.92\pm2.61$ & $58.32\pm0.44$ & $49.57\pm1.76$ & $41.17\pm2.67$ & $38.66\pm3.80$ 
 & $90.20\pm0.56$ & $89.38\pm0.49$ & $89.73\pm0.09$ & $89.81\pm0.12$ & $35.15\pm0.35$ \\
 & \rgp & $90.15\pm0.02$ & $\textbf{61.91}\pm1.32$ & $58.52\pm1.34$ & $54.28\pm0.68$ & $54.41\pm0.92$ 
 & $\textbf{74.48}\pm0.54$ & $65.24\pm0.88$ & $67.27\pm2.00$ & $66.38\pm0.93$ & $66.56\pm0.82$ 
 & $90.16\pm0.01$ & $\textbf{81.87}\pm4.19$ & $\textbf{53.66}\pm1.25$ & $53.49\pm0.09$ & $54.37\pm0.49$ \\
 & \gep & $90.16\pm0.00$ & $90.16\pm0.01$ & $90.16\pm0.00$ & $35.11\pm0.20$ & $\textbf{31.90}\pm0.24$ & $88.68\pm2.19$ & $85.19\pm0.20$ & $\textbf{46.72}\pm0.73$ & $\textbf{30.45}\pm0.36$ & $\textbf{26.64}\pm0.93$ & - & - & - & - & - \\
 & \alloc & $90.04\pm0.30$ & $89.89\pm0.18$ & $47.97\pm0.57$ & $38.49\pm0.47$ & $39.16\pm0.88$ 
 & $80.04\pm2.19$ & $\textbf{57.88}\pm0.86$ & $48.86\pm1.03$ & $43.83\pm1.52$ & $42.22\pm2.19$ 
 & $90.06\pm0.05$ & $89.57\pm0.05$ & $89.99\pm0.08$ & $\textbf{51.42}\pm0.58$ & $35.46\pm0.42$ \\
 & \adpclip & $89.71\pm0.23$ & $89.79\pm0.26$ & $64.50\pm2.33$ & $35.64\pm0.82$ & $34.12\pm0.42$ 
 & $86.57\pm1.43$ & $64.08\pm0.77$ & $48.05\pm1.15$ & $37.17\pm1.27$ & $33.55\pm1.98$ 
 & $\textbf{89.70}\pm0.34$ & $89.86\pm0.68$ & $90.21\pm0.19$ & $89.69\pm0.01$ & $44.47\pm0.07$ \\
 & \pate & $90.19\pm1.30$ & $91.70\pm1.44$ & $89.25\pm0.53$ & $83.30\pm2.42$ & $83.06\pm0.54$ 
 & $88.34\pm0.41$ & $87.60\pm1.13$ & $85.99\pm2.08$ & $82.05\pm1.18$ & $83.50\pm1.43$ 
 & $90.05\pm1.23$ & $91.60\pm0.49$ & $91.06\pm1.11$ & $89.92\pm1.56$ & $90.02\pm2.79$ \\
 & \knn & $89.52\pm0.45$ & $89.38\pm0.12$ & $88.94\pm0.17$ & $90.19\pm0.05$ & $90.29\pm0.40$ 
 & $87.77\pm1.70$ & $81.35\pm1.56$ & $77.38\pm1.27$ & $74.96\pm0.34$ & $74.43\pm1.27$ 
 & $89.85\pm1.27$ & $87.26\pm2.55$ & $85.42\pm2.04$ & $84.81\pm1.32$ & $84.41\pm1.56$ \\
\bottomrule
\end{tabular}

    }
    \vspace{-0.2cm}
\end{table*}

\mypara{Overview}
\autoref{tab:acc} reports an overview of algorithms' utility loss across model architectures, datasets, and privacy budgets.
Due to space limitations, we only show part of the experimental results.
The rest results can be found in \autoref{apx:additional-results} (\autoref{tab:acc-apx}, which shows the similar trend as \autoref{tab:acc}.).
The experimental results for \gep on \inception and \vgg are unavailable due to memory limit.
For brevity, we use a $\langle Alg, Model, Dataset,\epsilon \rangle$ tuple to denote the $Model$ trained with $Alg$ on $Dataset$ in the case of privacy budget $\epsilon$.
For instance, $\langle \rgp, \resnet, MNIST, 0.2 \rangle$ indicates the \resnet model trained by \rgp with a privacy budget of 0.2 on MNIST.

We observe that the utility loss decreases with increasing privacy budget for all algorithms, which intuitively shows that the noise scale hurts the model's utility.
However, the utility loss varies widely across algorithms for the same privacy budget.
We analyze improved DPML algorithms' utility loss across four categories in the following.
\textit{NonPrivate} in figures denotes the model trained by normal SGD without DP.

\begin{figure*}[!t]
    \centering
    \begin{subfigure}[t]{\textwidth}
    \centering
        \includegraphics[width=0.9\linewidth]{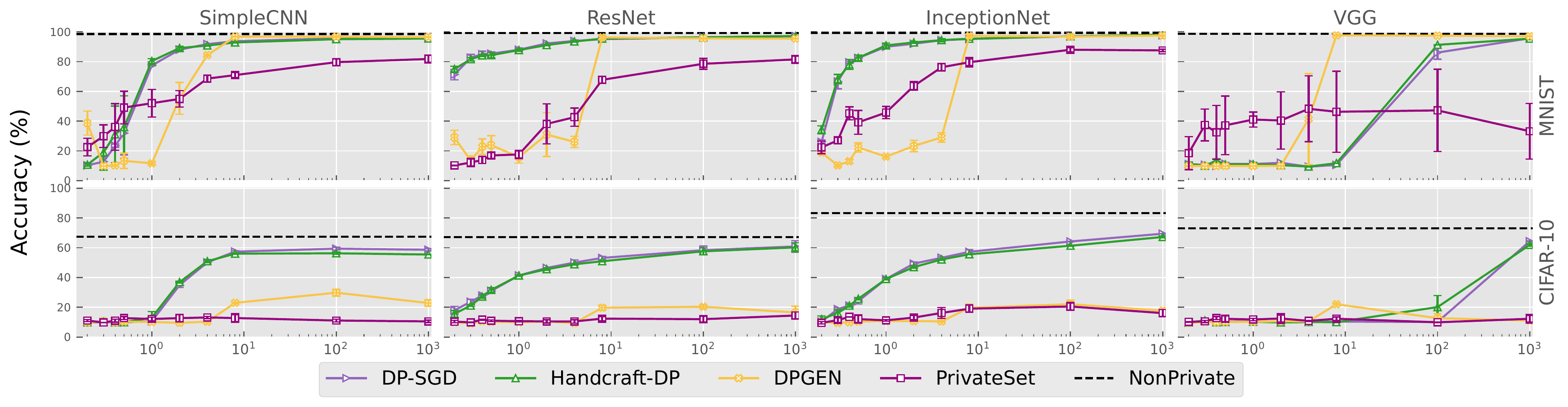}
        \vspace{-0.2cm}
        \caption{Data Preparation}
        \label{fig:acc_data_preparation}
    \end{subfigure}
    \begin{subfigure}[t]{\textwidth}
    \centering
        \includegraphics[width=0.9\linewidth]{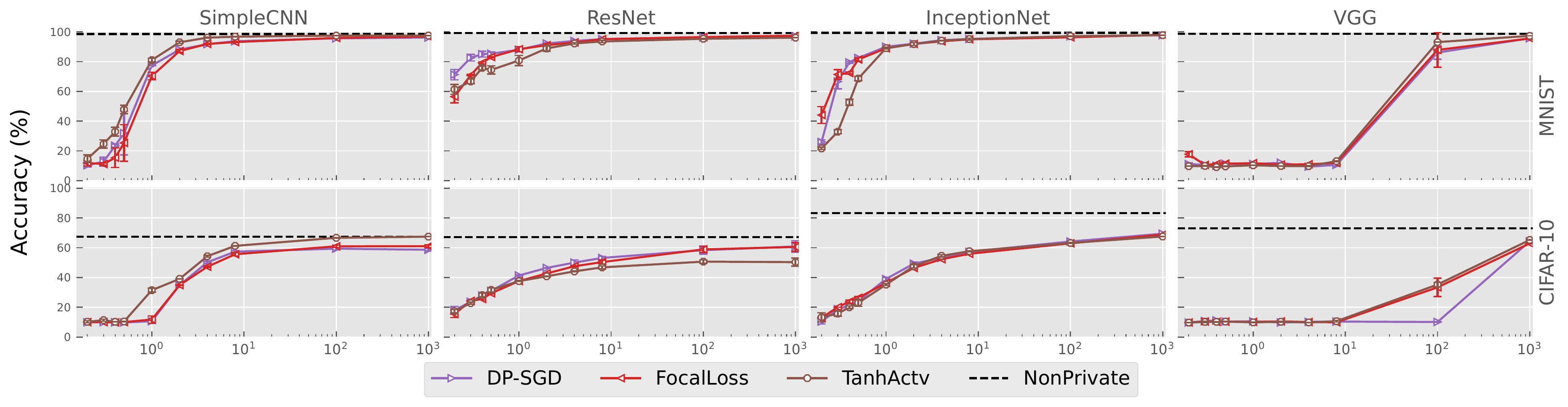}
        \vspace{-0.2cm}
        \caption{Model Design}
        \label{fig:acc_model_design}
    \end{subfigure}
    \begin{subfigure}[t]{\textwidth}
    \centering
        \includegraphics[width=0.9\linewidth]{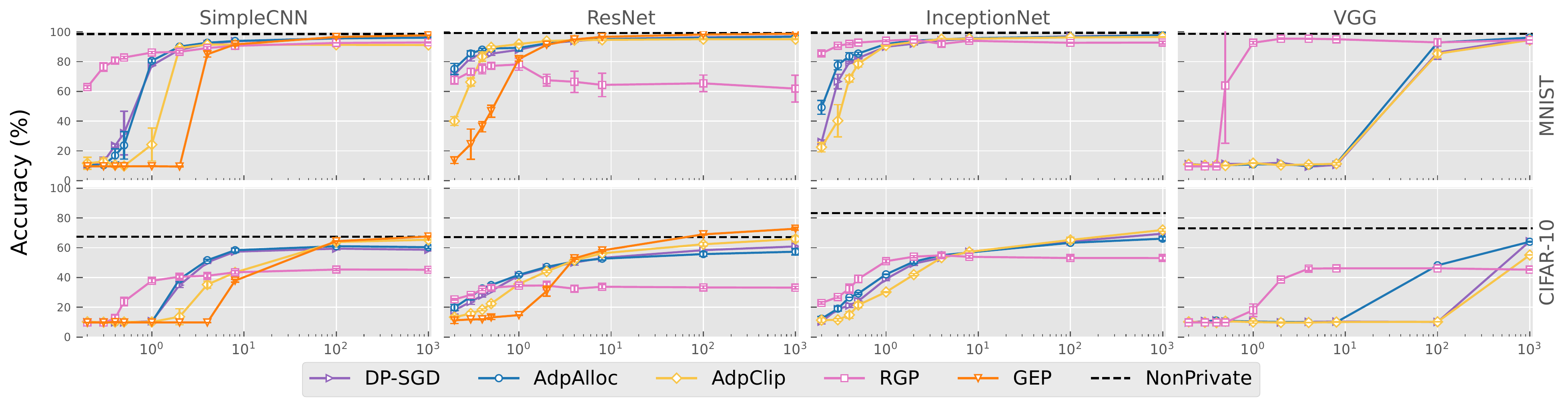}
        \vspace{-0.2cm}
        \caption{Model Training}
        \label{fig:acc_model_train}
    \end{subfigure}
    \begin{subfigure}[t]{\textwidth}
    \centering
        \includegraphics[width=0.9\linewidth]{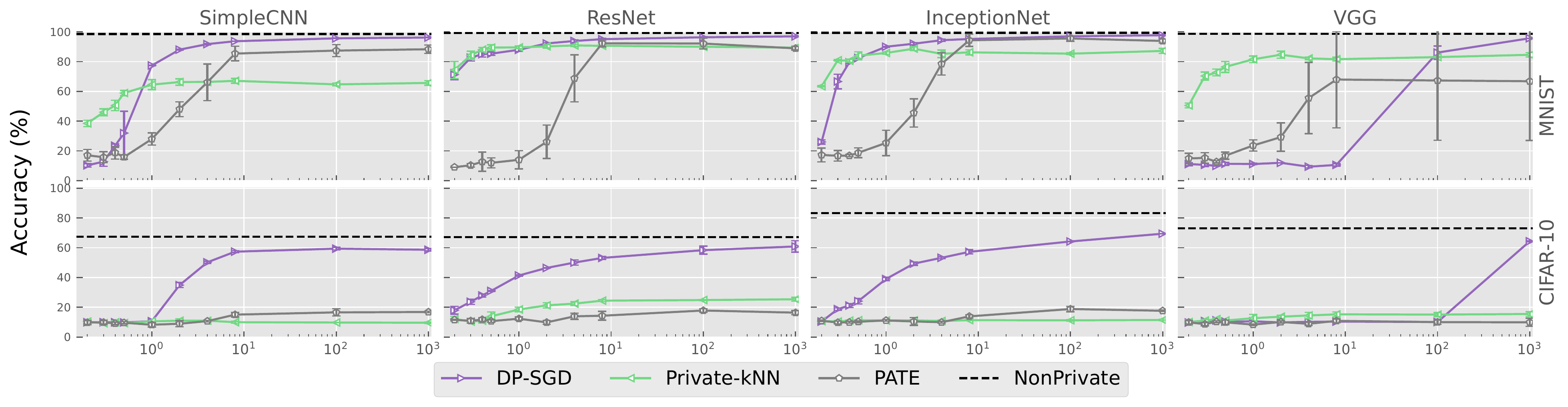}
        \vspace{-0.2cm}
        \caption{Model Ensemble}
        \label{fig:acc_model_ensemble}
    \end{subfigure}
    \vspace{-0.4cm}
    \caption{Accuracy comparison of the DPML algorithms in four categories, where the x-axis represents privacy budgets.
    }
\label{fig:acc}
\vspace{-0.3cm}
\end{figure*}

\mypara{Data Preparation}
Initially, in \cite{chen2022dpgen}, the classifier was trained on private data in order to label the synthetic data, and then the labeled dataset was used to train the target model.
This is similar to labeling public data through teacher ensemble in~\cite{papernot2017semi}, which will consume additional privacy budgets.
However, \cite{chen2022dpgen} does not count this part.
In our implementation, we use data that does not overlap with private data to train the labeling model.

\autoref{fig:acc_data_preparation} illustrates the accuracy comparison between algorithms in the data preparation category and vanilla DP-SGD.
The plot shows that \wavelet outperforms \dpgen and \privateset in low privacy budget generally.
\wavelet's accuracy is equivalent to vanilla DP-SGD and has a slight advantage on \vgg.
The performance of \dpgen and \privateset is highly relative to the quality of synthetic data.
When manually inspecting the generated data, we observe that there exist images with wrong labels and many similar, even identical images (\eg, mode collapse).
More effort on hyperparameter tuning and manual data filtering for DP synthetic algorithms can improve the performance.

Moreover, Tramer \etal propose using the non-learned handcrafted feature to train a linear model with DP-SGD~\cite{tramer2020differentially}.
Thus, we perform the same experiment for \wavelet on simple MLP.
The experiment results on CIFAR-10 are shown as \autoref{tab:acc-handcraft-cmp} in \autoref{apx:additional-results}.
Comparing other model architectures, we observe that the simple MLP only has an advantage when the privacy budget is relatively small (\eg, $\epsilon < 0.5$ ).
Thus, we exclude the MLP in subsequent experiments to maintain uniformity with other algorithms.

\mypara{Model Design}
\autoref{fig:acc_model_design} illustrates the performance of algorithms in the model design category and vanilla DP-SGD.

In general, \sigmoid outperforms vanilla DP-SGD and \loss on \simple and \vgg.
\textit{However, neither \sigmoid nor \loss performs better than vanilla DP-SGD on \resnet and \inception, \sigmoid's performance is even much worse than vanilla DP-SGD on \resnet.
\cite{papernot2021tempered} shows that \sigmoid has a better utility-privacy trade-off on their models, whose architecture is similar to \simple.}
The difference among the architectures is that \resnet and \inception both have GroupNorm layers while the others do not.

\begin{figure}[!t]
    \centering
    \includegraphics[width=0.9\linewidth]{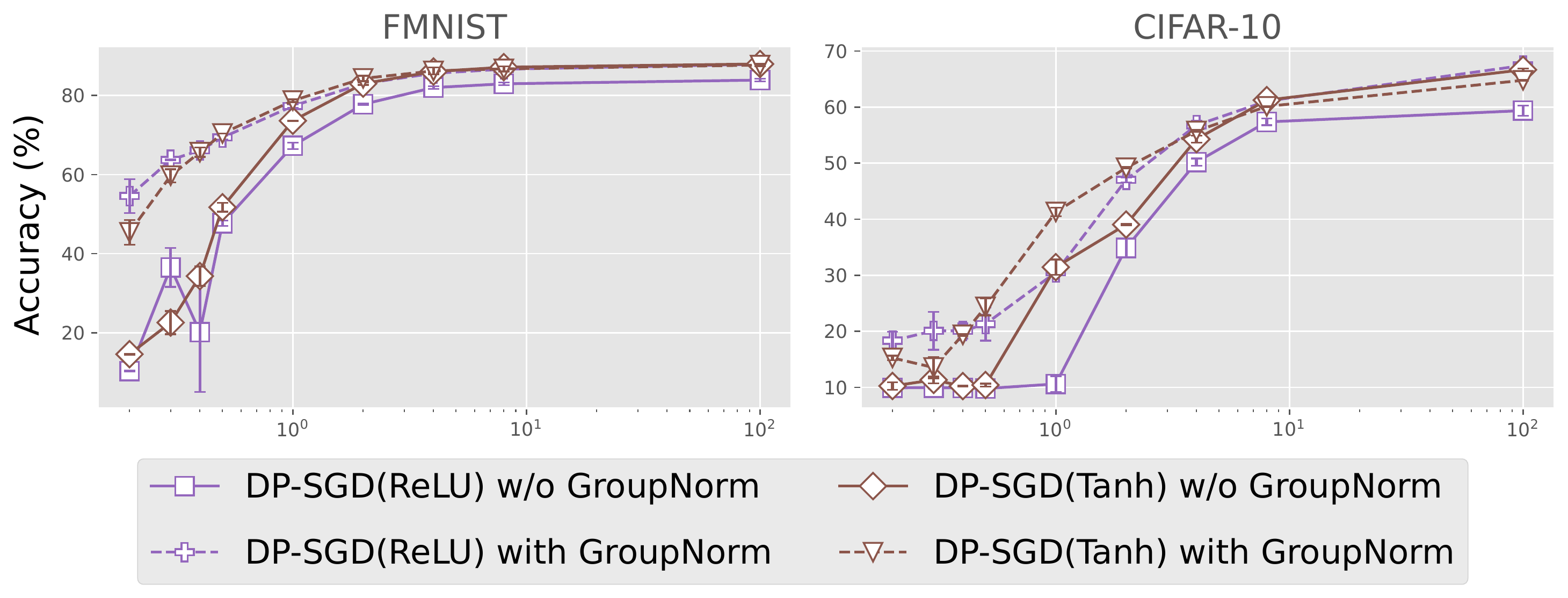}
    \vspace{-0.3cm}
    \caption{Accuracy of \simple models with/without GroupNorm layer trained by DP-SGD with ReLU and Tanh activation function across varies privacy budget.}
    \label{tab:tanh_norm}
    \vspace{-0.6cm}
\end{figure}

To figure out the impact of the GroupNorm layer and activation function, 
we add the GroupNorm layer before the activation function of the \simple and evaluate the performance of the vanilla DP-SGD (DP-SGD with ReLU) and \sigmoid (DP-SGD with Tanh) respectively (in \autoref{tab:tanh_norm}).
We observe that the GroupNorm layer improves the accuracy of the model overall.
However, the improvement gap shrinks as the privacy budget increases when using Tanh as an activation function, \eg DP-SGD (Tanh) w/o GroupNorm outperforms DP-SGD (Tanh) with GroupNorm when the privacy budget is greater than 10.
The connection between the activation function and the normalization layer needs further exploration.

\mypara{Model Training}
\autoref{fig:acc_model_train} illustrates the accuracy comparison of algorithms in the model training category and vanilla DP-SGD.
When the privacy budget is large, the accuracy of \gep exceeds the baseline in some settings (\eg $\langle \gep, \resnet, \operatorname{CIFAR-10},1000 \rangle$) because of leveraging public data.

When the privacy budget is small, \rgp is the only algorithm in this category to achieve acceptable performance on \vgg.
Model parameter dimensionality reduction is an effective technique to solve large models' inability to adapt to DP.
Nevertheless, there is a significant performance degradation when the privacy budget is large for \rgp.
We suspect the reason is that reparameterization not only reduces the noise scale in private training but also leads to information loss in the gradient.
When the noise scale is small, the information loss caused by reparametrization is higher than the mitigation effect on noise perturbation.

\autoref{tab:rgp_nodp} in \autoref{apx:additional-results} reports the accuracy of \rgp (w/o DP) and vanilla SGD, and the difference between them is whether using reparametrization.
We train models by using \rgp (w/o DP) and vanilla SGD, respectively, 
\xspace, and the difference between them is whether using reparametrization.
Overall, the accuracy of \rgp (w/o DP) is lower than that of SGD under the same settings across all datasets and model architectures.
The results can be found in \autoref{tab:rgp_nodp} in \autoref{apx:additional-results}.
The results echo our previous speculation that reparametrization reduces noise scale in private training but impairs performance in non-private settings.

\mypara{Model Ensemble}
\autoref{fig:acc_model_ensemble} illustrates the accuracy of the algorithms in the model ensemble category and vanilla DP-SGD.

Note that \knn and \pate use noise screening technique~\cite{zhu2020private,papernot2018scalable}, which ignores the data with low confidence in teacher ensembles to improve the utility-privacy tradeoff.
We do not use this technique in our implementation because the privacy budget is given in our settings and the noise scale is precomputed, which requires a fixed number of queries.

When implemented on \vgg, \knn can preserve an equivalent performance as other models, whereas \pate's performance plunges to random guesses.
A large number of teachers can impair the noise effect, while the amount of data allocated to each teacher model is too small for a large model such as \vgg to converge.
The results echo the introduction in \autoref{subsec:model-ensemble}, \pate is hard to get a good trade-off on the number of teacher models.

When implemented on other model architectures, \knn \xspace outperforms \pate with a low privacy budget and vice versa with a high privacy budget.
\textit{\pate and \knn both show higher accuracy at some specific settings \cite{papernot2017semi,zhu2020private}. However, they both fail to obtain a better utility-privacy trade-off than vanilla DP-SGD at most settings in our measurements.
We suspect that semi-supervised training techniques introduce more randomness and require fine-grained hyperparameter tunning, which leads to a high standard deviation as our experimental results show.}

\subsection{Evaluation on Defensive Capabilities}
\label{subsec:defense-analysis}

\begin{table*}
  \centering
  \setlength{\abovecaptionskip}{0.2cm}
  \caption{Overview of algorithms' tailored AUC in black-box MIA on different model architectures and privacy budgets. 
  In every setting, we bold the value with the best performance (with the \textbf{smallest} value).
  The experimental results for GEP on InceptionNet and VGG are unavailable due to memory limits.}
  \label{tab:mia}
  \resizebox{\linewidth}{!}{
    \begin{tabular}{c|l|ccccc|ccccc|ccccc}
\toprule
\multirow{2}{*}{} & \multirow{2}{*}{} & \multicolumn{5}{c|}{SimpleCNN} & \multicolumn{5}{c|}{ResNet} & \multicolumn{5}{c}{VGG} \\
 &  & 0.2 & 1 & 4 & 100 & 1000 & 0.2 & 1 & 4 & 100 & 1000 & 0.2 & 1 & 4 & 100 & 1000 \\
\midrule
\multirow[c]{12}{*}{\rotatebox{90}{CIFAR-10}} 
 & \wavelet & $\textbf{0.50}\pm0.00$ & $\textbf{0.50}\pm0.00$ & $0.51\pm0.01$ & $0.53\pm0.00$ & $0.53\pm0.00$ 
 & $\textbf{0.50}\pm0.00$ & $\textbf{0.50}\pm0.00$ & $\textbf{0.50}\pm0.00$ & $0.52\pm0.00$ & $0.53\pm0.00$ 
 & $\textbf{0.50}\pm0.00$ & $\textbf{0.50}\pm0.00$ & $\textbf{0.50}\pm0.00$ & $\textbf{0.50}\pm0.00$ & $0.53\pm0.01$ \\
 & \privateset & $\textbf{0.50}\pm0.00$ & $0.51\pm0.00$ & $\textbf{0.50}\pm0.00$ & $\textbf{0.50}\pm0.00$ & $\textbf{0.50}\pm0.00$ 
 & $\textbf{0.50}\pm0.00$ & $\textbf{0.50}\pm0.00$ & $\textbf{0.50}\pm0.00$ & $\textbf{0.50}\pm0.00$ & $0.51\pm0.01$ 
 & $\textbf{0.50}\pm0.00$ & $\textbf{0.50}\pm0.00$ & $\textbf{0.50}\pm0.00$ & $0.51\pm0.00$ & $\textbf{0.50}\pm0.00$ \\
 & \dpgen & $0.51\pm0.00$ & $\textbf{\textbf{0.50}}\pm0.00$ & $0.51\pm0.00$ & $\textbf{\textbf{0.50}}\pm0.00$ & $\textbf{0.50}\pm0.00$ 
 & $\textbf{0.50}\pm0.00$ & $\textbf{0.50}\pm0.00$ & $\textbf{0.50}\pm0.00$ & $\textbf{0.50}\pm0.00$ & $\textbf{0.50}\pm0.00$ 
 & $\textbf{0.50}\pm0.00$ & $\textbf{0.50}\pm0.00$ & $0.51\pm0.00$ & $\textbf{0.50}\pm0.00$ & $\textbf{0.50}\pm0.01$ \\
 & \sigmoid & $\textbf{0.50}\pm0.00$ & $\textbf{0.50}\pm0.00$ & $0.51\pm0.00$ & $0.53\pm0.00$ & $0.53\pm0.00$ 
 & $\textbf{0.50}\pm0.00$ & $\textbf{0.50}\pm0.00$ & $0.51\pm0.00$ & $0.52\pm0.00$ & $0.55\pm0.01$  
 & $\textbf{0.50}\pm0.00$ & $\textbf{0.50}\pm0.00$ & $\textbf{0.50}\pm0.00$ & $0.51\pm0.00$ & $0.54\pm0.00$ \\
 & \loss & $\textbf{0.50}\pm0.00$ & $0.51\pm0.00$ & $0.51\pm0.00$ & $0.52\pm0.00$ & $0.53\pm0.00$ 
 & $\textbf{0.50}\pm0.00$ & $\textbf{0.50}\pm0.00$ & $0.51\pm0.00$ & $0.52\pm0.00$ & $0.52\pm0.00$ 
 & $\textbf{0.50}\pm0.00$ & $\textbf{0.50}\pm0.00$ & $\textbf{0.50}\pm0.00$ & $\textbf{0.50}\pm0.00$ & $0.53\pm0.00$ \\
 & \dpsgd & $0.51\pm0.00$ & $\textbf{0.50}\pm0.00$ & $0.51\pm0.01$ & $0.53\pm0.00$ & $0.53\pm0.00$ 
 & $\textbf{0.50}\pm0.00$ & $\textbf{0.50}\pm0.00$ & $0.51\pm0.00$ & $0.52\pm0.00$ & $0.53\pm0.00$  
 & $\textbf{0.50}\pm0.00$ & $\textbf{0.50}\pm0.00$ & $\textbf{0.50}\pm0.00$ & $\textbf{0.50}\pm0.00$ & $0.53\pm0.01$ \\
 & \rgp & $\textbf{0.50}\pm0.00$ & $\textbf{0.50}\pm0.00$ & $\textbf{0.50}\pm0.00$ & $\textbf{0.50}\pm0.00$ & $\textbf{0.50}\pm0.00$ 
 & $\textbf{0.50}\pm0.00$ & $\textbf{0.50}\pm0.00$ & $\textbf{0.50}\pm0.00$ & $\textbf{0.50}\pm0.00$ & $\textbf{0.50}\pm0.00$ 
 & $\textbf{0.50}\pm0.00$ & $\textbf{0.50}\pm0.00$ & $0.51\pm0.00$ & $\textbf{0.50}\pm0.00$ & $\textbf{0.50}\pm0.00$ \\
 & \gep & $\textbf{0.50}\pm0.00$ & $\textbf{0.50}\pm0.00$ & $\textbf{0.50}\pm0.00$ & $0.52\pm0.00$ & $0.57\pm0.00$ & $0.51\pm0.00$ & $\textbf{0.50}\pm0.00$ & $\textbf{0.50}\pm0.00$ & $0.53\pm0.00$ & $0.54\pm0.00$ & - & - & - & - & - \\
 & \alloc & $\textbf{0.50}\pm0.00$ & $\textbf{0.50}\pm0.00$ & $0.51\pm0.00$ & $0.53\pm0.00$ & $0.53\pm0.01$ 
 & $\textbf{0.50}\pm0.00$ & $\textbf{0.50}\pm0.00$ & $0.51\pm0.00$ & $0.52\pm0.00$ & $0.53\pm0.01$ 
 & $\textbf{0.50}\pm0.00$ & $\textbf{0.50}\pm0.00$ & $\textbf{0.50}\pm0.00$ & $0.51\pm0.00$ & $0.54\pm0.00$ \\
 & \adpclip & $\textbf{0.50}\pm0.00$ & $\textbf{0.50}\pm0.00$ & $\textbf{0.50}\pm0.00$ & $0.53\pm0.00$ & $0.56\pm0.00$ 
 & $0.51\pm0.01$ & $\textbf{0.50}\pm0.00$ & $0.51\pm0.00$ & $0.52\pm0.00$ & $0.54\pm0.00$ 
 & $\textbf{0.50}\pm0.00$ & $\textbf{0.50}\pm0.00$ & $\textbf{0.50}\pm0.00$ & $\textbf{0.50}\pm0.00$ & $0.53\pm0.00$ \\
 & \pate & $\textbf{0.50}\pm0.00$ & $0.52\pm0.01$ & $0.51\pm0.01$ & $\textbf{0.50}\pm0.00$ & $0.51\pm0.00$ 
 & $0.51\pm0.01$ & $\textbf{0.50}\pm0.00$ & $\textbf{0.50}\pm0.00$ & $\textbf{0.50}\pm0.00$ & $\textbf{0.50}\pm0.01$ 
 & $0.51\pm0.01$ & $0.51\pm0.01$ & $\textbf{0.50}\pm0.01$ & $\textbf{0.50}\pm0.00$ & $\textbf{0.50}\pm0.00$ \\
 & \knn & $\textbf{0.50}\pm0.00$ & $\textbf{0.50}\pm0.00$ & $\textbf{0.50}\pm0.00$ & $\textbf{0.50}\pm0.00$ & $0.51\pm0.01$ 
 & $\textbf{0.50}\pm0.00$ & $\textbf{0.50}\pm0.00$ & $\textbf{0.50}\pm0.01$ & $\textbf{0.50}\pm0.00$ & $0.51\pm0.01$ 
 & $\textbf{0.50}\pm0.00$ & $\textbf{0.50}\pm0.00$ & $0.51\pm0.01$ & $\textbf{0.50}\pm0.00$ & $\textbf{0.50}\pm0.00$ \\
\bottomrule
\end{tabular}

  }
  \vspace{-0.3cm}
\end{table*}

\begin{figure*}[!t]
    \centering
    \begin{subfigure}[t]{\textwidth}
    \centering
        \includegraphics[width=0.9\linewidth]{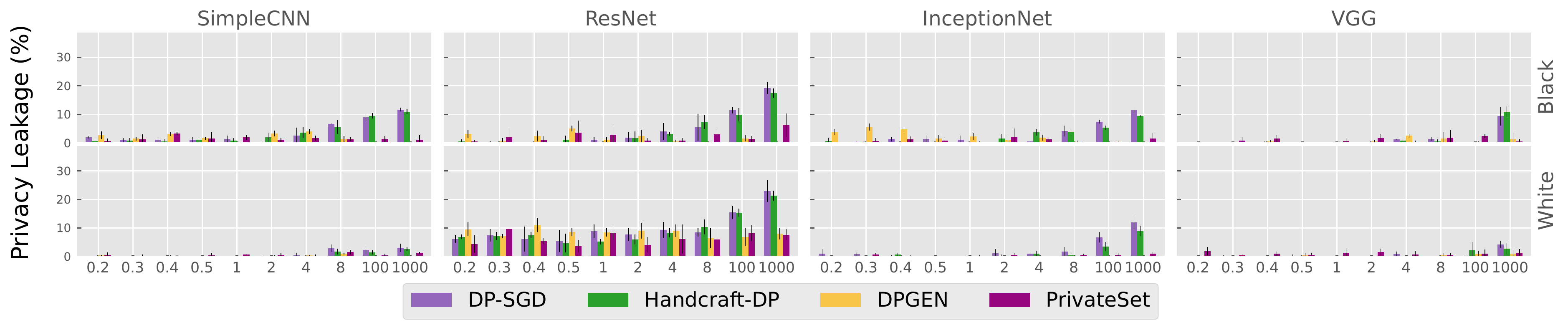}
        \vspace{-0.2cm}
        \caption{Data Preparation}
        \label{fig:multi_mia_data_preparation}
    \end{subfigure}
    \begin{subfigure}[t]{\textwidth}
    \centering
        \includegraphics[width=0.9\linewidth]{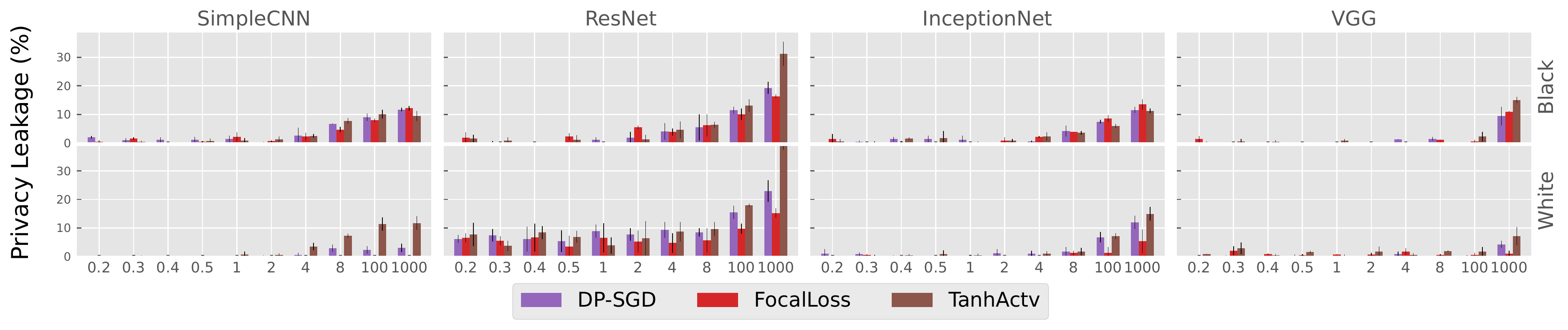}
        \vspace{-0.2cm}
        \caption{Model Design}
        \label{fig:multi_mia_model_design}
    \end{subfigure}
    \begin{subfigure}[t]{\textwidth}
    \centering
        \includegraphics[width=0.9\linewidth]{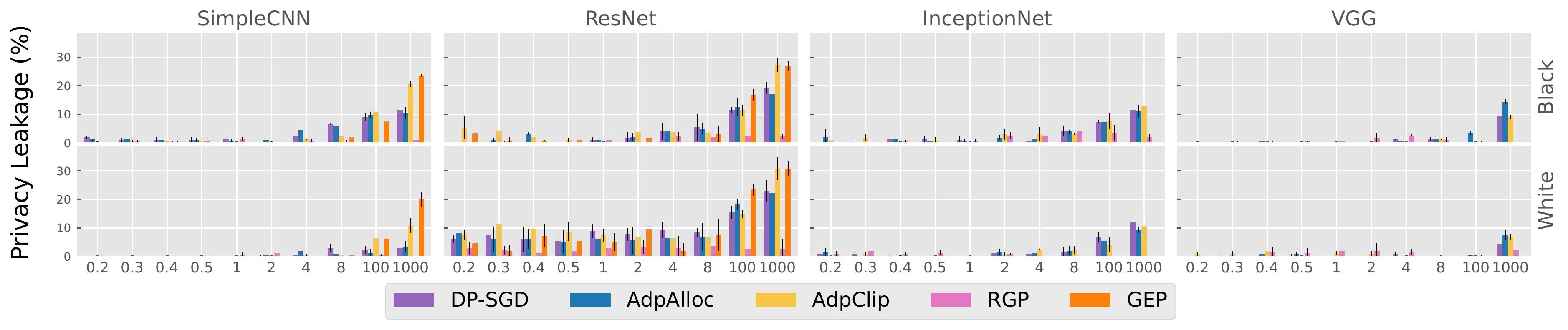}
        \vspace{-0.2cm}
        \caption{Model Training}
        \label{fig:multi_mia_model_train}
    \end{subfigure}
    \begin{subfigure}[t]{\textwidth}
    \centering
        \includegraphics[width=0.9\linewidth]{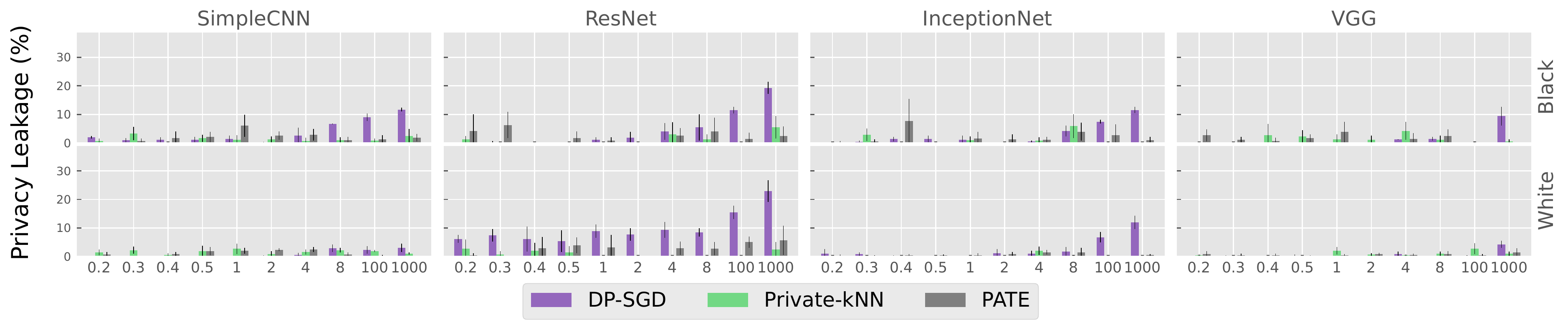}
        \vspace{-0.2cm}
        \caption{Model Ensemble}
        \label{fig:multi_mia_model_ensemble}
    \end{subfigure}
    \vspace{-0.4cm}
    \caption{Privacy leakage (under MIA) of DPML algorithms in four categories when given different privacy budgets.}
\label{fig:multi_mia}
\vspace{-0.3cm}
\end{figure*}

We report the tailored AUC of the black-box MIAs on CIFAR-10 in \autoref{tab:mia}, and put the results of other datasets and more settings into \autoref{apx:additional-results} for having the similar trends
(\autoref{tab:mia-black-apx} for more datasets under black-box MIAs, \autoref{tab:mia-white} for white-box MIAs).
Note that the tailored AUC of attacking the non-private model on the MNIST dataset is already very close to 0.5, so we omit the results on MNIST in this section. 
Generally, all algorithms' tailored AUC is around 0.5, which means a strong defense against the MIA compared to the baseline results~\autoref{tab:baseline}.

\autoref{fig:multi_mia} illustrates the privacy leakage of models trained by algorithms in a per-category manner. 
Compared to vanilla DP-SGD, the modification of \rgp and \loss change the feature of confidence vectors, resulting in training and testing data having a different distribution for the attack model.
Thus, \rgp and \loss have a remarkable advantage over black-box and white-box attacks in general.
Refer to \autoref{fig:multi_mia_model_ensemble} and \autoref{fig:multi_mia_data_preparation}.
We observe that \pate, \knn, \dpgen, and \privateset remain nearly free of privacy leakage.
It is because the target models do not access private data.
\pate and \knn use the knowledge transferred from teacher ensemble, and \dpgen and \privateset only access generated data.

\begin{table}[!t]
  \centering
  \caption{Impact of per-sample clipping on model utility and defense to attacks.
  The table reports the accuracy and the AUC of models on CIFAR-10 with different privacy guarantees. 
  Inf indicates normal SGD; 
  Inf (Clip) denotes normal SGD with per-sample clipping.}
  \vspace{-0.3cm}
  \label{tab:mia-clip}
  \resizebox{0.8\linewidth}{!}{
    \begin{tabular}{ccccccc}
    \toprule
     &  & 8 & 100 & 1000 & Inf(clip) & Inf \\
    \midrule
    \multirow[c]{2}{*}{\textbf{SimpleCNN}} & ACC (\%) & 58.20 & 60.66 & 60.44 & 57.96 & 69.22 \\
     & AUC & 0.52 & 0.52 & 0.53 & 0.52 & 0.78 \\
     \midrule
    \multirow[c]{2}{*}{\textbf{ResNet}} & ACC (\%) & 53.80 & 61.50 & 65.90 & 57.42 & 69.70 \\
     & AUC & 0.51 & 0.52 & 0.53 & 0.54 & 0.65 \\
    \midrule
    \multirow[c]{2}{*}{\textbf{InceptionNet}} & ACC (\%) & 58.00 & 64.60 & 69.40 & 72.80 & 83.68 \\
     & AUC & 0.51 & 0.51 & 0.52 & 0.58 & 0.71 \\
    \midrule
    \multirow[c]{2}{*}{\textbf{VGG}} & ACC (\%) & 10.36 & 10.02 & 64.66 & 67.72 & 71.36 \\
     & AUC & 0.50 & 0.50 & 0.52 & 0.59 & 0.78 \\
    \bottomrule
\end{tabular}

  }
  \vspace{-0.4cm}
\end{table}

\mypara{Role of Sensitivity-bounding Techniques}
To explore the role of sensitivity-bounding techniques in defending MIAs, we conduct attacks on a model trained with normal SGD and per-sample clipping to explore the impact of per-sample clipping on the defense. 
The results are shown in \autoref{tab:mia-clip}.

We observe that the per-sample clipping has a strong defense ability against MIAs with acceptable accuracy degradation compared to the non-private model.
Moreover, the defensive effects and accuracy degradation are model dependent.
For example, \textit{Inf(clip)} performs comparably to $\epsilon=8$ on \simple, but when applied to other models, the performance is worse than when $ \epsilon=1000$.

We suspect the reason why the per-sample clipping technique can defend against MIAs is that it reduces the overfitting of the model. 
During the training process, applying gradient descent without clipping guides the model to the direction that overfits the training samples; 
while clipping the gradient makes the model move more conservatively and less overfit to the training samples.
Note that the models trained by SGD with per-sample clipping have a defense ability against MIAs but do not satisfy the DP guarantee.

\subsection{The Role of the Architecture}
\label{subsec:model-impact}

\begin{figure*}[!t]
    \centering
    \begin{subfigure}[t]{\textwidth}
    \centering
        \includegraphics[width=0.9\linewidth]{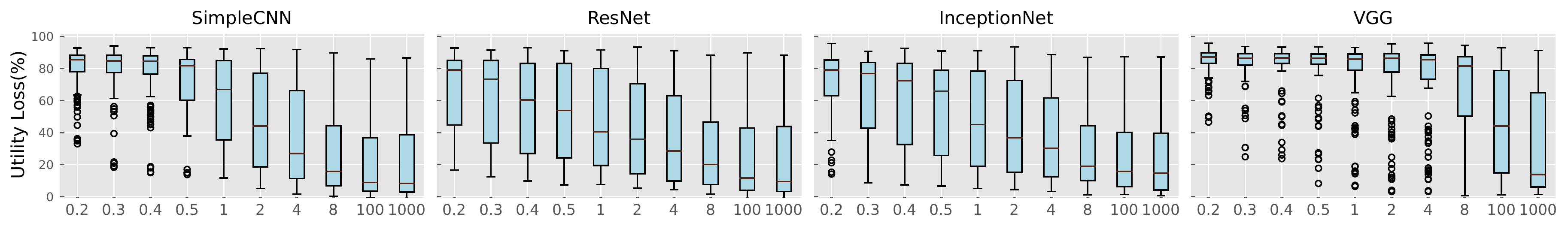}
        \vspace{-0.2cm}
        \caption{Utility Loss}
        \label{subfig:utility_netbox}
    \end{subfigure}
    \begin{subfigure}[t]{\textwidth}
    \centering
        \includegraphics[width=0.9\linewidth]{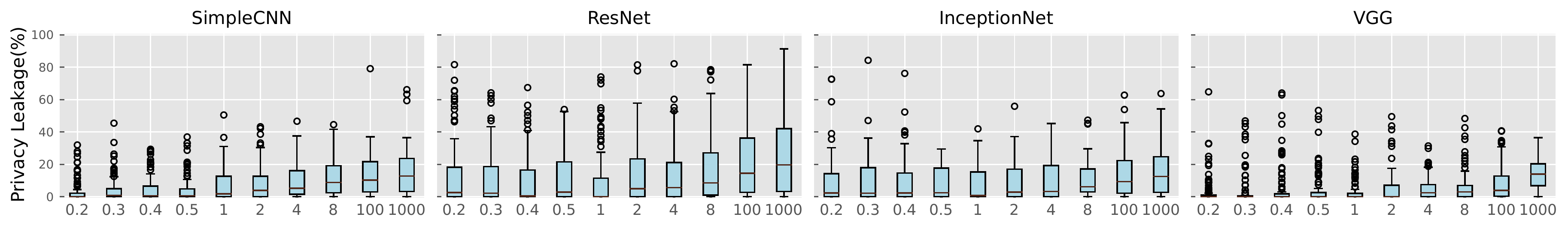}
        \vspace{-0.2cm}
        \caption{Privacy Leakage}
        \label{subfig:mia_netbox}
    \end{subfigure}
    \vspace{-0.4cm}
    \caption{Boxplot of utility loss and privacy leakage on all DPML algorithms with various privacy budgets and four network architectures .}
\label{fig:netbox}
\end{figure*}

\mypara{Architecture Complexity}
According to baseline accuracy in \autoref{tab:baseline}, the model's performance can be ordered as \inception $>$ \vgg $\approx$ \resnet $>$ \simple.

\mypara{Architecture versus Utility Loss}
To figure out the impact of model architecture on algorithm performance, we illustrate the boxplot for the utility loss overall algorithms, network, and dataset jointly vary with the privacy budget as \autoref{subfig:utility_netbox}.

We observe that the utility loss is similar for \resnet and \inception across different privacy budgets.
When the privacy budget is small ($\epsilon \leq 1$), the performance of \simple and \vgg is worse than that of \resnet and \inception.
As the noise amount becomes smaller ($\epsilon > 1$), the performance gap between \simple, \resnet, and \inception narrows.
The performance of \vgg, the largest model in our assessment, is still poor unless perturbed noise is negligible ($\epsilon \geq 100 $), while the privacy protection provided by DP is also meaningless.
Further, we explored the test accuracy of \resnet with different numbers of parameters under different privacy budgets.
Due to space limitaions, detailed results can be viewed at ~\autoref{fig:acc-params} in \autoref{apx:additional-results}.
Generally, the smaller the privacy budget and the more model parameters, the worse the model accuracy when training with vanilla DP-SGD.

\mypara{Architecture versus Privacy Leakage}
We also present a boxplot for the privacy leakage of all algorithms on different network architectures across privacy budgets as \autoref{subfig:mia_netbox}.
We observe no strong correlation between privacy leakage and model architecture.
\vgg has the lowest privacy leakage because many algorithms fail to converge on \vgg, leading to the following attack failure.

\subsection{The Role of the Datasets}
\label{subsec:dataset-impact}

\begin{figure*}[!t]
    \centering
    \begin{subfigure}[t]{\textwidth}
    \centering
        \includegraphics[width=0.9\linewidth]{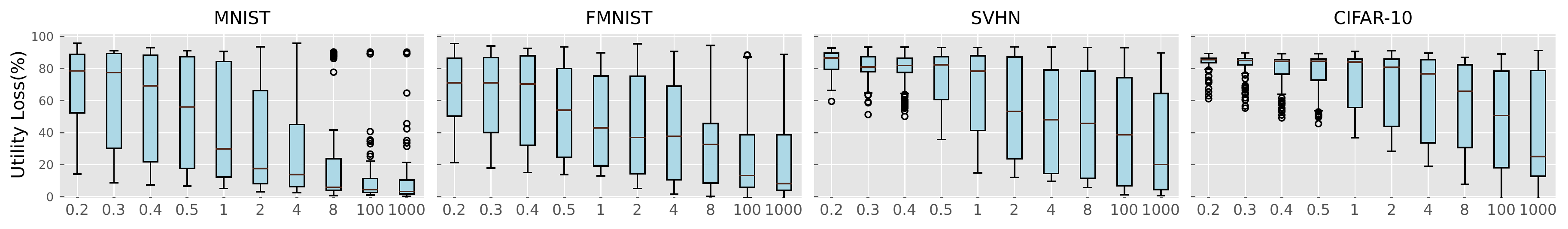}
        \vspace{-0.2cm}
        \caption{Utility Loss}
        \label{subfig:utility_databox}
    \end{subfigure}
    \begin{subfigure}[t]{\textwidth}
    \centering
        \includegraphics[width=0.9\linewidth]{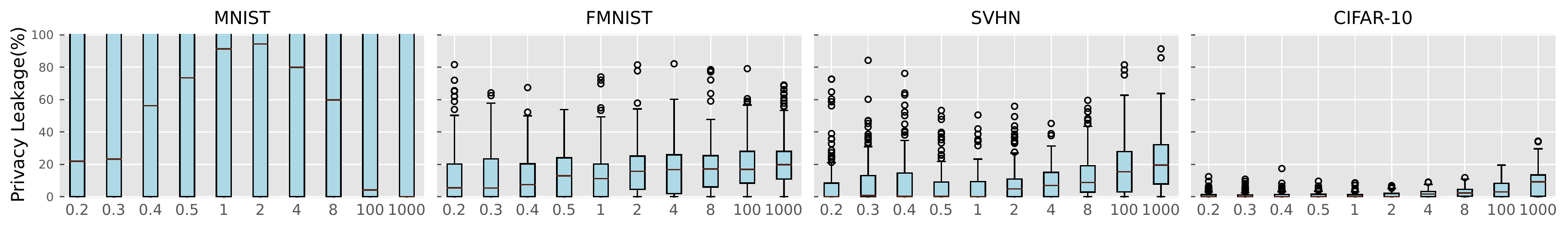}
        \vspace{-0.2cm}
        \caption{Privacy Leakage.
        The tailored AUCs of MIAs on MNIST is around 0.5, whether with or without DP, which leads to privacy leakage close to 100\%.}
        \label{subfig:mia_databox}
    \end{subfigure}
    \vspace{-0.3cm}
    \caption{Boxplot of utility loss and privacy leakage on all DPML algorithms with various privacy budgets for different datasets.}
\label{fig:databox}
\vspace{-0.4cm}
\end{figure*}

\mypara{Dataset Complexity}
As mentioned before, we resize all the samples in each dataset to $32\times32$ pixels.
MNIST and FMNIST are simpler than SVHN and CIFAR10 as they only contain gray-scale images.
When the number of channels is the same, MNIST and SVHN are easier than FMNIST and CIFAR10, respectively, because the contents of MNIST and FMNIST are digital numbers.
The accuracy of baseline models in \autoref{tab:baseline} shows the same conclusion.

\mypara{Dataset versus Utility Loss}
To explore the impact of the dataset on the DPML algorithm, we plot the relationship between dataset complexity and model utility loss in \autoref{subfig:utility_databox}.

As shown in the plots, the algorithm's performance on these datasets is correlated with the dataset complexity, with worse performance on the harder dataset.
Even with a very large privacy budget ($\epsilon = 100$), nearly half of the private models had a utility loss of more than 30\% on CIFAR10 compared to the non-private setting.

\mypara{Dataset versus Privacy Leakage}
We plot the relationship between dataset complexity and model privacy leakage in \autoref{subfig:mia_databox}.
We observe that more complex datasets lead to less privacy leakage.
One reason is that a complex dataset is harder to converge under private settings, and attackers cannot obtain enough information to infer.
Additionally, more complex datasets lead to better MIA performance~\cite{liu2021ml-doctor} under non-private settings, leading to a smaller privacy leakage value.
The tailored AUCs of MIAs on MNIST is around 0.5, whether with or without DP, which leads to privacy leakage close to 100\%.

\subsection{Comparison with Label DP}
Label Differential Privacy (Label DP) is a variant of DP where the data labels are considered sensitive and must be protected.
The definition of label differential privacy is:
\begin{definition}
(Label Differential Privacy). A randomized training algorithm M taking a dataset as input is said to be $(\epsilon, \delta)$-label differentially private, if for any two training datasets D and $\text{D}'$ that differ in the label of a single example, 
$$ Pr[\text{M}(D)\in S] \leq e^{\epsilon }  Pr[\text{M}(D') \in S] + \delta .$$
\end{definition}
If $\delta = 0$, then M is said to be $\epsilon$-label differentially private ($\epsilon$-LabelDP).
Label DP and DP synthetic algorithms share similar paradigms but differ in  generating synthetic datasets by satisfying Label DP instead of standard DP. 
Our evaluation covers two state-of-the-art Label DP algorithms: \lpmst~\cite{ghazi2021deep} and \alibi~\cite{malek2021antipodes},  to explore the difference between Label DP and standard DP algorithms.
It is worth noting that the Label-DP satisfies bounded DP.
We convert the privacy budget for equivalence while comparing it with other algorithms, and the figure shows the privacy budget in unbounded DP (\eg $\langle \rgp, \resnet, \text{CIFAR-10},1000 \rangle$ and $\langle \lpmst, \resnet, \text{CIFAR-10},2000 \rangle$ share the same horizontal coordinate, 1000).
The concrete algorithm description can be found in \autoref{apx:labeldp-funcs}

\begin{figure*}[!t]
    \centering
    \begin{subfigure}[t]{\textwidth}
    \centering
        \includegraphics[width=0.9\linewidth]{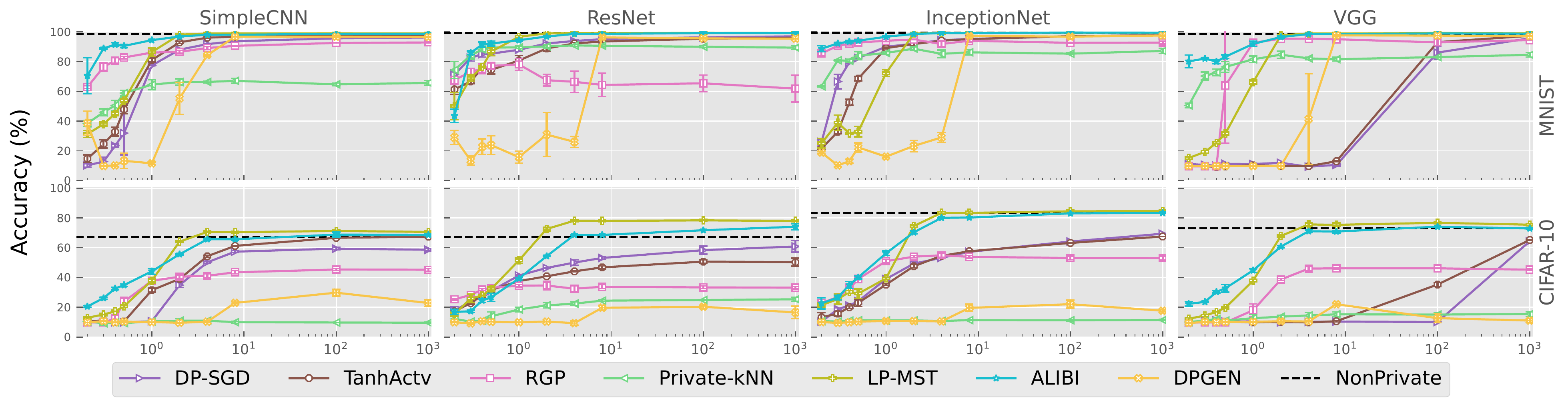}
        \vspace{-0.2cm}
        \caption{Accuracy}
        \label{fig:acc_label_cmp}
    \end{subfigure}
    \begin{subfigure}[t]{\textwidth}
    \centering
        \includegraphics[width=0.9\linewidth]{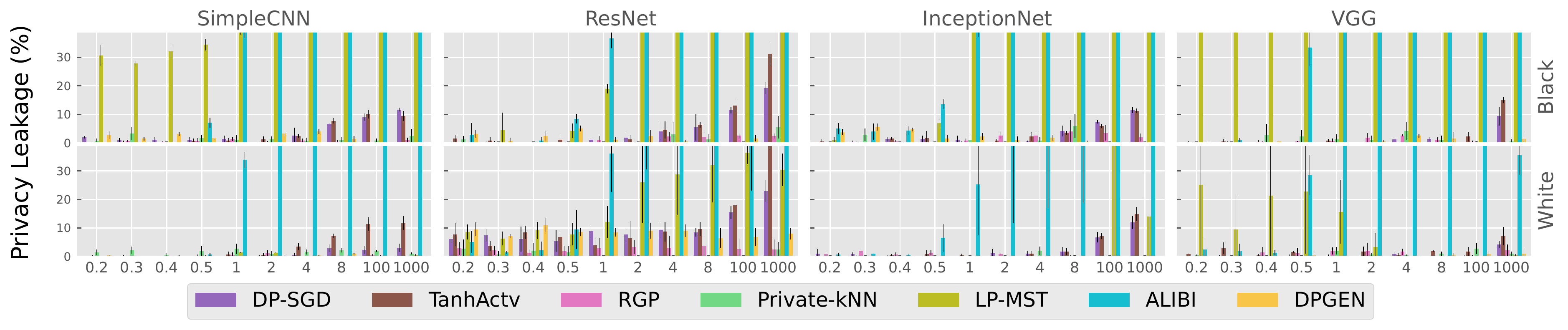}
        \vspace{-0.2cm}
        \caption{Privacy Leakage}
        \label{fig:mia_label_cmp}
    \end{subfigure}
    \vspace{-0.3cm}
    \caption{Comparison of Label DP algorithms (\lpmst, \alibi) and vanilla DP-SGD, \sigmoid, \rgp, \knn, and \dpgen under different privacy budgets.}
\label{fig:label_cmp}
\vspace{-0.5cm}
\end{figure*}

\autoref{fig:acc_label_cmp} illustrates the comparison of accuracy between Label DP algorithms and vanilla DP-SGD,
\sigmoid, \rgp, \knn, and \dpgen.
We notice that the accuracy of \lpmst and \alibi can approach or even exceed baseline when the privacy budget is not very large, \eg the accuracy of $\langle \lpmst, \resnet, \operatorname{CIFAR-10},4 \rangle $ is 71.82 larger than the baseline of 66.56. 
There are two reasons behind this.
One is that noise only affects labels. 
The training process gradually becomes the same as non-private training as the private budget increase.
The other is that the techniques used to mitigate the effects of wrong labels usually also improve the model's generalization, such as mixup~\cite{zhang2017mixup} used in \lpmst~\cite{ghazi2021deep}.

\autoref{fig:mia_label_cmp} illustrates the comparison of black-box MIA on Label DP algorithms and vanilla DP-SGD, \sigmoid, \rgp, \knn, and \dpgen with the metric of privacy leakage.
We observe that Label DP algorithms have higher privacy leakage than standard DP algorithms, which is natural for Label DP because of no protection provided to data.
\vspace*{-0.2cm}

\subsection{Takeaways} \label{subsec:summary}
In the following, we summarize important insights obtained from our measurements and provide some actionable advice to future DPML practitioners.
\begin{itemize}[topsep=1pt,itemsep=1pt,nolistsep]

    \item Different improvement techniques can affect the privacy-utility trade-offs of the algorithm from different perspectives.
    Concretely, parameter dimension reduction in the model training category improves the performance of DPML on large models but impairs utility when the privacy budget is large.
    Thus, \rgp is a good choice for those who want to provide a DP guarantee for large models.
    On the other hand, algorithms in the model ensemble category and DP synthetic algorithms can be used when stronger defense against MIAs is desired.
    However, more effort on manual data filtering for DP synthetic algorithms is needed for better utility.
    
    \item In general, the DPML algorithms provide an effective defense against practical MIAs in both black-box and white-box manner.
    The defense performance hardly decreases when the privacy budget increases.
    The reason is that sensitivity-bounding techniques such as gradient clipping play an important role in defense. 
    More specifically, improved algorithms that do not directly access private data are better at defending against attacks, such as algorithms in the model ensemble category and DP synthetic algorithms.
    In addition, improved algorithms that affect the attack features of MIAs can achieve additional defensive capabilities.
    For instance, the confidence vector distribution of \loss is different from that of shadow models, which causes \loss to be more robust to attacks.
    All algorithms that provide the standard DP guarantee can defend MIAs effectively.
    
    \item Some model architecture design choices for non-private ML models are ineffective for private ML models.
    More specifically, a large model scale degrades utility for most DPML algorithms.
    In addition, using Tanh and GroupNorm can reduce the utility loss on vanilla DP-SGD.
    However, we also find that using both Tanh and GroupNorm has a negative effect.
    What model architectures are suitable for DPML is still a research question to be explored.
    When applying DP to ML models, \resnet and \inception are preferred architectures to attempt.

    \item In general, learning data distribution from more complex datasets is more difficult than that from easier datasets for all DP algorithms.
    Compared with the non-private setting, applying DP makes it even more difficult to learn from complex datasets.
    Leveraging external datasets (\eg, pre-train on public dataset~\cite{abadi2016deep} and public data embedding~\cite{yu2021not}) can be helpful to improve the utility of the model on complex datasets.
    Therefore, designing DPML algorithms to better learn from complex datasets is an interesting future research direction.
    
    \item 
    Label DP algorithms achieve better model utility than standard DP algorithms, which is expected since label DP algorithms loosen the constraint on adjacent datasets.
    However, the defense effectiveness of label DP algorithms is worse than that of standard DP algorithms since they only protect the privacy of the label instead of the privacy of the training sample.
    Label DP should only be used when the label is sensitive, not the data itself, and there is no need to defend against MIAs.
\end{itemize}

\section{Discussion}
\label{sec:discussion}

In this section, we discuss several potential research directions to inspire interested readers to explore relavent domains.

\mypara{Emsembled DPML Algorithms}
As discussed in \autoref{subsec:tax-overview}, the improved DPML algorithms in different phases of our taxonomy are independent of each other;
thus, one interesting future work is to combine the improvements in different phases to achieve better performance.
Shamsabadi \etal~\cite{shamsabadi2021losing} take the first step and show that combining a handcrafted feature extractor\cite{tramer2020differentially} in the data preparation phase and optimal loss function in the model design phase can effectively improve the model utility.
It would be exciting to follow our taxonomy and combine algorithms at different phases to achieve even better performance.

\mypara{Extension to Other Domains}
Our current measurement primarily focuses on image classification tasks, it would be interesting to leverage \sysname to measure the performance of DPML algorithms in other domains, such as natural language processing (NLP) and graph neural networks (GNN).

\mypara{DPML Algorithms for Large Models}
With the development of deep learning, the model scale increases rapidly, especially in the NLP field.
For instance, the famous GPT-3 model contains 175B parameters~\cite{brown2020language}.
However, our measurements show that most of the current DPML algorithms suffer from low model utilities. 
Furthermore, DP-SGD-based algorithms require calculating per-sample clipping of the gradients, which significantly increases the training time and memory consumption.
Therefore, designing high-utility and efficient DPML algorithms for large models is of significant importance in the future.

\section{Related Work}

\mypara{Differential Privacy}
Differential privacy (DP)~\cite{dwork2008Differential,dwork2014algorithmic} is a widely used rigorous mathematical definition to formalize and measure privacy guarantees based on a parameter called \textit{privacy budget}.
It has been adopted for a number of data analysis tasks, such as synthetic dataset generation~\cite{YZDCCS23,WZWHBCZ23,ZWLHBHCZ21,DHZFCZG23}, marginal release~\cite{ZWLHC18}, range query~\cite{DZBLJCC21}, and stream data analysis~\cite{WCZSCLLJ21}.
Some studies propose integrating DP with traditional machine learning algorithms, such as naive Bayes and Linear Support Vector Machine (SVM)~\cite{vaidya2013differentially,chaudhuri2008privacy,chaudhuri2011differentially}.
Abadi \etal propose vanilla DP-SGD~\cite{abadi2016deep} as the first general DPML algorithm.
Recent studies try to mitigate DP's impairment on utility by proposing new algorithms~\cite{papernot2017semi,tramer2020differentially,yu2021not,zhu2020private} or relax DP definition for specific scenarios~\cite{machanavajjhala2008privacy,ghazi2021deep,dwork2010differential}.

\mypara{Membership Inference Attacks}
The adversary in MIAs aims to infer whether a given data sample is used to train the target model.
Currently, the MIA is one of the critical methods to assess the privacy risk of ML models~\cite{shokri2017membership,salem2018ml,nasr2019comprehensive,CZWBHZ21,HZSBLZ22,CZWBZ23}.
According to the accessibility to the target model, the MIA can be categorized into black-box and white-box attacks.
Shokri \etal~\cite{shokri2017membership} propose the first black-box MIA against ML models.
They propose to train multiple shadow models to simulate the behavior of the target model and use shadow models to generate the data used to train the attack model.
Salem \etal~\cite{salem2018ml} simplify their method by using one shadow dataset and one shadow model.
Nasr \etal~\cite{nasr2019comprehensive} first propose white-box MIAs, where the adversary knows the internal parameters of the target model.

\mypara{DPML Measurement}
Several DPML measurement studies concentrate on different perspectives~\cite{jayaraman2019evaluating,iyengar2019towards,zhao2020not,jarin2022dp}.
Jayaraman \etal~\cite{jayaraman2019evaluating} analyzed the difference of privacy leakage of relaxed variants of differential privacy.
They explore the difference in privacy leakage when using the same algorithm with different DP definitions.
Iyengar \etal~\cite{iyengar2019towards} evaluate several differentially private convex optimization algorithms.  
The work of Zhao \etal~\cite{zhao2020not} and Jarin \etal~\cite{jarin2022dp} analyze the performance of naive noise perturbation in different stages of the training pipeline.

ML-Doctor~\cite{liu2021ml-doctor} also investigates the defenses and attacks against ML models.
However, we have different objectives.
ML-Doctor aims to evaluate the effectiveness of different types of defenses against attacks. 
For DPML, they only evaluate the vanilla DP-SGD, and their only conclusion is that DP-SGD can defend against MIAs while failing for other attacks without considering the impact on model utility. 
On the other hand, \sysname conducts more fine-grained taxonomy and evaluation on different DPML algorithms and aims to evaluate the trade-off between model utility, privacy guarantee, and defense effectiveness. 
This can better facilitate future research on DPML.
As such, we obtained more insights on how to design proper DPML algorithms to trade off the above triangle, as stated in \autoref{subsec:summary}. 

\vspace{-0.2cm}

\section{Conclusion}
This paper establishes a taxonomy of improved DPML algorithms along the ML life cycle for four types: data preparation, model design, model training, and model ensemble.
Based on taxonomy, we propose the first holistic measurement of improved DPML algorithms' performance on utility and defense capability against MIAs on image classification tasks.
Our extensive measurement study covers \nalgs DPML algorithms, two attacks, four model architectures, four datasets, and various privacy budget configurations.
We also cover state-of-the-art label DP in the evaluation.

Among other things, we found that different improvement techniques can affect the privacy-utility trade-off of the algorithm from different perspectives.
We also show that DP can effectively defend against MIAs and sensitivity-bounding
techniques such as per-sample gradient clipping play an important role in defense.
Moreover, some model architecture design choices for non-private ML models are ineffective for private ML models.
In addition, label DP has less utility loss but is fragile to MIAs.

We implement a modular re-usable software, \sysname, which contains all algorithms and attacks.
\sysname enables sensitive data owners to deploy DPML algorithms and serves as a benchmark tool for researchers and practitioners.
Currently, while \sysname focuses on image classification models, we plan to extend other types of DP models, such as language models~\cite{yu2022differentially,li2021large}, graph neural networks~\cite{ZCBSZ22,CZWBHZ22,SHZCYBZS22}, and generative models~\cite{he2021fedgraphnn,torkzadehmahani2019dpcgan}.

\section*{Acknowledgement}
This work is supported in part by the National Natural Science Foundation of China (NSFC) under No. 62302441, the Funding for Postdoctoral Scientific Research Projects in Zhejiang Province (ZJ2022072), and ZJU – DAS-Security Joint Research Institute of Frontier Technologies, the Helmholtz Association within the project ``Trustworthy Federated Data Analytics'' (TFDA) (No. ZT-I-OO1 4), and CISPA-Stanford Center for Cybersecurity (FKZ:13N1S0762).

\bibliographystyle{unsrt}

\bibliography{ref_arxiv}

\appendix

\section{Hyperparameter Settings}
\label{apx:hyper-setting}
\autoref{tab:hyper-setting} reports the detailed hyperparameter settings.
Settings of \dpgen and \privateset are for classifier training.
We follow the author's setting for generated algorithms. 

\begin{table}[!h]
  \centering
  \caption{Detailed hyperparameter settings.
  Settings of \dpgen and \privateset are for classifier training.
  We follow the author's setting for generated algorithms. }
  \label{tab:hyper-setting}
  \resizebox{\linewidth}{!}{
    % Table generated by Excel2LaTeX from sheet 'Hyperparameter'

    \begin{tabular}{c|cccc}
    \toprule
          & \textbf{Learning Rate} & \textbf{Batch Size} & \textbf{Epoch} & \textbf{Additional} \\
    \hline
    \multirow{2}[2]{*}{\textbf{vanilla DP-SGD}} & \multirow{2}[2]{*}{0.01} & \multirow{2}[2]{*}{256} & \multicolumn{1}{c}{MNIST,FMNIST:60 } & \multirow{2}[2]{*}{} \\
          &       &       & \multicolumn{1}{c}{SVHN,CIFAR-10:90} &  \\
    \hline
    \multirow{2}[2]{*}{\sigmoid} & \multirow{2}[2]{*}{0.01} & \multirow{2}[2]{*}{256} & \multicolumn{1}{c}{MNIST,FMNIST:60 } & \multirow{2}[2]{*}{} \\
          &       &       & \multicolumn{1}{c}{SVHN,CIFAR-10:90} &  \\
    \hline
    \multirow{2}[2]{*}{\alloc} & \multirow{2}[2]{*}{0.01} & \multirow{2}[2]{*}{256} & \multicolumn{1}{c}{MNIST,FMNIST:60 } & ExpDecay \\
          &       &       & \multicolumn{1}{c}{SVHN,CIFAR-10:90} & \multicolumn{1}{c}{k=0.01} \\
    \hline
    \multirow{4}[2]{*}{\adpclip} & \multirow{4}[2]{*}{0.01} & \multirow{4}[2]{*}{256} & \multicolumn{1}{c}{\multirow{2}[1]{*}{MNIST,FMNIST:60 }} & \multicolumn{1}{c}{target\_unclipped\_quantile=0.7} \\
          &       &       &       & \multicolumn{1}{c}{clipbound\_learning\_rate=0.1} \\
          &       &       & \multicolumn{1}{c}{\multirow{2}[1]{*}{SVHN,CIFAR-10:90}} & \multicolumn{1}{c}{max\_clipbound=10} \\
          &       &       &       & \multicolumn{1}{c}{min\_clipbound=0.05} \\
    \hline
    \multirow{2}[2]{*}{\loss} & \multirow{2}[2]{*}{0.01} & \multirow{2}[2]{*}{256} & \multicolumn{1}{c}{MNIST,FMNIST:60 } & \multirow{2}[2]{*}{weight\_decay=1e-4} \\
          &       &       & \multicolumn{1}{c}{SVHN,CIFAR-10:90} &  \\
    \hline
    \multirow{2}[2]{*}{\textbf{Handcrafted}} & \multirow{2}[2]{*}{0.01} & \multirow{2}[2]{*}{256} & \multicolumn{1}{c}{MNIST,FMNIST:60 } & \multirow{2}[2]{*}{} \\
          &       &       & \multicolumn{1}{c}{SVHN,CIFAR-10:90} &  \\
    \hline
    \multirow{4}[2]{*}{\gep} & \multirow{4}[2]{*}{0.1} & \multirow{4}[2]{*}{256} & \multicolumn{1}{c}{\multirow{2}[1]{*}{MNIST,FMNIST:60 }} &  num\_groups=3 \\
          &       &       &       & num\_bases=1000 \\
          &       &       & \multicolumn{1}{c}{\multirow{2}[1]{*}{SVHN,CIFAR-10:90}} & weight\_decay=2e-4 \\
          &       &       &       & \multicolumn{1}{c}{aux\_data\_size=2000} \\
    \hline
    \multirow{3}[2]{*}{\rgp} & \multirow{3}[2]{*}{0.1} & \multirow{3}[2]{*}{256} & \multicolumn{1}{c}{\multirow{2}[1]{*}{MNIST,FMNIST:60 }} & \multicolumn{1}{c}{width=1} \\
          &       &       &       & \multicolumn{1}{c}{rank=16} \\
          &       &       & \multicolumn{1}{c}{SVHN,CIFAR-10:90} & \multicolumn{1}{c}{weight\_decay=1e-4} \\
    \hline
    \textbf{PATE} & 0.001 & 200   & 500   & n\_teacher=100 \\
    \hline
    \multirow{2}[2]{*}{\knn} & \multirow{2}[2]{*}{0.01} & \multirow{2}[2]{*}{512} & \multirow{2}[2]{*}{500} & iteration=2 \\
          &       &       &       & \multicolumn{1}{c}{ sample\_prob=0.15} \\
    \hline
    \dpgen & 0.01  & 1024  & 100   &  \\
    \hline
    \privateset & 0.01  & 10    & 300   & samples\_per\_class=10 \\
    \hline
    \lpmst & 0.01 & 256 & 200 & stage=2 \\
    \hline
    \multirow{2}[2]{*}{\alibi} & \multirow{2}[2]{*}{0.01} & \multirow{2}[2]{*}{256} & \multicolumn{1}{c}{MNIST,FMNIST:60 } & \multirow{2}[2]{*}{post\_process=mapwithprior} \\
          &       &       & \multicolumn{1}{c}{SVHN,CIFAR-10:90} &  \\
    \bottomrule
    \end{tabular}%

  }
\end{table}

\section{Dataset Description}
\label{apx:dataset}
\begin{itemize}
    \setlength{\itemsep}{1pt}
    \setlength{\parsep}{1pt}
    \setlength{\parskip}{1pt}
    \item \textbf{MNIST} comprises 60000 training samples and 10000 test samples. 
    Each sample is a 28x28 pixel gray handwritten numeral picture.
    \item \textbf{Fashion-MNIST (FMNIST)} has the same size, format, and train\\/test set division as the MNIST. 
    It covers front images of products from 10 different clothing categories.
    It has 60000 training samples and 10000 test samples.
    \item \textbf{CIFAR-10} consists of 10 categories of real-world objects of color images, and the size of each picture is 32$\times$32. 
    There are 50000 training images and 10000 test images in the dataset.
    \item \textbf{Street View House Number (SVHN)} is the house number extracted from the Google Street view image. 
    It can be seen as a colorful and more realistic version of MNIST.
    It comprises 73257 training samples and 26032 test samples, which are 32$\times$32 RGB images. 
    We trim the testset size to 10000 while keeping distribution consistent with the original testset.
\end{itemize}

\begin{enumerate}
    \setlength{\itemsep}{1pt}
    \setlength{\parsep}{1pt}
    \setlength{\parskip}{1pt}
    
    \item \textbf{Target Training Dataset} is regarded as private data and member samples while evaluating the performance of MIAs.
    \item \textbf{Target Testing Dataset} is used to evaluate the utility performance of the model.
    It is also used to evaluate the performance of MIAs as non-member samples.
    \item \textbf{Shadow Training Dataset} is used to train shadow models as auxiliary datasets of adversaries and then generate training data as members for attack models.
    \item \textbf{Shadow Testing Dataset} is used to generate training data as non-members for attack models.
\end{enumerate}

\section{Details of Label DP Algorithms}
\label{apx:labeldp-funcs}

\begin{itemize}
    \item \lpmst
    Ghazi \etal~\cite{ghazi2021deep} introduced \textit{RRWithPrior}, a Randomized Response (RR)~\cite{warner1965randomized} based algorithm, to perform label perturbation, to determine whether the label of each data sample is obtained by the RR mechanism or randomly generated.
    To mitigate the effects of mislabeling, \lpmst leverages a multi-stage training strategy.
    
    \item \alibi
    Malek \etal~\cite{malek2021antipodes} provide label DP guarantee by applying additive Laplace noise to a one-hot encoded label.
    To mitigate the effects of the perturbed label, they apply Bayesian post-processing to the output of the Laplace mechanism to mitigate the effect of mislabeling.
\end{itemize}

\section{Additional Results}
\label{apx:additional-results}

\mypara{Results of \wavelet on MLP}
\autoref{tab:acc-handcraft-cmp} shows the results comparison of \wavelet among 5 model architectures on CIFAR-10.

\begin{table}[!t]
    \centering
    \caption{Test accuracy of 5 model architectures on CIFAR-10 when given various privacy budgets.}
    \label{tab:acc-handcraft-cmp}
    \resizebox{0.85\linewidth}{!}{
        \begin{tabular}{cccccccc}
        \toprule
         & 0.2 & 0.4 & 1 & 4 & 100 & 1000 & Inf \\
        \midrule
        MLP & \textbf{27.30} & \textbf{32.50} & 38.98 & 46.70 & 54.02 & 57.48 & 57.32 \\
        SimpleCNN & 9.66 & 9.98 & 10.24 & 51.12 & 57.06 & 55.66 & 65.18 \\
        ResNet & 14.78 & 28.04 & \textbf{42.02} & 49.36 & \textbf{60.80} & 64.58 & 72.16 \\
        InceptionNet & 15.08 & 22.14 & 39.54 & \textbf{52.22} & 60.78 & \textbf{66.42} & \textbf{81.58} \\
        VGG & 9.70 & 10.16 & 10.32 & 10.22 & 30.08 & 60.90 & 72.26 \\
        \bottomrule
        \end{tabular}
    }
\end{table}

\begin{figure}[h]
    \centering
    \setlength{\abovecaptionskip}{0.2cm}
    \includegraphics[width=0.85\linewidth]{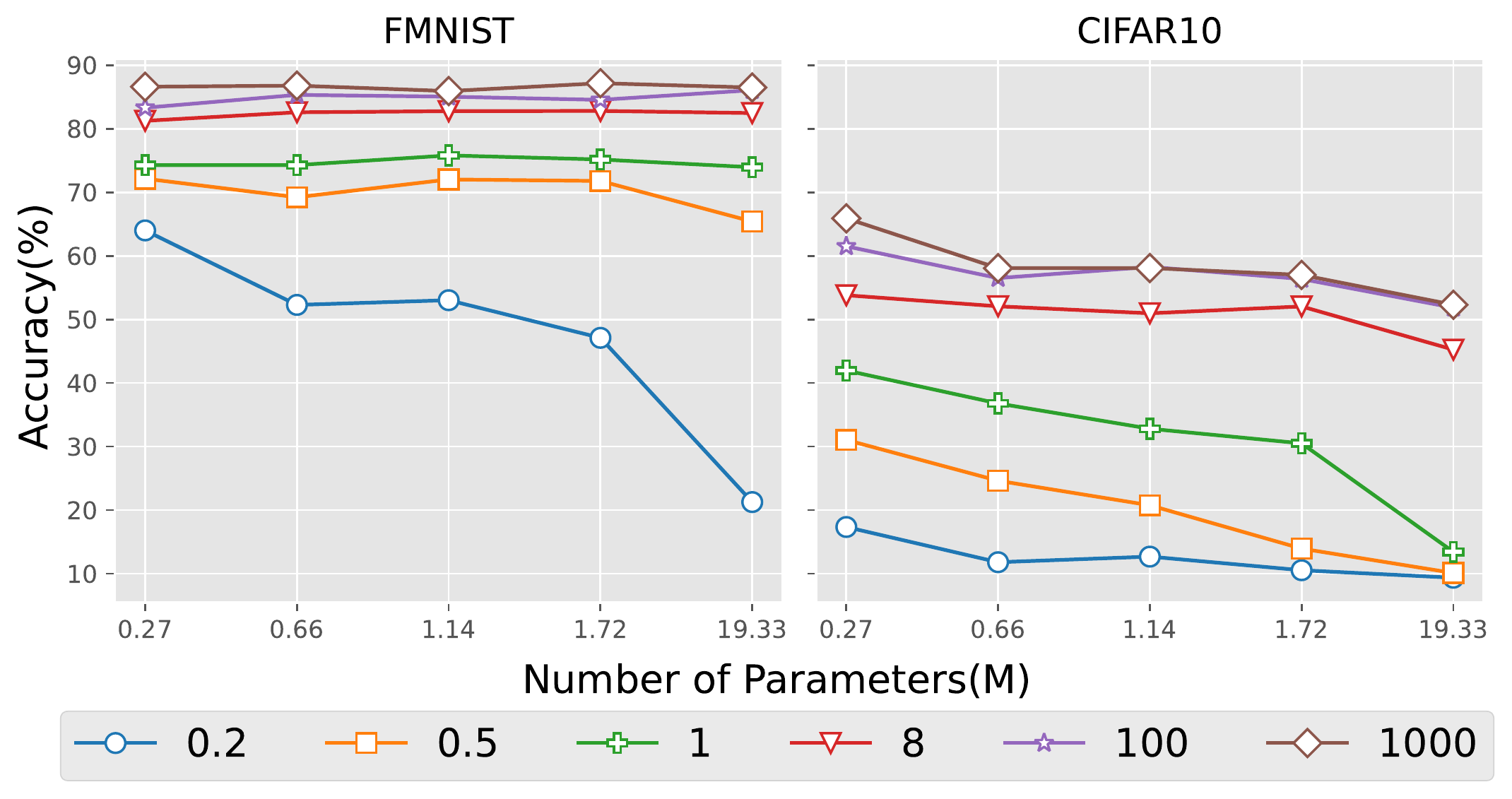}
    \caption{Accuracy of \resnet with the different number of parameters trained by vanilla DP-SGD under different $\epsilon$ levels.}
    \label{fig:acc-params}
    \vspace{-0.45cm}
\end{figure}

\mypara{Impact of Model Parameter Amounts on Accuracy}
\autoref{fig:acc-params} shows the test accuracy of \resnet with the different numbers of parameters trained by vanilla DP-SGD under different privacy budgets.

\mypara{Accuracy of \rgp without DP}
To figure out the effect of reparametrization in a non-private setting, \autoref{tab:rgp_nodp} reports the accuracy comparison between \rgp without DP and vanilla DP-SGD.

\begin{table}[h]
  \centering
  \caption{Accuracy of \rgp (w/o DP) and vanilla SGD. Other settings keep the same as \autoref{tab:acc}.}
    \label{tab:rgp_nodp}
  \resizebox{\linewidth}{!}{
    \begin{tabular}{c|cc|cc|cc|cc}
    \toprule
     & \multicolumn{2}{c|}{\textbf{SimpleCNN}} & \multicolumn{2}{c|}{\textbf{ResNet}} & \multicolumn{2}{c|}{\textbf{InceptionNet}} & \multicolumn{2}{c}{\textbf{VGG}} \\
     & RGP(w/o DP) & SGD & RGP(w/o DP) & SGD & RGP(w/o DP) & SGD & RGP(w/o DP) & SGD \\
    \midrule
    MNIST & 95.50 & \textbf{98.42} & 97.78 & \textbf{99.24} & 99.04 & \textbf{99.18} & 98.56 & \textbf{98.68} \\
    FMNIST & 85.70 & \textbf{88.04} & 86.62 & \textbf{88.60} & 90.76 & \textbf{91.70} & 88.72 & \textbf{90.48} \\
    SVHN & 73.70 & \textbf{87.69} & 90.58 & \textbf{93.84} & 93.09 & \textbf{94.90} & 88.14 & \textbf{89.77} \\
    CIFAR-10 & 50.40 & \textbf{69.22} & 59.94 & \textbf{68.16} & 75.74 & \textbf{83.68} & 68.32 & \textbf{71.36} \\
    \bottomrule
    \end{tabular}
    }
\end{table}

\begin{figure*}[!t]
    \centering
    \begin{subfigure}[t]{\textwidth}
    \centering
        \includegraphics[width=0.9\linewidth]{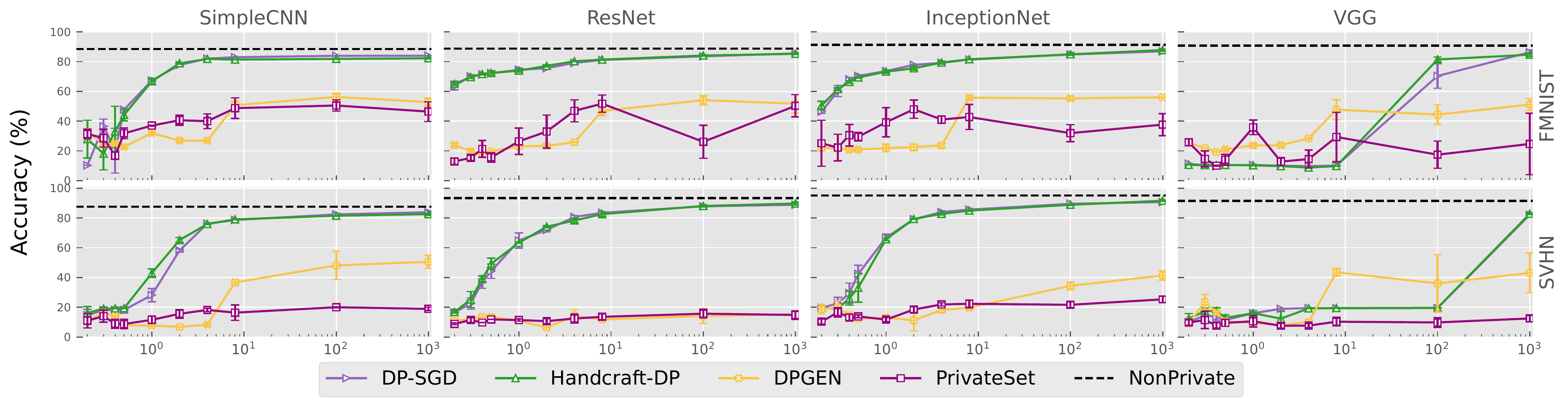}
        \caption{Data Preparation}
        \label{fig:acc_data_preparation_apx}
    \end{subfigure}
    \begin{subfigure}[t]{\textwidth}
    \centering
        \includegraphics[width=0.9\linewidth]{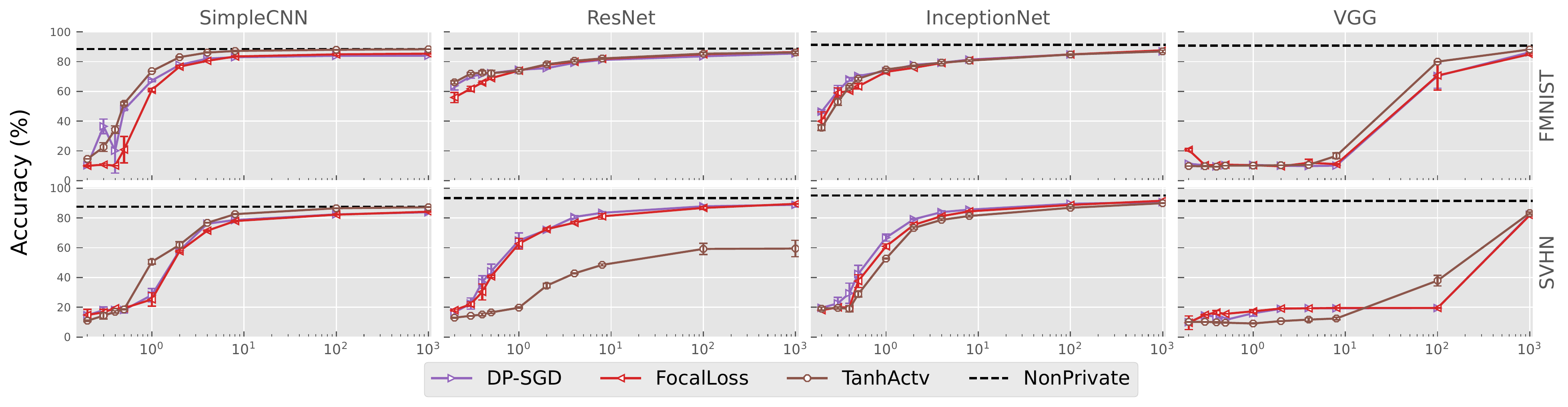}
        \caption{Model Design}
        \label{fig:acc_model_design_apx}
    \end{subfigure}
    \begin{subfigure}[t]{\textwidth}
    \centering
        \includegraphics[width=0.9\linewidth]{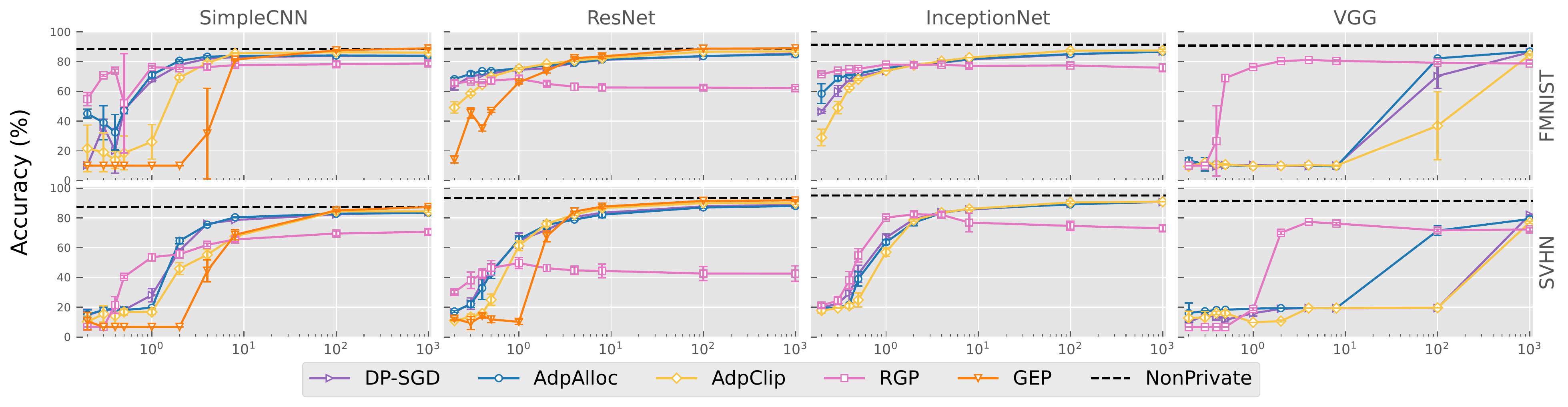}
        \caption{Model Training}
        \label{fig:acc_model_train_apx}
    \end{subfigure}
    \begin{subfigure}[t]{\textwidth}
    \centering
        \includegraphics[width=0.9\linewidth]{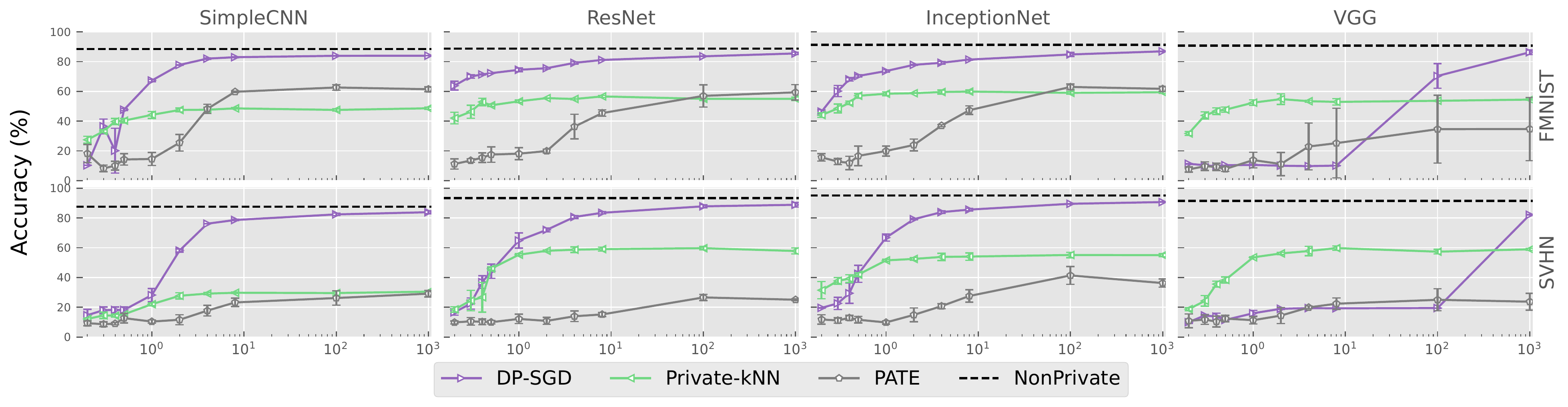}
        \caption{Model Ensemble}
        \label{fig:acc_model_ensemble_apx}
    \end{subfigure}
    \vspace{-0.4cm}
    \caption{Accuracy comparison on FMNIST and SVHN. As a supplementary to \autoref{fig:acc}. }
\label{fig:acc_apx}
\end{figure*}

\mypara{Accuracy on FMNIST and SVHN}
\autoref{fig:acc_apx} shows the accuracy of DPML algorithms on FMNIST and SVHN in terms of categories.
As a supplementary to \autoref{fig:acc}.

\mypara{Utility loss on FMNIST and SVHN}
\autoref{tab:acc-apx} shows the results of utility loss on FMNIST and SVHN as the supplement of \autoref{tab:acc}.

\begin{table*}[!t]
    \centering
    \caption{Overview of algorithms' utility loss on different model architectures, datasets, and privacy budget.  
    For each privacy budget, we bold the value with the best performance (with the \textbf{smallest} value of utility loss).
    Supplement to \autoref{tab:acc}.}
    \vspace{-0.3cm}
    \label{tab:acc-apx}
    \resizebox{\textwidth}{!}
    {
        \begin{tabular}{c|l|ccccc|ccccc|ccccc|ccccc}
\toprule
\multirow{2}{*}{} & \multirow{2}{*}{} & \multicolumn{5}{c|}{SimpleCNN} & \multicolumn{5}{c|}{ResNet} & \multicolumn{5}{c|}{InceptionNet} & \multicolumn{5}{c}{VGG} \\
 &  & 0.2 & 1 & 4 & 100 & 1000 & 0.2 & 1 & 4 & 100 & 1000 & 0.2 & 1 & 4 & 100 & 1000 & 0.2 & 1 & 4 & 100 & 1000 \\
\midrule
\multirow[c]{12}{*}{\rotatebox{90}{FMNIST}} & \dpgen & $66.83\pm1.90$ & $67.80\pm0.84$ & $72.94\pm0.77$ & $43.30\pm2.60$ & $46.79\pm2.73$ & $75.84\pm1.15$ & $76.69\pm2.18$ & $73.98\pm0.31$ & $45.47\pm3.09$ & $47.87\pm6.09$ & $77.64\pm2.13$ & $77.92\pm2.63$ & $76.09\pm1.60$ & $44.21\pm1.41$ & $43.57\pm0.23$ & $74.17\pm0.26$ & $76.04\pm1.32$ & $71.35\pm0.31$ & $55.22\pm6.61$ & $48.50\pm4.07$ \\
 & \privateset & $68.35\pm3.18$ & $62.69\pm0.28$ & $59.71\pm4.98$ & $49.09\pm3.80$ & $53.22\pm6.59$ & $86.93\pm2.07$ & $73.35\pm8.89$ & $52.70\pm7.45$ & $73.66\pm11.15$ & $49.31\pm7.46$ & $74.68\pm15.51$ & $60.46\pm9.81$ & $58.60\pm2.53$ & $67.81\pm5.72$ & $62.05\pm7.41$ & $73.91\pm2.25$ & $63.89\pm4.96$ & $85.38\pm5.99$ & $82.40\pm9.16$ & $75.13\pm20.65$ \\
 & \wavelet & $71.89\pm12.70$ & $32.68\pm1.89$ & $17.55\pm0.24$ & $17.50\pm0.25$ & $17.12\pm0.50$ & $34.50\pm1.06$ & $25.48\pm1.10$ & $19.12\pm0.31$ & $15.24\pm0.51$ & $14.09\pm0.36$ & $49.09\pm2.86$ & $26.15\pm1.12$ & $19.99\pm1.00$ & $14.44\pm0.73$ & $\textbf{11.53}\pm0.75$ & $89.30\pm0.07$ & $89.65\pm0.33$ & $91.13\pm0.40$ & $17.91\pm1.62$ & $14.67\pm1.31$ \\
 & \sigmoid & $84.79\pm1.74$ & $26.01\pm1.07$ & $\textbf{12.85}\pm0.29$ & $\textbf{10.99}\pm0.32$ & $11.00\pm0.08$ & $33.38\pm0.99$ & $25.50\pm0.71$ & $18.79\pm0.76$ & $14.00\pm0.90$ & $13.14\pm0.85$ & $64.14\pm1.88$ & $24.70\pm0.67$ & $20.01\pm0.73$ & $14.52\pm0.48$ & $12.42\pm0.79$ & $90.08\pm0.09$ & $89.81\pm0.06$ & $89.31\pm0.19$ & $19.38\pm0.20$ & $\textbf{11.11}\pm0.04$ \\
 & \loss & $89.94\pm0.47$ & $38.54\pm1.12$ & $18.85\pm0.04$ & $14.44\pm0.65$ & $13.95\pm0.43$ & $43.65\pm3.44$ & $25.48\pm0.61$ & $19.46\pm0.65$ & $14.37\pm0.20$ & $12.88\pm0.35$ & $59.75\pm6.38$ & $26.32\pm1.17$ & $20.31\pm1.31$ & $14.48\pm0.95$ & $11.72\pm0.30$ & $79.11\pm0.91$ & $89.53\pm0.47$ & $87.89\pm2.39$ & $28.78\pm9.79$ & $14.35\pm0.71$ \\
 & \dpsgd & $77.79\pm17.00$ & $31.26\pm1.65$ & $17.24\pm0.62$ & $15.72\pm0.56$ & $15.83\pm0.58$ & $35.65\pm2.99$ & $24.92\pm1.21$ & $20.21\pm0.77$ & $15.76\pm0.28$ & $13.83\pm0.82$ & $53.24\pm1.26$ & $25.75\pm0.84$ & $20.15\pm0.87$ & $14.51\pm1.33$ & $12.40\pm0.15$ & $88.56\pm0.19$ & $89.34\pm0.25$ & $90.15\pm0.14$ & $29.09\pm8.31$ & $13.02\pm1.54$ \\
 & \alloc & $54.58\pm3.02$ & $28.31\pm1.91$ & $16.03\pm0.59$ & $15.23\pm0.67$ & $15.45\pm0.61$ & $\textbf{31.21}\pm0.90$ & $\textbf{23.88}\pm1.02$ & $20.28\pm0.19$ & $15.63\pm0.25$ & $14.38\pm0.05$ & $41.08\pm6.55$ & $23.67\pm0.35$ & $19.88\pm0.48$ & $14.29\pm0.24$ & $12.68\pm0.21$ & $86.30\pm1.76$ & $90.32\pm0.00$ & $89.82\pm0.24$ & $\textbf{17.14}\pm0.34$ & $12.48\pm0.06$ \\
 & \adpclip & $78.18\pm15.67$ & $73.69\pm11.55$ & $20.30\pm1.37$ & $13.00\pm0.31$ & $13.26\pm0.33$ & $50.48\pm3.76$ & $24.08\pm0.61$ & $18.71\pm0.49$ & $12.65\pm0.15$ & $12.46\pm0.53$ & $70.79\pm5.57$ & $25.55\pm0.19$ & $\textbf{19.02}\pm0.82$ & $\textbf{11.97}\pm0.58$ & $11.81\pm0.78$ & $90.34\pm0.37$ & $90.13\pm0.15$ & $89.43\pm0.03$ & $62.83\pm22.77$ & $14.81\pm0.57$ \\
 & \gep & $89.85\pm0.05$ & $89.85\pm0.05$ & $68.11\pm30.47$ & $11.69\pm0.46$ & $\textbf{10.35}\pm0.54$ & $85.69\pm2.27$ & $33.40\pm1.20$ & $\textbf{17.09}\pm0.99$ & $\textbf{10.60}\pm0.31$ & $\textbf{10.33}\pm0.20$ & - & - & - & - & - & - & - & - & - & - \\
 & \rgp & $\textbf{44.80}\pm4.49$ & $\textbf{23.10}\pm0.76$ & $23.04\pm1.88$ & $21.12\pm1.65$ & $20.67\pm1.66$ & $33.95\pm1.86$ & $30.97\pm2.00$ & $36.36\pm2.44$ & $37.00\pm1.28$ & $37.32\pm1.08$ & $\textbf{28.00}\pm0.73$ & $\textbf{21.33}\pm0.33$ & $21.42\pm1.10$ & $21.98\pm0.65$ & $23.52\pm2.67$ & $89.82\pm0.05$ & $\textbf{23.08}\pm0.45$ & $\textbf{18.28}\pm0.46$ & $20.10\pm0.85$ & $20.70\pm0.42$ \\
 & \pate & $81.62\pm6.09$ & $85.35\pm4.39$ & $51.37\pm3.09$ & $36.89\pm1.88$ & $38.07\pm1.62$ & $88.71\pm3.48$ & $81.72\pm4.08$ & $63.40\pm8.27$ & $42.57\pm7.48$ & $40.15\pm5.16$ & $84.24\pm2.54$ & $79.90\pm3.35$ & $62.70\pm0.67$ & $36.52\pm1.90$ & $37.80\pm1.75$ & $92.41\pm1.76$ & $86.05\pm5.23$ & $76.88\pm15.73$ & $65.12\pm22.83$ & $65.02\pm21.25$ \\
 & \knn & $72.31\pm2.36$ & $55.44\pm2.34$ & $51.94\pm1.77$ & $52.08\pm0.75$ & $50.97\pm0.83$ & $57.59\pm3.89$ & $46.23\pm0.81$ & $44.69\pm0.54$ & $44.55\pm1.04$ & $44.55\pm1.20$ & $55.54\pm1.59$ & $41.12\pm1.26$ & $39.95\pm1.36$ & $40.52\pm0.86$ & $40.12\pm0.86$ & $\textbf{68.04}\pm1.31$ & $47.17\pm1.93$ & $46.16\pm0.65$ & $45.90\pm0.38$ & $45.09\pm0.49$ \\
\cline{1-22}
\multirow[c]{12}{*}{\rotatebox{90}{SVHN}} & \dpgen & $89.23\pm0.13$ & $92.20\pm0.39$ & $91.60\pm1.03$ & $51.49\pm9.52$ & $49.05\pm4.32$ & $88.80\pm0.10$ & $89.02\pm0.77$ & $85.70\pm4.31$ & $85.75\pm5.11$ & $84.56\pm1.80$ & $81.12\pm3.22$ & $86.86\pm1.87$ & $81.70\pm1.76$ & $65.34\pm2.70$ & $58.32\pm3.10$ & $90.93\pm0.90$ & $90.63\pm0.84$ & $89.91\pm1.43$ & $63.70\pm19.35$ & $56.55\pm13.36$ \\
 & \privateset & $88.96\pm4.87$ & $88.36\pm2.55$ & $81.66\pm1.12$ & $79.83\pm0.29$ & $80.94\pm1.15$ & $91.14\pm0.62$ & $88.50\pm0.74$ & $87.49\pm3.02$ & $84.20\pm2.91$ & $85.08\pm2.90$ & $89.60\pm1.67$ & $88.21\pm1.45$ & $77.93\pm1.06$ & $78.17\pm1.98$ & $74.50\pm2.11$ & $90.23\pm1.85$ & $89.34\pm3.79$ & $92.26\pm1.48$ & $90.11\pm3.31$ & $87.44\pm2.08$ \\
 & \wavelet & $\textbf{83.78}\pm4.33$ & $56.72\pm2.80$ & $23.46\pm0.46$ & $17.85\pm0.20$ & $16.77\pm0.75$ & $82.99\pm0.76$ & $36.15\pm0.51$ & $21.09\pm1.58$ & $11.23\pm0.75$ & $9.56\pm1.07$ & $80.24\pm0.11$ & $33.78\pm0.96$ & $16.57\pm0.83$ & $10.44\pm0.28$ & $7.71\pm0.51$ & $87.61\pm3.42$ & $83.85\pm0.70$ & $80.64\pm0.18$ & $80.24\pm0.14$ & $16.50\pm0.56$ \\
 & \sigmoid & $88.96\pm0.38$ & $49.05\pm1.51$ & $\textbf{22.64}\pm0.24$ & $\textbf{12.76}\pm0.16$ & $12.11\pm0.22$ & $86.90\pm0.30$ & $80.12\pm0.51$ & $56.90\pm0.35$ & $40.32\pm3.74$ & $40.07\pm5.48$ & $80.82\pm1.03$ & $46.88\pm0.32$ & $20.66\pm0.61$ & $12.46\pm0.66$ & $9.35\pm0.22$ & $89.75\pm0.13$ & $90.83\pm0.39$ & $88.21\pm1.12$ & $61.78\pm3.59$ & $\textbf{16.02}\pm0.68$ \\
 & \loss & $85.07\pm3.75$ & $74.53\pm4.55$ & $27.98\pm0.45$ & $17.12\pm0.36$ & $15.11\pm0.27$ & $81.79\pm0.68$ & $36.78\pm3.60$ & $22.59\pm0.74$ & $12.54\pm0.97$ & $9.80\pm0.25$ & $81.85\pm0.41$ & $38.60\pm1.63$ & $18.12\pm0.52$ & $10.53\pm0.18$ & $\textbf{7.67}\pm0.33$ & $90.38\pm4.59$ & $82.58\pm1.02$ & $80.60\pm0.01$ & $80.34\pm0.14$ & $17.56\pm0.52$ \\
 & \dpsgd & $85.22\pm3.90$ & $71.58\pm4.43$ & $23.23\pm0.10$ & $16.93\pm0.80$ & $15.46\pm0.95$ & $83.27\pm1.93$ & $34.64\pm5.15$ & $18.68\pm1.01$ & $11.47\pm1.08$ & $10.43\pm1.56$ & $80.25\pm0.16$ & $32.68\pm2.33$ & $\textbf{15.40}\pm0.84$ & $9.75\pm0.16$ & $8.58\pm0.20$ & $89.94\pm0.77$ & $83.92\pm1.92$ & $80.42\pm0.28$ & $80.38\pm0.07$ & $17.19\pm0.13$ \\
 & \alloc & $84.76\pm3.26$ & $80.19\pm0.13$ & $23.95\pm0.48$ & $16.51\pm0.46$ & $15.83\pm0.65$ & $82.66\pm0.68$ & $\textbf{33.80}\pm2.21$ & $20.45\pm0.81$ & $12.23\pm0.73$ & $11.25\pm0.73$ & $80.19\pm0.12$ & $35.80\pm1.66$ & $16.02\pm0.91$ & $10.22\pm0.38$ & $8.31\pm0.23$ & $83.74\pm6.73$ & $80.77\pm0.26$ & $80.40\pm0.05$ & $27.93\pm3.34$ & $20.09\pm1.53$ \\
 & \adpclip & $89.67\pm3.85$ & $83.11\pm1.84$ & $44.26\pm3.69$ & $14.78\pm0.27$ & $14.86\pm0.65$ & $88.90\pm0.57$ & $38.17\pm3.19$ & $17.06\pm0.66$ & $9.27\pm0.61$ & $9.31\pm0.81$ & $82.00\pm1.83$ & $42.49\pm2.77$ & $15.79\pm0.45$ & $\textbf{8.83}\pm0.83$ & $8.38\pm0.13$ & $87.13\pm4.53$ & $90.06\pm0.05$ & $80.42\pm0.11$ & $80.20\pm0.14$ & $22.33\pm0.73$ \\
 & \gep & $89.25\pm5.81$ & $93.17\pm0.01$ & $55.13\pm7.38$ & $14.25\pm0.83$ & $\textbf{11.93}\pm0.66$ & $87.56\pm1.05$ & $89.73\pm1.65$ & $\textbf{15.02}\pm0.24$ & $\textbf{7.77}\pm0.38$ & $\textbf{7.44}\pm0.57$ & - & - & - & - & - & - & - & - & - & - \\
 & \rgp & $93.17\pm0.04$ & $\textbf{45.96}\pm2.54$ & $37.55\pm1.08$ & $29.87\pm1.53$ & $28.75\pm1.73$ & $\textbf{69.61}\pm1.38$ & $49.92\pm3.67$ & $54.89\pm2.85$ & $57.00\pm4.73$ & $57.07\pm5.13$ & $78.54\pm1.42$ & $\textbf{19.12}\pm0.88$ & $17.20\pm1.51$ & $24.74\pm3.14$ & $26.37\pm1.83$ & $93.24\pm0.04$ & $81.11\pm0.77$ & $\textbf{22.03}\pm0.26$ & $\textbf{27.76}\pm0.91$ & $27.16\pm1.08$ \\
 & \pate & $90.64\pm1.66$ & $89.54\pm1.08$ & $82.13\pm3.54$ & $73.59\pm4.85$ & $70.62\pm2.16$ & $90.10\pm0.98$ & $87.74\pm3.17$ & $85.94\pm3.54$ & $73.20\pm2.04$ & $74.73\pm0.54$ & $88.11\pm3.20$ & $90.13\pm1.10$ & $79.08\pm1.78$ & $58.29\pm6.01$ & $63.46\pm2.76$ & $89.05\pm4.66$ & $88.61\pm2.47$ & $79.95\pm0.94$ & $74.80\pm7.42$ & $76.07\pm5.64$ \\
 & \knn & $87.78\pm0.71$ & $77.79\pm1.02$ & $70.58\pm0.65$ & $70.21\pm1.42$ & $69.29\pm0.55$ & $81.30\pm1.82$ & $44.26\pm0.67$ & $40.86\pm1.87$ & $39.84\pm1.14$ & $41.74\pm2.01$ & $\textbf{68.28}\pm5.75$ & $48.09\pm0.69$ & $45.75\pm2.42$ & $44.43\pm1.83$ & $44.55\pm1.06$ & $\textbf{81.14}\pm1.03$ & $\textbf{46.04}\pm0.37$ & $41.76\pm3.15$ & $42.19\pm1.55$ & $40.58\pm0.65$ \\
\bottomrule
\end{tabular}

    }
    \vspace{-0.2cm}
\end{table*}

\mypara{Extra Results of MIAs}
\autoref{tab:mia-white} reports the tailored AUC in white-box style on all model architectures, datasets, and privacy budget.
\autoref{tab:mia-black-apx} reports the tailored AUC in black-box on FMNIST and SVHN.

\begin{table*}[!t]
    \centering
  \caption{Overview of algorithms' tailored AUC in black-box on on FMNIST and SVHN.}.
  \vspace{-0.3cm}
  \resizebox{\linewidth}{!}{
    \begin{tabular}{c|l|ccccc|ccccc|ccccc|ccccc}
\toprule
\multirow{2}{*}{} & \multirow{2}{*}{} & \multicolumn{5}{c|}{SimpleCNN} & \multicolumn{5}{c|}{ResNet} & \multicolumn{5}{c|}{InceptionNet} & \multicolumn{5}{c}{VGG} \\
 &  & 0.2 & 1 & 4 & 100 & 1000 & 0.2 & 1 & 4 & 100 & 1000 & 0.2 & 1 & 4 & 100 & 1000 & 0.2 & 1 & 4 & 100 & 1000 \\
\midrule
\multirow[c]{12}{*}{FMNIST} 
 & \wavelet & $0.50\pm0.00$ & $0.51\pm0.01$ & $0.51\pm0.00$ & $0.51\pm0.00$ & $0.51\pm0.00$ & $0.51\pm0.00$ & $0.50\pm0.01$ & $0.51\pm0.00$ & $0.51\pm0.00$ & $0.50\pm0.00$ & $0.51\pm0.00$ & $0.51\pm0.00$ & $0.51\pm0.00$ & $0.51\pm0.00$ & $0.51\pm0.01$ & $0.50\pm0.00$ & $0.50\pm0.00$ & $0.50\pm0.00$ & $0.51\pm0.01$ & $0.51\pm0.00$ \\
 & \privateset & $0.51\pm0.00$ & $0.51\pm0.00$ & $0.50\pm0.01$ & $0.51\pm0.00$ & $0.51\pm0.00$ & $0.50\pm0.00$ & $0.50\pm0.00$ & $0.51\pm0.00$ & $0.50\pm0.00$ & $0.51\pm0.00$ & $0.50\pm0.00$ & $0.51\pm0.00$ & $0.51\pm0.01$ & $0.51\pm0.00$ & $0.51\pm0.00$ & $0.50\pm0.00$ & $0.50\pm0.00$ & $0.50\pm0.00$ & $0.51\pm0.00$ & $0.51\pm0.00$ \\
 & \dpgen & $0.50\pm0.00$ & $0.50\pm0.00$ & $0.51\pm0.00$ & $0.51\pm0.00$ & $0.51\pm0.00$ & $0.52\pm0.00$ & $0.52\pm0.00$ & $0.52\pm0.00$ & $0.51\pm0.00$ & $0.51\pm0.00$ & $0.51\pm0.00$ & $0.51\pm0.00$ & $0.51\pm0.00$ & $0.52\pm0.00$ & $0.52\pm0.00$ & $0.51\pm0.00$ & $0.51\pm0.00$ & $0.51\pm0.00$ & $0.52\pm0.01$ & $0.51\pm0.00$ \\
 & \sigmoid & $0.50\pm0.00$ & $0.51\pm0.01$ & $0.51\pm0.00$ & $0.51\pm0.01$ & $0.51\pm0.00$ & $0.51\pm0.00$ & $0.51\pm0.01$ & $0.51\pm0.00$ & $0.51\pm0.00$ & $0.52\pm0.00$ & $0.51\pm0.01$ & $0.51\pm0.01$ & $0.51\pm0.01$ & $0.51\pm0.00$ & $0.51\pm0.00$ & $0.50\pm0.00$ & $0.50\pm0.00$ & $0.50\pm0.00$ & $0.51\pm0.01$ & $0.52\pm0.00$ \\
 & \loss & $0.50\pm0.01$ & $0.50\pm0.00$ & $0.51\pm0.00$ & $0.51\pm0.00$ & $0.52\pm0.00$ & $0.51\pm0.00$ & $0.51\pm0.00$ & $0.51\pm0.00$ & $0.51\pm0.00$ & $0.51\pm0.00$ & $0.51\pm0.00$ & $0.51\pm0.01$ & $0.51\pm0.00$ & $0.51\pm0.00$ & $0.51\pm0.00$ & $0.50\pm0.00$ & $0.50\pm0.00$ & $0.51\pm0.01$ & $0.51\pm0.00$ & $0.52\pm0.00$ \\
 & \dpsgd & $0.50\pm0.00$ & $0.51\pm0.00$ & $0.51\pm0.00$ & $0.51\pm0.00$ & $0.51\pm0.00$ & $0.51\pm0.00$ & $0.50\pm0.00$ & $0.51\pm0.00$ & $0.51\pm0.00$ & $0.51\pm0.00$ & $0.51\pm0.00$ & $0.51\pm0.00$ & $0.51\pm0.01$ & $0.51\pm0.00$ & $0.51\pm0.00$ & $0.50\pm0.00$ & $0.50\pm0.00$ & $0.50\pm0.00$ & $0.51\pm0.00$ & $0.52\pm0.00$ \\
 & \rgp & $0.50\pm0.00$ & $0.51\pm0.00$ & $0.51\pm0.01$ & $0.51\pm0.00$ & $0.51\pm0.00$ & $0.51\pm0.01$ & $0.51\pm0.00$ & $0.51\pm0.01$ & $0.51\pm0.00$ & $0.51\pm0.00$ & $0.52\pm0.00$ & $0.51\pm0.00$ & $0.51\pm0.00$ & $0.51\pm0.00$ & $0.51\pm0.01$ & $0.50\pm0.00$ & $0.51\pm0.00$ & $0.52\pm0.00$ & $0.52\pm0.00$ & $0.52\pm0.00$ \\
 & \gep & $0.50\pm0.00$ & $0.50\pm0.00$ & $0.50\pm0.00$ & $0.51\pm0.00$ & $0.53\pm0.01$ & $0.50\pm0.00$ & $0.51\pm0.00$ & $0.51\pm0.00$ & $0.51\pm0.00$ & $0.51\pm0.00$ & - & - & - & - & - & - & - & - & - & - \\
 & \alloc & $0.50\pm0.00$ & $0.51\pm0.00$ & $0.51\pm0.01$ & $0.51\pm0.00$ & $0.51\pm0.00$ & $0.51\pm0.01$ & $0.51\pm0.00$ & $0.51\pm0.01$ & $0.51\pm0.00$ & $0.51\pm0.00$ & $0.51\pm0.00$ & $0.51\pm0.00$ & $0.51\pm0.00$ & $0.51\pm0.01$ & $0.51\pm0.00$ & $0.50\pm0.00$ & $0.50\pm0.00$ & $0.50\pm0.00$ & $0.51\pm0.00$ & $0.51\pm0.00$ \\
 & \adpclip & $0.50\pm0.00$ & $0.50\pm0.00$ & $0.51\pm0.00$ & $0.51\pm0.00$ & $0.51\pm0.00$ & $0.51\pm0.00$ & $0.51\pm0.00$ & $0.50\pm0.00$ & $0.51\pm0.00$ & $0.51\pm0.00$ & $0.51\pm0.01$ & $0.51\pm0.00$ & $0.51\pm0.00$ & $0.51\pm0.00$ & $0.51\pm0.00$ & $0.50\pm0.00$ & $0.50\pm0.00$ & $0.50\pm0.00$ & $0.50\pm0.00$ & $0.52\pm0.00$ \\
 & \pate & $0.50\pm0.00$ & $0.50\pm0.01$ & $0.50\pm0.00$ & $0.52\pm0.01$ & $0.51\pm0.01$ & $0.51\pm0.01$ & $0.50\pm0.00$ & $0.50\pm0.00$ & $0.50\pm0.00$ & $0.50\pm0.00$ & $0.50\pm0.00$ & $0.50\pm0.00$ & $0.50\pm0.01$ & $0.50\pm0.00$ & $0.50\pm0.00$ & $0.52\pm0.01$ & $0.51\pm0.01$ & $0.50\pm0.00$ & $0.50\pm0.00$ & $0.50\pm0.01$ \\
 & \knn & $0.50\pm0.00$ & $0.51\pm0.01$ & $0.51\pm0.01$ & $0.50\pm0.01$ & $0.50\pm0.00$ & $0.50\pm0.00$ & $0.50\pm0.00$ & $0.50\pm0.00$ & $0.50\pm0.00$ & $0.50\pm0.00$ & $0.50\pm0.00$ & $0.50\pm0.00$ & $0.51\pm0.00$ & $0.51\pm0.01$ & $0.50\pm0.00$ & $0.50\pm0.00$ & $0.50\pm0.00$ & $0.50\pm0.00$ & $0.50\pm0.00$ & $0.51\pm0.01$ \\
\cline{1-22}
\multirow[c]{12}{*}{SVHN} 
 & \wavelet & $0.50\pm0.00$ & $0.51\pm0.01$ & $0.51\pm0.00$ & $0.53\pm0.01$ & $0.53\pm0.01$ & $0.50\pm0.00$ & $0.50\pm0.00$ & $0.51\pm0.00$ & $0.52\pm0.01$ & $0.53\pm0.01$ & $0.51\pm0.01$ & $0.50\pm0.00$ & $0.51\pm0.01$ & $0.51\pm0.00$ & $0.52\pm0.01$ & $0.50\pm0.00$ & $0.50\pm0.00$ & $0.52\pm0.00$ & $0.50\pm0.00$ & $0.53\pm0.01$ \\
 & \privateset & $0.51\pm0.01$ & $0.53\pm0.03$ & $0.52\pm0.01$ & $0.53\pm0.01$ & $0.52\pm0.01$ & $0.50\pm0.00$ & $0.51\pm0.01$ & $0.50\pm0.00$ & $0.51\pm0.01$ & $0.51\pm0.01$ & $0.50\pm0.00$ & $0.51\pm0.01$ & $0.51\pm0.00$ & $0.50\pm0.00$ & $0.51\pm0.00$ & $0.54\pm0.02$ & $0.53\pm0.02$ & $0.52\pm0.01$ & $0.54\pm0.01$ & $0.54\pm0.01$ \\
 & \dpgen & $0.50\pm0.00$ & $0.50\pm0.01$ & $0.52\pm0.01$ & $0.50\pm0.01$ & $0.50\pm0.00$ & $0.54\pm0.01$ & $0.50\pm0.00$ & $0.51\pm0.01$ & $0.51\pm0.00$ & $0.52\pm0.01$ & $0.50\pm0.00$ & $0.51\pm0.01$ & $0.52\pm0.01$ & $0.52\pm0.02$ & $0.52\pm0.01$ & $0.50\pm0.00$ & $0.50\pm0.00$ & $0.52\pm0.01$ & $0.53\pm0.00$ & $0.51\pm0.01$ \\
 & \sigmoid & $0.50\pm0.00$ & $0.51\pm0.01$ & $0.53\pm0.00$ & $0.54\pm0.01$ & $0.53\pm0.01$ & $0.50\pm0.00$ & $0.52\pm0.01$ & $0.53\pm0.00$ & $0.56\pm0.00$ & $0.57\pm0.01$ & $0.51\pm0.02$ & $0.50\pm0.00$ & $0.51\pm0.01$ & $0.52\pm0.00$ & $0.53\pm0.00$ & $0.50\pm0.00$ & $0.50\pm0.00$ & $0.50\pm0.00$ & $0.51\pm0.00$ & $0.53\pm0.00$ \\
 & \loss & $0.50\pm0.00$ & $0.51\pm0.01$ & $0.52\pm0.01$ & $0.53\pm0.00$ & $0.52\pm0.00$ & $0.51\pm0.01$ & $0.50\pm0.00$ & $0.50\pm0.00$ & $0.51\pm0.00$ & $0.52\pm0.00$ & $0.51\pm0.02$ & $0.50\pm0.00$ & $0.50\pm0.00$ & $0.52\pm0.01$ & $0.52\pm0.00$ & $0.50\pm0.00$ & $0.50\pm0.00$ & $0.51\pm0.00$ & $0.50\pm0.00$ & $0.53\pm0.00$ \\
 & \dpsgd & $0.50\pm0.00$ & $0.52\pm0.00$ & $0.51\pm0.00$ & $0.53\pm0.01$ & $0.53\pm0.01$ & $0.51\pm0.00$ & $0.50\pm0.00$ & $0.50\pm0.00$ & $0.52\pm0.00$ & $0.53\pm0.01$ & $0.51\pm0.02$ & $0.50\pm0.00$ & $0.50\pm0.00$ & $0.52\pm0.01$ & $0.53\pm0.00$ & $0.50\pm0.00$ & $0.50\pm0.00$ & $0.51\pm0.00$ & $0.50\pm0.00$ & $0.52\pm0.00$ \\
 & \rgp & $0.50\pm0.00$ & $0.51\pm0.01$ & $0.51\pm0.00$ & $0.50\pm0.00$ & $0.51\pm0.00$ & $0.50\pm0.00$ & $0.50\pm0.00$ & $0.50\pm0.00$ & $0.51\pm0.02$ & $0.51\pm0.02$ & $0.51\pm0.01$ & $0.50\pm0.00$ & $0.50\pm0.00$ & $0.50\pm0.00$ & $0.50\pm0.00$ & $0.50\pm0.00$ & $0.51\pm0.01$ & $0.51\pm0.00$ & $0.51\pm0.00$ & $0.51\pm0.00$ \\
 & \gep & $0.50\pm0.00$ & $0.50\pm0.00$ & $0.51\pm0.01$ & $0.51\pm0.00$ & $0.52\pm0.00$ & $0.51\pm0.00$ & $0.50\pm0.00$ & $0.51\pm0.01$ & $0.52\pm0.01$ & $0.51\pm0.01$ & - & - & - & - & - & - & - & - & - & - \\
 & \alloc & $0.50\pm0.00$ & $0.50\pm0.00$ & $0.52\pm0.01$ & $0.52\pm0.01$ & $0.53\pm0.01$ & $0.50\pm0.00$ & $0.50\pm0.00$ & $0.50\pm0.01$ & $0.52\pm0.00$ & $0.53\pm0.01$ & $0.51\pm0.01$ & $0.50\pm0.00$ & $0.50\pm0.00$ & $0.52\pm0.00$ & $0.52\pm0.01$ & $0.50\pm0.00$ & $0.50\pm0.00$ & $0.50\pm0.00$ & $0.51\pm0.01$ & $0.53\pm0.00$ \\
 & \adpclip & $0.50\pm0.00$ & $0.50\pm0.00$ & $0.51\pm0.01$ & $0.53\pm0.01$ & $0.53\pm0.01$ & $0.51\pm0.00$ & $0.50\pm0.00$ & $0.51\pm0.00$ & $0.53\pm0.01$ & $0.53\pm0.01$ & $0.51\pm0.01$ & $0.50\pm0.00$ & $0.51\pm0.01$ & $0.51\pm0.00$ & $0.51\pm0.00$ & $0.50\pm0.00$ & $0.50\pm0.00$ & $0.50\pm0.00$ & $0.50\pm0.00$ & $0.52\pm0.00$ \\
 & \pate & $0.52\pm0.01$ & $0.52\pm0.01$ & $0.51\pm0.01$ & $0.50\pm0.00$ & $0.50\pm0.00$ & $0.50\pm0.00$ & $0.50\pm0.00$ & $0.51\pm0.01$ & $0.50\pm0.00$ & $0.50\pm0.00$ & $0.50\pm0.00$ & $0.51\pm0.01$ & $0.50\pm0.00$ & $0.50\pm0.00$ & $0.50\pm0.00$ & $0.52\pm0.01$ & $0.51\pm0.01$ & $0.51\pm0.01$ & $0.53\pm0.01$ & $0.53\pm0.01$ \\
 & \knn & $0.52\pm0.01$ & $0.51\pm0.01$ & $0.51\pm0.00$ & $0.50\pm0.00$ & $0.50\pm0.00$ & $0.50\pm0.00$ & $0.50\pm0.00$ & $0.50\pm0.00$ & $0.50\pm0.00$ & $0.50\pm0.00$ & $0.50\pm0.00$ & $0.51\pm0.01$ & $0.51\pm0.00$ & $0.51\pm0.01$ & $0.51\pm0.02$ & $0.51\pm0.00$ & $0.52\pm0.01$ & $0.52\pm0.01$ & $0.51\pm0.01$ & $0.50\pm0.00$ \\
\bottomrule
\end{tabular}

  }
  \label{tab:mia-black-apx}
\end{table*}

\begin{table*}[!t]
  \centering
  \caption{Overview of algorithms' tailored AUC in white-box style on all model architectures, datasets, and privacy budget}.

  \vspace{-0.3cm}
  \resizebox{\linewidth}{!}{
    \begin{tabular}{c|l|ccccc|ccccc|ccccc|ccccc}
\toprule
\multirow{2}{*}{} & \multirow{2}{*}{} & \multicolumn{5}{c|}{SimpleCNN} & \multicolumn{5}{c|}{ResNet} & \multicolumn{5}{c|}{InceptionNet} & \multicolumn{5}{c}{VGG} \\
 &  & 0.2 & 1 & 4 & 100 & 1000 & 0.2 & 1 & 4 & 100 & 1000 & 0.2 & 1 & 4 & 100 & 1000 & 0.2 & 1 & 4 & 100 & 1000 \\
\midrule
\multirow[c]{12}{*}{\rotatebox{90}{FMNIST}} 
 & \wavelet & $\textbf{0.50}\pm0.00$ & $\textbf{0.50}\pm0.00$ & $0.51\pm0.00$ & $0.51\pm0.00$ & $0.51\pm0.01$ & $0.51\pm0.00$ & $0.51\pm0.00$ & $0.51\pm0.00$ & $\textbf{0.50}\pm0.00$ & $\textbf{0.50}\pm0.00$ & $\textbf{0.50}\pm0.00$ & $\textbf{0.50}\pm0.00$ & $\textbf{0.50}\pm0.00$ & $\textbf{0.50}\pm0.00$ & $\textbf{0.50}\pm0.00$ & $\textbf{0.50}\pm0.00$ & $\textbf{0.50}\pm0.00$ & $\textbf{0.50}\pm0.00$ & $\textbf{0.50}\pm0.00$ & $0.51\pm0.01$ \\
 & \privateset & $0.51\pm0.00$ & $\textbf{0.50}\pm0.00$ & $\textbf{0.50}\pm0.00$ & $\textbf{0.50}\pm0.00$ & $0.51\pm0.00$ & $\textbf{0.50}\pm0.00$ & $\textbf{0.50}\pm0.00$ & $0.51\pm0.00$ & $\textbf{0.50}\pm0.00$ & $0.51\pm0.01$ & $\textbf{0.50}\pm0.00$ & $\textbf{0.50}\pm0.00$ & $\textbf{0.50}\pm0.00$ & $\textbf{0.50}\pm0.00$ & $0.51\pm0.00$ & $\textbf{0.50}\pm0.00$ & $\textbf{0.50}\pm0.00$ & $\textbf{0.50}\pm0.00$ & $\textbf{0.50}\pm0.00$ & $\textbf{0.50}\pm0.00$ \\
& \dpgen & $\textbf{0.50}\pm0.00$ & $\textbf{0.50}\pm0.00$ & $\textbf{0.50}\pm0.00$ & $\textbf{0.50}\pm0.00$ & $\textbf{0.50}\pm0.00$ & $0.51\pm0.00$ & $0.51\pm0.00$ & $0.51\pm0.00$ & $0.51\pm0.01$ & $0.51\pm0.00$ & $\textbf{0.50}\pm0.00$ & $\textbf{0.50}\pm0.00$ & $\textbf{0.50}\pm0.00$ & $\textbf{0.50}\pm0.00$ & $\textbf{0.50}\pm0.00$ & $\textbf{0.50}\pm0.00$ & $\textbf{0.50}\pm0.00$ & $\textbf{0.50}\pm0.00$ & $\textbf{0.50}\pm0.00$ & $0.51\pm0.01$ \\
 & \sigmoid & $\textbf{0.50}\pm0.00$ & $0.51\pm0.00$ & $\textbf{0.50}\pm0.00$ & $\textbf{0.50}\pm0.00$ & $\textbf{0.50}\pm0.00$ & $\textbf{0.50}\pm0.00$ & $0.51\pm0.00$ & $0.51\pm0.01$ & $0.51\pm0.00$ & $0.51\pm0.00$ & $\textbf{0.50}\pm0.00$ & $\textbf{0.50}\pm0.00$ & $\textbf{0.50}\pm0.00$ & $\textbf{0.50}\pm0.00$ & $\textbf{0.50}\pm0.00$ & $\textbf{0.50}\pm0.00$ & $\textbf{0.50}\pm0.00$ & $\textbf{0.50}\pm0.00$ & $0.51\pm0.00$ & $0.51\pm0.00$ \\
 & \loss & $\textbf{0.50}\pm0.00$ & $\textbf{0.50}\pm0.01$ & $0.51\pm0.01$ & $0.51\pm0.00$ & $0.51\pm0.00$ & $\textbf{0.50}\pm0.00$ & $\textbf{0.50}\pm0.00$ & $\textbf{0.50}\pm0.00$ & $\textbf{0.50}\pm0.00$ & $0.51\pm0.00$ & $0.51\pm0.00$ & $\textbf{0.50}\pm0.01$ & $\textbf{0.50}\pm0.01$ & $\textbf{0.50}\pm0.00$ & $\textbf{0.50}\pm0.00$ & $\textbf{0.50}\pm0.00$ & $\textbf{0.50}\pm0.00$ & $\textbf{0.50}\pm0.00$ & $0.51\pm0.00$ & $0.51\pm0.00$ \\
 & \dpsgd & $\textbf{0.50}\pm0.00$ & $\textbf{0.50}\pm0.01$ & $0.51\pm0.00$ & $0.51\pm0.00$ & $0.51\pm0.00$ & $\textbf{0.50}\pm0.00$ & $0.51\pm0.00$ & $0.51\pm0.00$ & $0.51\pm0.00$ & $0.51\pm0.00$ & $\textbf{0.50}\pm0.00$ & $\textbf{0.50}\pm0.00$ & $\textbf{0.50}\pm0.00$ & $\textbf{0.50}\pm0.00$ & $\textbf{0.50}\pm0.00$ & $\textbf{0.50}\pm0.00$ & $\textbf{0.50}\pm0.00$ & $\textbf{0.50}\pm0.00$ & $0.51\pm0.01$ & $0.51\pm0.01$ \\
 & \rgp & $\textbf{0.50}\pm0.00$ & $\textbf{0.50}\pm0.00$ & $\textbf{0.50}\pm0.00$ & $\textbf{0.50}\pm0.00$ & $\textbf{0.50}\pm0.00$ & $\textbf{0.50}\pm0.00$ & $\textbf{0.50}\pm0.00$ & $\textbf{0.50}\pm0.00$ & $\textbf{0.50}\pm0.00$ & $\textbf{0.50}\pm0.00$ & $\textbf{0.50}\pm0.01$ & $\textbf{0.50}\pm0.00$ & $\textbf{0.50}\pm0.00$ & $\textbf{0.50}\pm0.00$ & $\textbf{0.50}\pm0.00$ & $\textbf{0.50}\pm0.00$ & $\textbf{0.50}\pm0.00$ & $\textbf{0.50}\pm0.00$ & $\textbf{0.50}\pm0.00$ & $\textbf{0.50}\pm0.00$ \\
 & \gep & $\textbf{0.50}\pm0.00$ & $\textbf{0.50}\pm0.00$ & $\textbf{0.50}\pm0.00$ & $0.51\pm0.00$ & $0.52\pm0.00$ & $\textbf{0.50}\pm0.00$ & $0.51\pm0.00$ & $0.51\pm0.00$ & $\textbf{0.50}\pm0.00$ & $\textbf{0.50}\pm0.00$ & - & - & - & - & - & - & - & - & - & - \\
 & \alloc & $\textbf{0.50}\pm0.00$ & $\textbf{0.50}\pm0.00$ & $0.51\pm0.00$ & $0.51\pm0.01$ & $0.51\pm0.00$ & $\textbf{0.50}\pm0.00$ & $0.51\pm0.00$ & $0.51\pm0.00$ & $0.51\pm0.00$ & $0.51\pm0.00$ & $\textbf{0.50}\pm0.00$ & $\textbf{0.50}\pm0.00$ & $\textbf{0.50}\pm0.00$ & $\textbf{0.50}\pm0.00$ & $\textbf{0.50}\pm0.00$ & $\textbf{0.50}\pm0.00$ & $\textbf{0.50}\pm0.00$ & $\textbf{0.50}\pm0.00$ & $0.51\pm0.01$ & $0.51\pm0.01$ \\
 & \adpclip & $\textbf{0.50}\pm0.00$ & $\textbf{0.50}\pm0.00$ & $\textbf{0.50}\pm0.00$ & $0.51\pm0.00$ & $0.51\pm0.00$ & $\textbf{0.50}\pm0.00$ & $0.51\pm0.00$ & $0.51\pm0.00$ & $0.51\pm0.00$ & $0.51\pm0.00$ & $\textbf{0.50}\pm0.00$ & $\textbf{0.50}\pm0.00$ & $\textbf{0.50}\pm0.00$ & $\textbf{0.50}\pm0.00$ & $\textbf{0.50}\pm0.00$ & $\textbf{0.50}\pm0.00$ & $\textbf{0.50}\pm0.00$ & $\textbf{0.50}\pm0.00$ & $\textbf{0.50}\pm0.00$ & $0.51\pm0.01$ \\
 & \pate & $\textbf{0.50}\pm0.00$ & $0.51\pm0.01$ & $\textbf{0.50}\pm0.00$ & $0.51\pm0.00$ & $0.51\pm0.01$ & $\textbf{0.50}\pm0.00$ & $\textbf{0.50}\pm0.00$ & $\textbf{0.50}\pm0.00$ & $\textbf{0.50}\pm0.00$ & $\textbf{0.50}\pm0.00$ & $0.51\pm0.01$ & $0.51\pm0.00$ & $\textbf{0.50}\pm0.00$ & $\textbf{0.50}\pm0.00$ & $\textbf{0.50}\pm0.01$ & $\textbf{0.50}\pm0.00$ & $\textbf{0.50}\pm0.00$ & $\textbf{0.50}\pm0.00$ & $\textbf{0.50}\pm0.00$ & $\textbf{0.50}\pm0.00$ \\
 & \knn & $\textbf{0.50}\pm0.00$ & $\textbf{0.50}\pm0.00$ & $\textbf{0.50}\pm0.00$ & $\textbf{0.50}\pm0.00$ & $\textbf{0.50}\pm0.00$ & $0.51\pm0.00$ & $0.51\pm0.00$ & $\textbf{0.50}\pm0.00$ & $\textbf{0.50}\pm0.00$ & $\textbf{0.50}\pm0.00$ & $\textbf{0.50}\pm0.00$ & $0.51\pm0.00$ & $0.51\pm0.01$ & $0.51\pm0.01$ & $\textbf{0.50}\pm0.00$ & $0.51\pm0.00$ & $\textbf{0.50}\pm0.01$ & $\textbf{0.50}\pm0.00$ & $\textbf{0.50}\pm0.00$ & $\textbf{0.50}\pm0.00$ \\
\cline{1-22}
\multirow[c]{12}{*}{\rotatebox{90}{SVHN}} 
 & \wavelet & $\textbf{0.50}\pm0.00$ & $\textbf{0.50}\pm0.00$ & $0.51\pm0.01$ & $0.52\pm0.01$ & $0.52\pm0.01$ & $\textbf{0.50}\pm0.00$ & $\textbf{0.50}\pm0.00$ & $0.51\pm0.01$ & $0.53\pm0.01$ & $0.53\pm0.00$ & $\textbf{0.50}\pm0.00$ & $\textbf{0.50}\pm0.00$ & $0.52\pm0.00$ & $0.53\pm0.00$ & $0.53\pm0.00$ & $\textbf{0.50}\pm0.00$ & $\textbf{0.50}\pm0.00$ & $\textbf{0.50}\pm0.00$ & $\textbf{0.50}\pm0.00$ & $0.52\pm0.00$ \\
 & \privateset & $\textbf{0.50}\pm0.00$ & $\textbf{0.50}\pm0.00$ & $\textbf{0.50}\pm0.00$ & $\textbf{0.50}\pm0.00$ & $\textbf{0.50}\pm0.00$ & $\textbf{0.50}\pm0.00$ & $\textbf{0.50}\pm0.00$ & $\textbf{0.50}\pm0.00$ & $\textbf{0.50}\pm0.00$ & $\textbf{0.50}\pm0.01$ & $\textbf{0.50}\pm0.00$ & $\textbf{0.50}\pm0.00$ & $\textbf{0.50}\pm0.00$ & $\textbf{0.50}\pm0.00$ & $\textbf{0.50}\pm0.00$ & $\textbf{0.50}\pm0.00$ & $0.51\pm0.02$ & $\textbf{0.50}\pm0.00$ & $0.51\pm0.01$ & $\textbf{0.50}\pm0.00$ \\
 & \dpgen & $\textbf{0.50}\pm0.00$ & $\textbf{0.50}\pm0.00$ & $\textbf{0.50}\pm0.00$ & $\textbf{0.50}\pm0.00$ & $\textbf{0.50}\pm0.00$ & $\textbf{0.50}\pm0.00$ & $\textbf{0.50}\pm0.00$ & $0.51\pm0.02$ & $\textbf{0.50}\pm0.00$ & $\textbf{0.50}\pm0.00$ & $\textbf{0.50}\pm0.00$ & $\textbf{0.50}\pm0.00$ & $\textbf{0.50}\pm0.00$ & $\textbf{0.50}\pm0.01$ & $\textbf{0.50}\pm0.00$ & $\textbf{0.50}\pm0.00$ & $\textbf{0.50}\pm0.00$ & $\textbf{0.50}\pm0.00$ & $\textbf{0.50}\pm0.00$ & $\textbf{0.50}\pm0.00$ \\
 & \sigmoid & $\textbf{0.50}\pm0.00$ & $0.51\pm0.01$ & $0.53\pm0.02$ & $0.55\pm0.03$ & $0.55\pm0.03$ & $\textbf{0.50}\pm0.00$ & $\textbf{0.50}\pm0.00$ & $0.51\pm0.02$ & $0.54\pm0.01$ & $0.56\pm0.01$ & $\textbf{0.50}\pm0.00$ & $\textbf{0.50}\pm0.00$ & $0.53\pm0.02$ & $0.52\pm0.01$ & $0.53\pm0.02$ & $\textbf{0.50}\pm0.00$ & $\textbf{0.50}\pm0.00$ & $\textbf{0.50}\pm0.00$ & $\textbf{0.50}\pm0.00$ & $0.51\pm0.01$ \\
 & \loss & $\textbf{0.50}\pm0.00$ & $\textbf{0.50}\pm0.00$ & $0.51\pm0.01$ & $0.51\pm0.01$ & $0.51\pm0.01$ & $0.52\pm0.02$ & $0.51\pm0.01$ & $0.51\pm0.01$ & $\textbf{0.50}\pm0.00$ & $0.51\pm0.01$ & $\textbf{0.50}\pm0.00$ & $0.51\pm0.00$ & $\textbf{0.50}\pm0.01$ & $0.51\pm0.01$ & $0.51\pm0.01$ & $\textbf{0.50}\pm0.00$ & $\textbf{0.50}\pm0.00$ & $\textbf{0.50}\pm0.00$ & $\textbf{0.50}\pm0.00$ & $0.52\pm0.01$ \\
 & \dpsgd & $\textbf{0.50}\pm0.00$ & $\textbf{0.50}\pm0.00$ & $0.51\pm0.00$ & $0.53\pm0.02$ & $0.53\pm0.02$ & $0.51\pm0.01$ & $\textbf{0.50}\pm0.00$ & $\textbf{0.50}\pm0.00$ & $0.52\pm0.01$ & $0.53\pm0.01$ & $\textbf{0.50}\pm0.00$ & $\textbf{0.50}\pm0.00$ & $0.51\pm0.00$ & $0.52\pm0.01$ & $0.52\pm0.01$ & $\textbf{0.50}\pm0.00$ & $\textbf{0.50}\pm0.00$ & $\textbf{0.50}\pm0.00$ & $\textbf{0.50}\pm0.00$ & $0.51\pm0.01$ \\
 & \rgp & $\textbf{0.50}\pm0.00$ & $\textbf{0.50}\pm0.01$ & $0.51\pm0.01$ & $0.51\pm0.01$ & $0.51\pm0.01$ & $\textbf{0.50}\pm0.00$ & $\textbf{0.50}\pm0.00$ & $\textbf{0.50}\pm0.00$ & $\textbf{0.50}\pm0.00$ & $\textbf{0.50}\pm0.00$ & $\textbf{0.50}\pm0.00$ & $\textbf{0.50}\pm0.00$ & $\textbf{0.50}\pm0.00$ & $\textbf{0.50}\pm0.00$ & $\textbf{0.50}\pm0.00$ & $\textbf{0.50}\pm0.00$ & $\textbf{0.50}\pm0.00$ & $\textbf{0.50}\pm0.00$ & $\textbf{0.50}\pm0.00$ & $\textbf{0.50}\pm0.00$ \\
 & \gep & $\textbf{0.50}\pm0.00$ & $\textbf{0.50}\pm0.00$ & $\textbf{0.50}\pm0.00$ & $0.52\pm0.01$ & $0.54\pm0.02$ & $0.51\pm0.02$ & $\textbf{0.50}\pm0.00$ & $0.52\pm0.01$ & $0.53\pm0.01$ & $0.52\pm0.01$ & - & - & - & - & - & - & - & - & - & - \\
 & \alloc & $\textbf{0.50}\pm0.00$ & $\textbf{0.50}\pm0.00$ & $0.52\pm0.01$ & $0.53\pm0.01$ & $0.53\pm0.01$ & $\textbf{0.50}\pm0.01$ & $\textbf{0.50}\pm0.00$ & $\textbf{0.50}\pm0.00$ & $0.52\pm0.01$ & $0.52\pm0.00$ & $\textbf{0.50}\pm0.00$ & $0.51\pm0.01$ & $0.52\pm0.01$ & $0.52\pm0.01$ & $0.52\pm0.01$ & $\textbf{0.50}\pm0.00$ & $\textbf{0.50}\pm0.00$ & $\textbf{0.50}\pm0.00$ & $\textbf{0.50}\pm0.00$ & $0.53\pm0.01$ \\
 & \adpclip & $\textbf{0.50}\pm0.00$ & $\textbf{0.50}\pm0.00$ & $0.51\pm0.01$ & $0.53\pm0.02$ & $0.53\pm0.02$ & $0.51\pm0.01$ & $\textbf{0.50}\pm0.00$ & $0.51\pm0.00$ & $0.52\pm0.00$ & $0.53\pm0.01$ & $\textbf{0.50}\pm0.00$ & $\textbf{0.50}\pm0.01$ & $0.51\pm0.00$ & $0.52\pm0.01$ & $0.52\pm0.01$ & $\textbf{0.50}\pm0.00$ & $\textbf{0.50}\pm0.00$ & $\textbf{0.50}\pm0.00$ & $\textbf{0.50}\pm0.00$ & $0.52\pm0.01$ \\
 & \pate & $\textbf{0.50}\pm0.00$ & $\textbf{0.50}\pm0.00$ & $\textbf{0.50}\pm0.00$ & $\textbf{0.50}\pm0.00$ & $\textbf{0.50}\pm0.00$ & $\textbf{0.50}\pm0.00$ & $\textbf{0.50}\pm0.00$ & $\textbf{0.50}\pm0.00$ & $\textbf{0.50}\pm0.00$ & $\textbf{0.50}\pm0.00$ & $\textbf{0.50}\pm0.00$ & $\textbf{0.50}\pm0.00$ & $\textbf{0.50}\pm0.00$ & $\textbf{0.50}\pm0.00$ & $\textbf{0.50}\pm0.00$ & $\textbf{0.50}\pm0.00$ & $\textbf{0.50}\pm0.00$ & $\textbf{0.50}\pm0.00$ & $\textbf{0.50}\pm0.00$ & $\textbf{0.50}\pm0.00$ \\
 & \knn & $\textbf{0.50}\pm0.00$ & $\textbf{0.50}\pm0.00$ & $\textbf{0.50}\pm0.00$ & $\textbf{0.50}\pm0.00$ & $\textbf{0.50}\pm0.00$ & $\textbf{0.50}\pm0.00$ & $\textbf{0.50}\pm0.00$ & $\textbf{0.50}\pm0.00$ & $\textbf{0.50}\pm0.00$ & $\textbf{0.50}\pm0.00$ & $\textbf{0.50}\pm0.00$ & $\textbf{0.50}\pm0.00$ & $\textbf{0.50}\pm0.00$ & $\textbf{0.50}\pm0.00$ & $\textbf{0.50}\pm0.00$ & $\textbf{0.50}\pm0.00$ & $\textbf{0.50}\pm0.00$ & $\textbf{0.50}\pm0.00$ & $\textbf{0.50}\pm0.00$ & $\textbf{0.50}\pm0.00$ \\
\cline{1-22}
\multirow[c]{12}{*}{\rotatebox{90}{CIFAR-10}} 
 & \wavelet & $\textbf{0.50}\pm0.00$ & $\textbf{0.50}\pm0.00$ & $\textbf{0.50}\pm0.00$ & $\textbf{0.50}\pm0.00$ & $0.51\pm0.00$ & $0.51\pm0.00$ & $0.51\pm0.00$ & $0.51\pm0.00$ & $0.52\pm0.00$ & $0.53\pm0.00$ & $\textbf{0.50}\pm0.00$ & $\textbf{0.50}\pm0.00$ & $\textbf{0.50}\pm0.00$ & $0.51\pm0.00$ & $0.52\pm0.00$ & $\textbf{0.50}\pm0.00$ & $\textbf{0.50}\pm0.00$ & $\textbf{0.50}\pm0.00$ & $\textbf{0.50}\pm0.01$ & $0.51\pm0.00$ \\
 & \privateset & $\textbf{0.50}\pm0.00$ & $\textbf{0.50}\pm0.00$ & $\textbf{0.50}\pm0.00$ & $\textbf{0.50}\pm0.00$ & $\textbf{0.50}\pm0.00$ & $0.51\pm0.00$ & $0.51\pm0.00$ & $0.51\pm0.01$ & $0.51\pm0.00$ & $0.51\pm0.00$ & $\textbf{0.50}\pm0.00$ & $\textbf{0.50}\pm0.00$ & $\textbf{0.50}\pm0.00$ & $\textbf{0.50}\pm0.00$ & $\textbf{0.50}\pm0.00$ & $\textbf{0.50}\pm0.00$ & $\textbf{0.50}\pm0.00$ & $\textbf{0.50}\pm0.00$ & $\textbf{0.50}\pm0.00$ & $\textbf{0.50}\pm0.00$ \\
 & \dpgen & $\textbf{0.50}\pm0.00$ & $\textbf{0.50}\pm0.00$ & $\textbf{0.50}\pm0.00$ & $\textbf{0.50}\pm0.00$ & $\textbf{0.50}\pm0.00$ & $0.51\pm0.00$ & $0.51\pm0.00$ & $0.51\pm0.00$ & $0.51\pm0.00$ & $0.51\pm0.00$ & $\textbf{0.50}\pm0.00$ & $\textbf{0.50}\pm0.00$ & $\textbf{0.50}\pm0.00$ & $\textbf{0.50}\pm0.00$ & $\textbf{0.50}\pm0.00$ & $\textbf{0.50}\pm0.00$ & $\textbf{0.50}\pm0.00$ & $\textbf{0.50}\pm0.00$ & $\textbf{0.50}\pm0.00$ & $\textbf{0.50}\pm0.00$ \\
 & \sigmoid & $\textbf{0.50}\pm0.00$ & $\textbf{0.50}\pm0.00$ & $0.51\pm0.00$ & $0.53\pm0.01$ & $0.53\pm0.01$ & $0.51\pm0.01$ & $\textbf{0.50}\pm0.00$ & $0.51\pm0.00$ & $0.52\pm0.00$ & $0.55\pm0.00$ & $\textbf{0.50}\pm0.00$ & $\textbf{0.50}\pm0.00$ & $\textbf{0.50}\pm0.00$ & $0.51\pm0.00$ & $0.53\pm0.00$ & $\textbf{0.50}\pm0.00$ & $\textbf{0.50}\pm0.00$ & $\textbf{0.50}\pm0.00$ & $\textbf{0.50}\pm0.00$ & $0.52\pm0.01$ \\
 & \loss & $\textbf{0.50}\pm0.00$ & $\textbf{0.50}\pm0.00$ & $\textbf{0.50}\pm0.00$ & $\textbf{0.50}\pm0.00$ & $\textbf{0.50}\pm0.00$ & $0.51\pm0.00$ & $0.51\pm0.01$ & $0.51\pm0.00$ & $0.51\pm0.00$ & $0.52\pm0.00$ & $\textbf{0.50}\pm0.00$ & $\textbf{0.50}\pm0.00$ & $\textbf{0.50}\pm0.00$ & $\textbf{0.50}\pm0.00$ & $0.51\pm0.01$ & $\textbf{0.50}\pm0.00$ & $\textbf{0.50}\pm0.00$ & $\textbf{0.50}\pm0.00$ & $\textbf{0.50}\pm0.00$ & $\textbf{0.50}\pm0.00$ \\
 & \dpsgd & $\textbf{0.50}\pm0.00$ & $\textbf{0.50}\pm0.00$ & $\textbf{0.50}\pm0.00$ & $0.51\pm0.00$ & $0.51\pm0.00$ & $0.51\pm0.00$ & $0.51\pm0.00$ & $0.51\pm0.00$ & $0.52\pm0.00$ & $0.53\pm0.00$ & $\textbf{0.50}\pm0.00$ & $\textbf{0.50}\pm0.00$ & $\textbf{0.50}\pm0.00$ & $0.51\pm0.00$ & $0.52\pm0.00$ & $\textbf{0.50}\pm0.00$ & $\textbf{0.50}\pm0.00$ & $\textbf{0.50}\pm0.00$ & $\textbf{0.50}\pm0.00$ & $0.51\pm0.00$ \\
 & \rgp & $\textbf{0.50}\pm0.00$ & $\textbf{0.50}\pm0.00$ & $\textbf{0.50}\pm0.00$ & $\textbf{0.50}\pm0.00$ & $\textbf{0.50}\pm0.00$ & $\textbf{0.50}\pm0.00$ & $\textbf{0.50}\pm0.00$ & $\textbf{0.50}\pm0.01$ & $\textbf{0.50}\pm0.00$ & $\textbf{0.50}\pm0.00$ & $\textbf{0.50}\pm0.00$ & $\textbf{0.50}\pm0.00$ & $\textbf{0.50}\pm0.00$ & $\textbf{0.50}\pm0.00$ & $\textbf{0.50}\pm0.00$ & $\textbf{0.50}\pm0.00$ & $\textbf{0.50}\pm0.00$ & $\textbf{0.50}\pm0.00$ & $\textbf{0.50}\pm0.00$ & $0.51\pm0.00$ \\
 & \gep & $\textbf{0.50}\pm0.00$ & $\textbf{0.50}\pm0.00$ & $\textbf{0.50}\pm0.00$ & $0.51\pm0.00$ & $0.54\pm0.01$ & $0.51\pm0.00$ & $0.51\pm0.00$ & $\textbf{0.50}\pm0.00$ & $0.53\pm0.00$ & $0.54\pm0.00$ & - & - & - & - & - & - & - & - & - & - \\
 & \alloc & $\textbf{0.50}\pm0.00$ & $\textbf{0.50}\pm0.00$ & $\textbf{0.50}\pm0.00$ & $\textbf{0.50}\pm0.00$ & $0.51\pm0.00$ & $0.51\pm0.00$ & $0.51\pm0.01$ & $0.51\pm0.01$ & $0.52\pm0.00$ & $0.53\pm0.00$ & $\textbf{0.50}\pm0.00$ & $\textbf{0.50}\pm0.00$ & $\textbf{0.50}\pm0.00$ & $0.51\pm0.00$ & $0.52\pm0.00$ & $\textbf{0.50}\pm0.00$ & $\textbf{0.50}\pm0.00$ & $\textbf{0.50}\pm0.00$ & $\textbf{0.50}\pm0.00$ & $0.52\pm0.00$ \\
 & \adpclip & $\textbf{0.50}\pm0.00$ & $\textbf{0.50}\pm0.00$ & $\textbf{0.50}\pm0.00$ & $0.51\pm0.00$ & $0.52\pm0.01$ & $0.51\pm0.00$ & $0.51\pm0.00$ & $0.51\pm0.00$ & $0.52\pm0.00$ & $0.54\pm0.01$ & $\textbf{0.50}\pm0.00$ & $\textbf{0.50}\pm0.00$ & $\textbf{0.50}\pm0.00$ & $0.51\pm0.00$ & $0.52\pm0.01$ & $\textbf{0.50}\pm0.00$ & $\textbf{0.50}\pm0.00$ & $\textbf{0.50}\pm0.00$ & $\textbf{0.50}\pm0.00$ & $0.52\pm0.00$ \\
 & \pate & $\textbf{0.50}\pm0.00$ & $\textbf{0.50}\pm0.00$ & $0.51\pm0.00$ & $\textbf{0.50}\pm0.00$ & $\textbf{0.50}\pm0.00$ & $\textbf{0.50}\pm0.00$ & $\textbf{0.50}\pm0.01$ & $\textbf{0.50}\pm0.00$ & $0.51\pm0.00$ & $0.51\pm0.01$ & $\textbf{0.50}\pm0.00$ & $\textbf{0.50}\pm0.00$ & $\textbf{0.50}\pm0.00$ & $\textbf{0.50}\pm0.00$ & $\textbf{0.50}\pm0.00$ & $\textbf{0.50}\pm0.00$ & $\textbf{0.50}\pm0.00$ & $\textbf{0.50}\pm0.00$ & $\textbf{0.50}\pm0.00$ & $\textbf{0.50}\pm0.00$ \\
 & \knn & $\textbf{0.50}\pm0.00$ & $0.51\pm0.00$ & $\textbf{0.50}\pm0.00$ & $\textbf{0.50}\pm0.00$ & $\textbf{0.50}\pm0.00$ & $\textbf{0.50}\pm0.00$ & $\textbf{0.50}\pm0.00$ & $\textbf{0.50}\pm0.00$ & $\textbf{0.50}\pm0.00$ & $\textbf{0.50}\pm0.00$ & $\textbf{0.50}\pm0.00$ & $\textbf{0.50}\pm0.00$ & $\textbf{0.50}\pm0.00$ & $\textbf{0.50}\pm0.00$ & $\textbf{0.50}\pm0.00$ & $\textbf{0.50}\pm0.00$ & $\textbf{0.50}\pm0.00$ & $\textbf{0.50}\pm0.00$ & $0.51\pm0.00$ & $\textbf{0.50}\pm0.00$ \\
\bottomrule
\end{tabular}

  }
  \label{tab:mia-white}
\end{table*}

\section{Model Architectures}
\label{apx:model-details}
\mypara{Target Model}
\autoref{tab:simplenet}, \autoref{tab:resnet}, \autoref{tab:inceptionnet} and \autoref{tab:vgg} show target model architecture, respectively.
For simplicity, the details of the block used in the network are shown in \autoref{tab:resnet-detail} and \autoref{tab:inception-detail}.

\mypara{Attack Model}
We present implementation details of attack models as follows:
\begin{itemize}
    \setlength{\itemsep}{1pt}
    \setlength{\parsep}{1pt}
    \setlength{\parskip}{1pt}
    \item \mypara{Black-Box}
    We refer to the model architecture of Liu \etal~\cite{liu2021ml-doctor}.
    The attack model receives two inputs: the target sample's sorted posteriors and a binary indicator on whether the target sample is predicted correctly.
    The attack model consists of three MLPs (Multi-layer Perceptron).
    Two processes the input to extract features and concatenated output features are fed into the third MLP to obtain the final prediction.
    \item \mypara{White-Box}
    We use a similar model architecture as the one used by Nasr \etal~\cite{nasr2019comprehensive}.
    There are four inputs for this attack model, including the target model's posteriors, classification loss, gradients of the parameters of the target model's last layer, and true labels in one-hot encoding.
    Each input is fed into a different neural network to extract the features respectively, and then the features are passed to the classifier after concatenation.
\end{itemize}

\begin{table}[!t] 
  \centering
  \caption{SimpleNet architecture and details of BasicBlock.}
  \vspace{-0.3cm}
  \resizebox{0.35\linewidth}{!}
  {\begin{tabular}{|c|c|}
    \hline
    \rowcolor[rgb]{ .816,  .808,  .808} \textbf{Layer Type} & \textbf{Architecture} \\
    \hline
    BasicBlock & filters=16 \\
    \hline
    BasicBlock & filters=32 \\
    \hline
    BasicBlock & filters=64 \\
    \hline
    Flatten &  \\
    \hline
    Relu FC & 500 units \\
    \hline
    FC    & 10 units \\
    \hline
\end{tabular}%
}
  \resizebox{0.55\linewidth}{!}
  {    \begin{tabular}{|c|c|}
    \hline
    \rowcolor[rgb]{ .816,  .808,  .808} \textbf{BasicBlock} & \textit{filters} \\
    \hline
    Conv2D & filters, kernel\_size=3, padding=1 \\
    \hline
    Activation & ReLU \\
    \hline
    MaxPooling & kernel\_size=2, stride=2 \\
    \hline
    \end{tabular}%
}
  \label{tab:simplenet}
\end{table}

\begin{table}[htb]
  \centering
  \caption{ResNet architecture.}
  \resizebox{0.45\linewidth}{!}
  {    \begin{tabular}{|c|c|}
    \hline
    \rowcolor[rgb]{ .816,  .808,  .808} \textbf{Layer Type} & \textbf{Architecture} \\
    \hline
    Input Layer & filters=16 \\
    \hline
    ResBlock 1  & filters=16 \\
    \hline
    ResBlock 1  & filters=16 \\
    \hline
    ResBlock 1 & filters=16 \\
    \hline
    ResBlock 2 & filters=32, stride=2 \\
    \hline
    ResBlock 2 & filters=32, stride=2 \\
    \hline
    ResBlock 2 & filters=32, stride=2 \\
    \hline
    ResBlock 2 & filters=64, stride=2 \\
    \hline
    ResBlock 2 & filters=64, stride=2 \\
    \hline
    ResBlock 2 & filters=64, stride=2 \\
    \hline
    AdaptiveAvgPool2D & output\_size=(1,1) \\
    \hline
    FC    & 10 units \\
    \hline
    \end{tabular}%
}
  \label{tab:resnet}
\end{table}

\begin{table}[htb]
  \centering
  \caption{Details of ResBlock for ResNet. (\textit{Shortcut} perform identity mapping, and their outputs are added to the outputs of the stacked layers~\cite{he2016deep})}
    \resizebox{0.75\linewidth}{!}
    {
        \begin{tabular}{|c|c|}
    \hline
    \rowcolor[rgb]{ .816,  .808,  .808} \textbf{Input Layer} & \textit{filters} \\
    \hline
    Conv2D & filters, kernel\_size=3 \\
    \hline
    GroupNorm & num\_groups=4, num\_channels=filters, affine=False \\
    \hline
    Activation & ReLU \\
    \hline
    \rowcolor[rgb]{ .816,  .808,  .808} \textbf{ResBlock 1} & \textit{filters} \\
    \hline
    Conv2D & filters, kernel\_size=3, stride=1 \\
    \hline
    GroupNorm & num\_groups=4, num\_channels=filters \\
    \hline
    Activation & ReLU \\
    \hline
    Conv2D & filters, kernel\_size=3, stride=1 \\
    \hline
    GroupNorm & num\_groups=4, num\_channels=filters \\
    \hline
    Shortcut &  \\
    \hline
    Activation & ReLU \\
    \hline
    \rowcolor[rgb]{ .816,  .808,  .808} \textbf{ResBlock 2} & \textit{filters, stride} \\
    \hline
    Conv2D & filters, kernel\_size=3, stride \\
    \hline
    GroupNorm & num\_groups=4, num\_channels=filters \\
    \hline
    Activation & ReLU \\
    \hline
    Conv2D & filters, kernel\_size=3, stride=1 \\
    \hline
    GroupNorm & num\_groups=4, num\_channels=filters \\
    \hline
    AvgPool2D & kernel\_size=1, stride \\
    \hline
    GroupNorm & num\_groups=4, num\_channels=filters \\
    \hline
    Shortcut &  \\
    \hline
    Activation & ReLU \\
    \hline
    \end{tabular}%
    }
  \label{tab:resnet-detail}
\end{table}

\begin{table}[htb]
  \centering
  \caption{InceptionNet architecture.}
    \resizebox{0.7\linewidth}{!}{
        \begin{tabular}{|c|c|}
    \hline
    \rowcolor[rgb]{ .816,  .808,  .808} \textbf{Layer Type} & \textbf{Architecture} \\
    \hline
    InceptionBlock & filters=32, kernel\_size=3, stride=1 \\
    \hline
    InceptionBlock & filters=32, kernel\_size=3, stride=1 \\
    \hline
    MaxPool2D & kernel\_size=2, stride=1 \\
    \hline
    \multicolumn{2}{|c|}{InceptionA} \\
    \hline
    \multicolumn{2}{|c|}{InceptionB} \\
    \hline
    \multicolumn{2}{|c|}{InceptionC} \\
    \hline
    \multicolumn{2}{|c|}{InceptionD} \\
    \hline
    \multicolumn{2}{|c|}{InceptionE} \\
    \hline
    AdaptiveAvgPool2d & output\_size=(1,1) \\
    \hline
    Dropout & p=0.5 \\
    \hline
    FC    & 10 units \\
    \hline
    \end{tabular}%
    }
  \label{tab:inceptionnet}
\end{table}

\begin{table}[htb]
  \centering
  \caption{VGG architecture and the details of the VGGBlock.}
  \resizebox{0.4\linewidth}{!}
    {    \begin{tabular}{|c|c|}
    \hline
    \rowcolor[rgb]{ .816,  .808,  .808} \textbf{Layer Type} & \textbf{Architecture} \\
    \hline
    VGGBlock & filters=64 \\
    \hline
    MaxPool2D & kernel\_size=2, stride=2 \\
    \hline
    VGGBlock & filters=128 \\
    \hline
    MaxPool2D & kernel\_size=2, stride=2 \\
    \hline
    VGGBlock & filters=256 \\
    \hline
    VGGBlock & filters=256 \\
    \hline
    MaxPool2D & kernel\_size=2, stride=2 \\
    \hline
    VGGBlock & filters=512 \\
    \hline
    VGGBlock & filters=512 \\
    \hline
    MaxPool2D & kernel\_size=2, stride=2 \\
    \hline
    VGGBlock & filters=512 \\
    \hline
    VGGBlock & filters=512 \\
    \hline
    MaxPool2D & kernel\_size=2, stride=2 \\
    \hline
    Flatten &  \\
    \hline
    FC    & 4096 units \\
    \hline
    Activation & ReLU \\
    \hline
    Dropout & p=0.5 \\
    \hline
    FC    & 4096 units \\
    \hline
    Activation & ReLU \\
    \hline
    Dropout & p=0.5 \\
    \hline
    FC    & 10 units \\
    \hline
    \end{tabular}%
}
  \resizebox{0.5\linewidth}{!}
    {\begin{tabular}{|c|c|}
    \hline
    \rowcolor[rgb]{ .816,  .808,  .808} \textbf{VGGBlock} & \textit{filters} \\
    \hline
    Conv2D & filters, kernel\_size=3, padding=1 \\
    \hline
    Activation & ReLU \\
    \hline
    \end{tabular}%
}
  \label{tab:vgg}
\end{table}

\begin{table*}[htb]
    \centering
    \caption{Details of InceptionBlock for InceptionNet.}
    \label{tab:inception-detail}
    \resizebox{0.9\linewidth}{!}{
            \begin{tabular}{|cccccccc|}
    \hline
    \rowcolor[rgb]{ .816,  .808,  .808} \multicolumn{1}{|c|}{\textbf{InceptionBlock}} & \multicolumn{7}{c|}{\textit{filters, kernel\_size, padding}} \\
    \hline
    \multicolumn{1}{|c|}{Conv2D} & \multicolumn{7}{c|}{\textit{filters, kernel\_size, padding}} \\
    \hline
    \multicolumn{1}{|c|}{GroupNorm} & \multicolumn{7}{c|}{num\_groups=4, num\_channels=filters} \\
    \hline
    \multicolumn{1}{|c|}{Activation} & \multicolumn{7}{c|}{ReLU} \\
    \hline
    \rowcolor[rgb]{ .816,  .808,  .808} \multicolumn{1}{|c|}{\textbf{InceptionA}} & \multicolumn{7}{c|}{} \\
    \hline
    \multicolumn{1}{|l|}{InceptionBlock} & \multicolumn{1}{l|}{filters=32, kernel\_size=1} & \multicolumn{1}{c|}{InceptionBlock} & \multicolumn{1}{l|}{filters=24, kernel\_size=1} & \multicolumn{1}{c|}{InceptionBlock} & \multicolumn{1}{l|}{filters=32, kernel\_size=1} & \multicolumn{1}{l|}{AvgPool2D} & \multicolumn{1}{l|}{kernel\_size=3,stride=1,padding=1} \\
    \hline
    \multicolumn{1}{|r|}{} & \multicolumn{1}{r|}{} & \multicolumn{1}{c|}{InceptionBlock} & \multicolumn{1}{l|}{filters=32, kernel\_size=5, padding=2} & \multicolumn{1}{c|}{InceptionBlock} & \multicolumn{1}{l|}{filters=48, kernel\_size=3, padding=1} & \multicolumn{1}{l|}{InceptionBlock} & \multicolumn{1}{l|}{filters=16, kernel\_size=1} \\
    \hline
    \multicolumn{1}{|r|}{} & \multicolumn{1}{r|}{} & \multicolumn{1}{c|}{} & \multicolumn{1}{r|}{} & \multicolumn{1}{c|}{InceptionBlock} & \multicolumn{1}{l|}{filters=48, kernel\_size=3, padding=1} & \multicolumn{1}{r|}{} &  \\
    \hline
    \multicolumn{8}{|c|}{Concat} \\
    \hline
    \rowcolor[rgb]{ .816,  .808,  .808} \multicolumn{1}{|c|}{\textbf{InceptionB}} & \multicolumn{7}{c|}{} \\
    \hline
    \multicolumn{1}{|c|}{InceptionBlock} & \multicolumn{1}{l|}{filters=96, kernel\_size=3, stride=2} & \multicolumn{1}{c|}{InceptionBlock} & \multicolumn{1}{l|}{filters=32, kernel\_size=1} & \multicolumn{1}{c|}{MaxPool2D} & \multicolumn{1}{l|}{kernel\_size=3, stride=2} & \multicolumn{1}{r|}{} &  \\
    \hline
    \multicolumn{1}{|c|}{} & \multicolumn{1}{r|}{} & \multicolumn{1}{c|}{InceptionBlock} & \multicolumn{1}{l|}{filters=48, kernel\_size=3, padding=1} & \multicolumn{1}{c|}{} & \multicolumn{1}{r|}{} & \multicolumn{1}{r|}{} &  \\
    \hline
    \multicolumn{1}{|c|}{} & \multicolumn{1}{r|}{} & \multicolumn{1}{c|}{InceptionBlock} & \multicolumn{1}{l|}{filters=48, kernel\_size=3, stride=2} & \multicolumn{1}{c|}{} & \multicolumn{1}{r|}{} & \multicolumn{1}{r|}{} &  \\
    \hline
    \multicolumn{8}{|c|}{Concat} \\
    \hline
    \rowcolor[rgb]{ .816,  .808,  .808} \multicolumn{1}{|c|}{\textbf{InceptionC}} & \multicolumn{7}{c|}{} \\
    \hline
    \multicolumn{1}{|c|}{InceptionBlock} & \multicolumn{1}{l|}{filters=48, kernel\_size=1} & \multicolumn{1}{c|}{InceptionBlock} & \multicolumn{1}{l|}{filters=48, kernel\_size=1} & \multicolumn{1}{c|}{AvgPool2D} & \multicolumn{1}{l|}{kernel\_size=3, stride=1, padding=1} & \multicolumn{1}{r|}{} &  \\
    \hline
    \multicolumn{1}{|c|}{} & \multicolumn{1}{r|}{} & \multicolumn{1}{c|}{InceptionBlock} & \multicolumn{1}{l|}{filters=48, kernel\_size=7, padding=3} & \multicolumn{1}{c|}{InceptionBlock} & \multicolumn{1}{l|}{filters=48, kernel\_size=1} & \multicolumn{1}{r|}{} &  \\
    \hline
    \multicolumn{1}{|c|}{} & \multicolumn{1}{r|}{} & \multicolumn{1}{c|}{InceptionBlock} & \multicolumn{1}{l|}{filters=48, kernel\_size=7, padding=3} & \multicolumn{1}{c|}{} & \multicolumn{1}{r|}{} & \multicolumn{1}{r|}{} &  \\
    \hline
    \multicolumn{8}{|c|}{Concat} \\
    \hline
    \rowcolor[rgb]{ .816,  .808,  .808} \multicolumn{1}{|c|}{\textbf{InceptionD}} & \multicolumn{7}{c|}{} \\
    \hline
    \multicolumn{1}{|c|}{InceptionBlock} & \multicolumn{1}{l|}{filters=48, kernel\_size=1} & \multicolumn{1}{c|}{InceptionBlock} & \multicolumn{1}{l|}{filters=96, kernel\_size=1} & \multicolumn{1}{c|}{MaxPool2D} & \multicolumn{1}{l|}{kernal\_size=3, stride=2} & \multicolumn{1}{r|}{} &  \\
    \hline
    \multicolumn{1}{|c|}{InceptionBlock} & \multicolumn{1}{l|}{filters=96, kernel\_size=3, stride=2} & \multicolumn{1}{c|}{InceptionBlock} & \multicolumn{1}{l|}{filters=96, kernel\_size=7, padding=3} & \multicolumn{1}{c|}{} & \multicolumn{1}{r|}{} & \multicolumn{1}{r|}{} &  \\
    \hline
    \multicolumn{1}{|c|}{} & \multicolumn{1}{r|}{} & \multicolumn{1}{c|}{InceptionBlock} & \multicolumn{1}{l|}{filters=96, kernel\_size=3, stride=2} & \multicolumn{1}{c|}{} & \multicolumn{1}{r|}{} & \multicolumn{1}{r|}{} &  \\
    \hline
    \multicolumn{8}{|c|}{Concat} \\
    \hline
    \rowcolor[rgb]{ .816,  .808,  .808} \multicolumn{1}{|c|}{\textbf{InceptionE}} & \multicolumn{7}{c|}{} \\
    \hline
    \multicolumn{1}{|c|}{InceptionBlock} & \multicolumn{1}{l|}{filters=80, kernel\_size=1} & \multicolumn{1}{c|}{InceptionBlock} & \multicolumn{1}{l|}{filters=96, kernel\_size=1} & \multicolumn{1}{c|}{InceptionBlock} & \multicolumn{1}{l|}{filters=112, kernel\_size=1} & \multicolumn{1}{l|}{AvgPool2D} & \multicolumn{1}{l|}{kernel\_size=3,stride=1,padding=1} \\
    \hline
    \multicolumn{1}{|c|}{} & \multicolumn{1}{r|}{} & \multicolumn{1}{c|}{InceptionBlock} & \multicolumn{1}{l|}{filters=96, kernel\_size=3, padding=1} & \multicolumn{1}{c|}{InceptionBlock} & \multicolumn{1}{l|}{filters=96, kernel\_size=3, padding=1} & \multicolumn{1}{l|}{InceptionBlock} & \multicolumn{1}{l|}{filters=48, kernel\_size=1} \\
    \hline
    \multicolumn{1}{|c|}{} & \multicolumn{1}{r|}{} & \multicolumn{1}{c|}{} & \multicolumn{1}{r|}{} & \multicolumn{1}{c|}{InceptionBlock} & \multicolumn{1}{l|}{filters=96, kernel\_size=3, padding=1} & \multicolumn{1}{r|}{} &  \\
    \hline
    \multicolumn{8}{|c|}{Concat} \\
    \hline
    \end{tabular}%

    }
\end{table*}

\end{document}